\documentclass{article}
\usepackage{graphicx}
\usepackage[margin=1in]{geometry}
\usepackage{booktabs}
\usepackage[dvipsnames]{xcolor}
\usepackage{amsmath,amsfonts,amssymb,amsthm}
\usepackage{bm}
\usepackage{subcaption}
\usepackage{multirow}
\usepackage{pifont}
\usepackage[colorlinks=true,linkcolor=blue,citecolor=blue]{hyperref}
\colorlet{mylinkcolor}{violet}
\colorlet{mycitecolor}{YellowOrange}
\colorlet{myurlcolor}{Aquamarine}

\newtheorem{remark}{Remark}

\newcommand{\cmark}{\ding{51}}%
\newcommand{\xmark}{{\color{red}\ding{55}}}%

\providecommand{\keywords}[1]
{
  \small	
  \textbf{\textit{Keywords---}} #1
}

\title{Quantification of total uncertainty in the physics-informed reconstruction of CVSim-6 physiology}
\author{Mario De Florio$^{1}$, Zongren Zou$^{1}$, Daniele E. Schiavazzi$^{2,*}$, George Em Karniadakis$^{1}$}
\date{}

\begin{document}

\maketitle

\begin{center}
    \begin{minipage}{0.8\linewidth}
        \begin{center}
            $^{1}$ Division of Applied Mathematics, Brown University, Providence, 02906 RI, USA \\
            $^{2}$ Department of Applied and Computational Mathematics and Statistics, University of Notre Dame, Notre Dame, 46556 IN, USA \\
            $^{*}$ Corresponding author: dschiavazzi@nd.edu 
        \end{center}
    \end{minipage}
\end{center}

\begin{abstract}
\noindent When predicting physical phenomena through simulation, quantification of the total uncertainty due to multiple sources is as crucial as making sure the underlying numerical model is accurate.
Possible sources include irreducible \emph{aleatoric} uncertainty due to noise in the data, \emph{epistemic} uncertainty induced by insufficient data or inadequate parameterization, and \emph{model-form} uncertainty related to the use of misspecified model equations. 
In addition, recently proposed approaches provide flexible ways to combine information from data with full or partial satisfaction of equations that typically encode physical principles. 
Physics-based regularization interacts in nontrivial ways with aleatoric, epistemic and model-form uncertainty and their combination, and a better understanding of this interaction is needed to improve the predictive performance of physics-informed digital twins that operate under real conditions.
To better understand this interaction, with a specific focus on biological and physiological models, this study investigates the decomposition of total uncertainty in the estimation of states and parameters of a differential system simulated with MC X-TFC, a new physics-informed approach for uncertainty quantification based on random projections and Monte-Carlo sampling.
After an introductory comparison between approaches for physics-informed estimation, MC X-TFC is applied to a six-compartment stiff ODE system, the CVSim-6 model, developed in the context of human physiology. The system is first analyzed by progressively removing data while estimating an increasing number of parameters, and subsequently by investigating total uncertainty under model-form misspecification of non-linear resistance in the pulmonary compartment. In particular, we focus on the interaction between the formulation of the discrepancy term and quantification of model-form uncertainty, and show how additional physics can help in the estimation process. 
The method demonstrates robustness and efficiency in estimating unknown states and parameters, even with limited, sparse, and noisy data. It also offers great flexibility in integrating data with physics for improved estimation, even in cases of model misspecification.
\end{abstract}


\section{Introduction}\label{sec:intro}

Characterizing uncertainty from multiple sources is crucial for developing computational models that can accurately predict real physical phenomena. However, not all sources of uncertainty are equally relevant across all applications.
Consistent with the discussion in~\cite{psaros2023uncertainty}, common sources in scientific machine learning applications relate to the quality of \emph{data}, assumptions in the \emph{equations} formulated to capture physical phenomena, and the \emph{estimator} used to infer relevant states or parameters.
Uncertainty from data is typically referred to as \emph{aleatoric}, and includes the inability to precisely characterize a physical quantity due to inaccurate measurements that manifest as noise, missing or scarce data that exacerbate the ill-posed character of inverse problems, and noise model misspecification. 
Uncertainty related to model misspecification or the effects of disregarding stochasticity is referred to as \emph{model-form} uncertainty. 
In addition, for data-driven estimators, prediction variability due to the network size, hyperparameter selection, and determination of optimal weights and biases in optimization or inference tasks constitute a form of \emph{epistemic} uncertainty. 

In this context, physics-informed neural networks (PINNs)~\cite{raissi2019physics} have emerged as widely used estimators for problems involving differential equations (DE) across diverse disciplines such as fluid mechanics~\cite{eivazi2022physics, raissi2020hidden, cheng2021deep, wessels2020neural, shukla2024neurosem}, epidemiology~\cite{linka2022bayesian, kharazmi2021identifiability, schiassi2021physics, zou2024neuraluq}, and beyond.
Leveraging modern machine learning techniques, including automatic differentiation and efficient optimization methods, PINNs have proven effective in solving differential equations, assimilating data, and tackling ill-posed inverse problems~\cite{lu2021physics,gao2022physics}, as discussed in a number of comprehensive reviews~\cite{cuomo2022scientific, karniadakis2021physics}.
Uncertainty quantification methods for neural networks, such as variational inference~\cite{blei2017variational, hoffman2013stochastic, ranganath2014black}, dropout~\cite{gal2016dropout} and deep ensemble~\cite{lakshminarayanan2017simple}, have also been integrated into PINNs for quantifying uncertainty arising from multiple sources. These include noisy and gappy data~\cite{yang2021b, zou2024leveraging}, physical model misspecification~\cite{zou2024correcting}, noisy inputs~\cite{zou2023uncertainty} and inherent uncertainty of neural network models~\cite{psaros2023uncertainty, zou2023hydra, meng2022learning}.

However, one of the main drawbacks of PINN-based approaches is the computational cost associate with back-propagation and the need to augment the loss function to account for initial and boundary conditions, adding complexity to the process of learning DE solutions. 
An alternative approach, the Theory of Functional Connections (TFC)~\cite{mortari2017theory,mortari2017least,mortari2019high,de2021theory,leake2020multivariate}, offers a \emph{constrained expression} to approximate differential equation solutions while analytically satisfying initial and boundary conditions.
Building on this framework, the eXtreme-TFC approach (X-TFC~\cite{schiassi2021extreme}) combines TFC with random-projection neural networks (a.k.a. extreme learning machine~\cite{huang2006extreme,dwivedi2020physics}), providing a fast and accurate method for inference and prediction in partially observed and possibly misspecified dynamical systems.
Random-projection neural networks have been widely employed in physics-informed frameworks for solving linear and nonlinear partial differential equations \cite{dong2021local, dwivedi2020physics, wang2024extreme, sun2024local, alvarez2024nonlinear, fabiani2023parsimonious, dong2022computing, calabro2021extreme}, inverse problems for parameter estimations and dynamical systems discovery \cite{dong2023method, patsatzis2024physics, liu2023bayesian}, and neural operators \cite{fabiani2024randonet}.
While X-TFC has demonstrated robustness and efficiency in various applications~\cite{drozd2021energy,de2021solutions,johnston2020fuel,drozd2019constrained}, including forward problems~\cite{de2022physics,schiassi2020new,de2021physics,de2022poiseuille,schiassi2022orbit,de2024physics,furfaro2022physics,schiassi2022physics,d2021pontryagin}, inverse problems for the estimation of parameters and missing terms in equations~\cite{schiassi2020physics,ahmadi2024ai}, and combined with symbolic regression for physics discovery~\cite{ahmadi2024ai,de2023ai}, its uncertainty quantification capabilities have been largely overlooked in the literature. 
This study aims to address this gap by investigating physics-informed estimation under total uncertainty in the context of numerical models with applications in biology and physiology.

Computational models in biology have a rich history, encompassing diverse areas such as epidemiology~\cite{chowell2017fitting}, species and population dynamics~\cite{hastings2013population}, gene regulatory networks~\cite{smolen2000mathematical}, metabolic pathways~\cite{wang2017review}, and phylogenetics~\cite{stamatakis2006phylogenetic}, among many others.
%
Similarly, the field of physiology has seen a constant growth in the complexity of model formulations following early studies by Harvey~\cite{harvey2014works}, Poiseuille~\cite{sutera1993history}, Frank~\cite{frank1990basic}, and many others. Two- or three-element Windkessel models represent basic examples of lumped parameter hemodynamic models.  These models approximate the Navier-Stokes equations in cylindrical coordinates, linearized around rest conditions~\cite{milivsic2004analysis}, and are analogous to equations describing current and voltage in electrical circuits.
More complex one-dimensional models~\cite{hughes1973one} offer a more accurate hemodynamic representation but still rely on approximations for minor losses related, e.g., to stenosis or bifurcations. Additionally, hemodynamics can be solved using complex multi-physics three-dimensional models with fluid-structure interaction~\cite{bazilevs2013computational,brown2023patient} or include cardiac electrophysiology~\cite{verzicco2022electro,zingaro2024electromechanics}.
In addition, a number of recent studies have focused on the quantification of uncertainty in the predictions from hemodynamic models due to variability in boundary conditions, material properties of vascular tissue~\cite{sank2011sc,schiavazzi2017generalized, seo2020effects, tran2019uncertainty} or anatomical model geometry~\cite{maher2021geometric}. More recent approaches have also focused on the solution of inverse problems, combined forward and inverse problems~\cite{harrod2021predictive, tran2016automated, schiavazzi2016patient, schiavazzi2016uncertainty}, multi-fidelity propagation and sensitivity analysis~\cite{fleeter2020multilevel, seo2020multifidelity, pfaller2022automated, zanoni2024improved, schaferglobal}, and probabilistic neural twins~\cite{lee2024probabilistic}.

The main contribution of this study is to propose a new X-TFC-based method for uncertainty quantification and to improve understanding of the interaction between total uncertainty and physics-informed regularization under a variable amount of data. We do so by focusing on time-dependent problems formulated as systems of ODEs.
We begin with a controlled computational  experiment to examine the behavior of X-TFC-based estimators under separate aleatoric, epistemic, and model-form uncertainties, comparing their performance to PINN and Bayesian PINN (B-PINN)estimators.
Next, we conduct an ablation study for  CVSim-6, a stiff differential ODE model, focusing on aleatoric and epistemic uncertainties. We assess prediction variability by systematically reducing the available data for six compartmental pressures while estimating parameters that characterize pulmonary venous resistance and aortic compliance.
We then consider the common situation arising in lumped-parameter hemodynamics models, in which a compartment with linear resistance is used as a simplified model of a complex vascular tree. To study the impact of this model simplification on the total uncertainty, we introduce a discrepancy function. Furthermore, we demonstrate how different modeling choices for such discrepancy directly influence the quantification of model-form uncertainty.

To the best of our knowledge, this is the first study in which the behavior of physics-informed neural estimators is investigated under total uncertainty for stiff differential systems in the context of lumped parameter hemodynamics.
This paper is organized as follows. Section~\ref{sec:cvsim6_model} introduces the CVSim-6 cardiovascular model, provides the formulation as a differential system, and discusses its stiffness. Section~\ref{sec:method} introduces the X-TFC methodology for gray-box identification, parameter estimation, and how it is used for uncertainty quantification. 
Section~\ref{sec:pedagogical} presents an introductory example to facilitate the reader's understanding of total uncertainty decomposition and to better explain the individual contributions of each uncertainty source.
Finally, the results for the uncertainty quantification for the CVSim-6 cardiovascular model are presented in Section~\ref{sec:results}, followed by a discussion and the conclusions in Section~\ref{sec:conclusions}.

\section{The CVSim-6 cardiovascular model}\label{sec:cvsim6_model}

\begin{figure}
\begin{subfigure}[c]{0.63\textwidth}
\centering
\includegraphics[width=\textwidth]{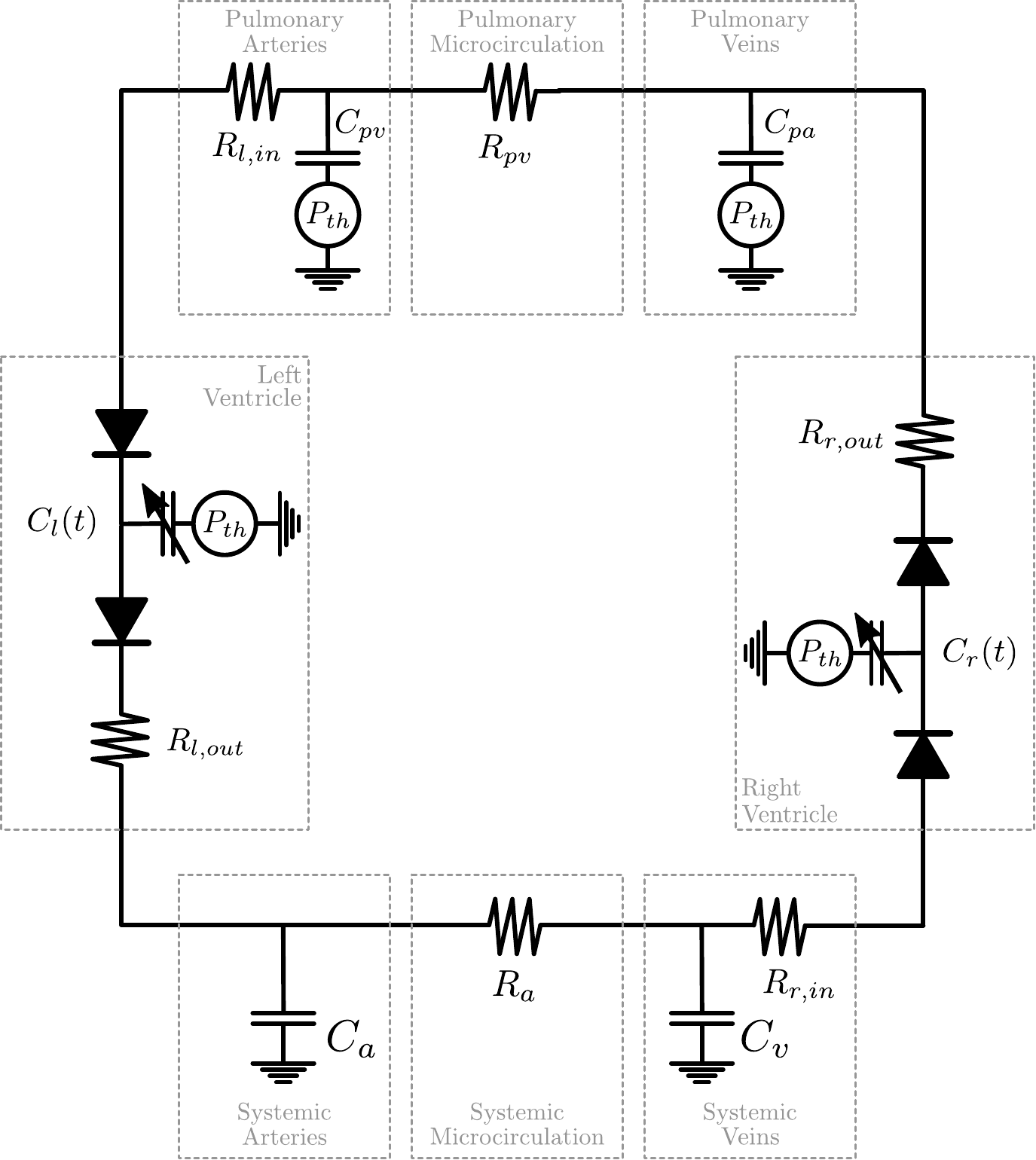}
\caption{Schematic diagram of the CVSim-6 circuit model layout.}
\label{fig:cvsim_circuit}
\end{subfigure}
\begin{subfigure}[c]{0.37\textwidth}
\centering
\includegraphics[width=\textwidth]{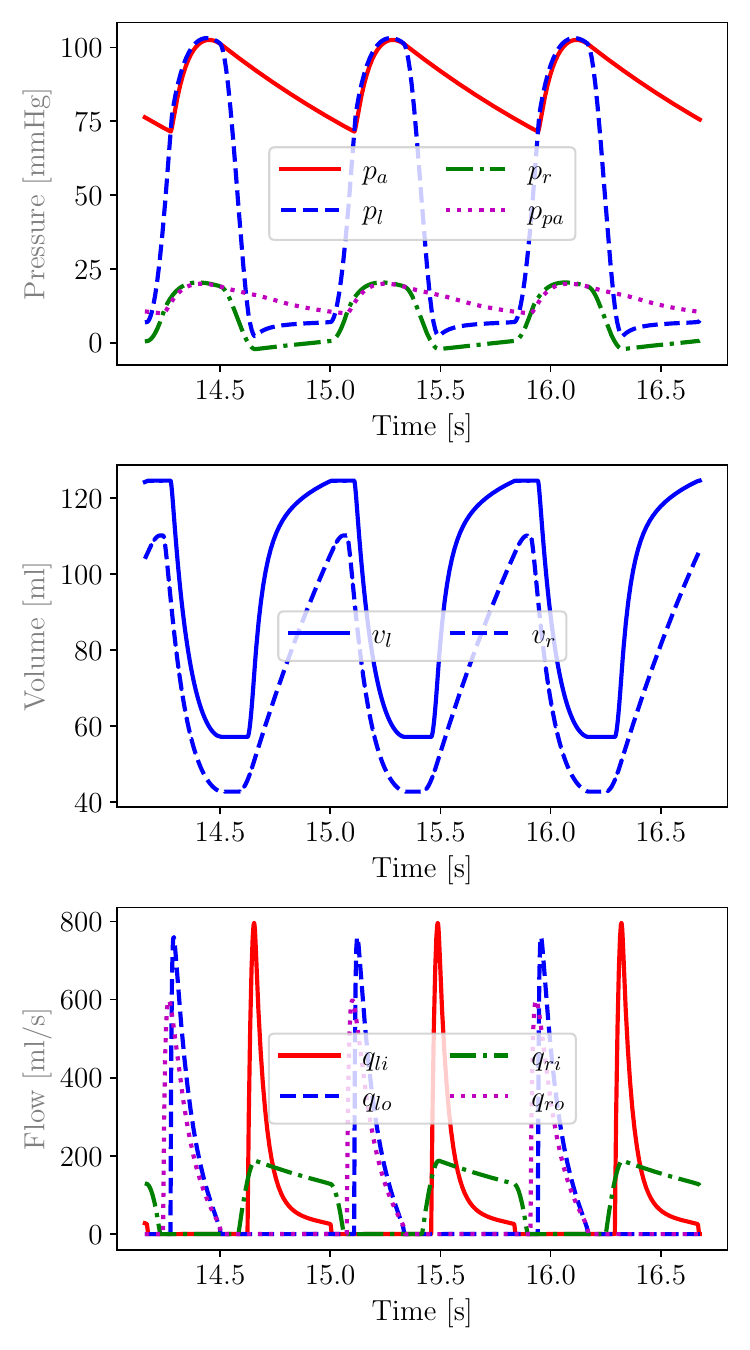}
\caption{CVSim-6 model solution for pressures (top), volumes (middle) and flows (bottom), corresponding to the default parameter set in Tables~\ref{table:params_basic}, \ref{table:params_cap}, \ref{table:params_res} and \ref{table:params_vols}.}
\label{fig:cvsim_def_sol}
\end{subfigure}
\caption{CVSim-6 model circuit and default output.}\label{fig:pulm_res}
\end{figure}

The CVSim-6 is a lumped-parameter hemodynamic model, originally developed for  teaching cardiovascular physiology~\cite{davis1991teaching,heldt2010cvsim}. 
It includes six compartments, where the subscripts $l$, $r$, $a$, $v$, $pa$, and $pv$ indicate quantities referred to the left ventricle, right ventricle, systemic arteries, systemic vein, pulmonary arteries, and pulmonary veins, respectively.
It consists of six differential equations (one per compartment) and 23 input parameters. These parameters, together with a specific combination (referred to as the \emph{default} set) providing outputs associated with the physiology of a healthy subject, are grouped in Tables~\ref{table:params_basic}, \ref{table:params_cap}, \ref{table:params_res} and \ref{table:params_vols}.
A schematic of the CVSim-6 model is also shown in Figure~\ref{fig:cvsim_circuit}.

The CVSim-6 model consists of a system of six ODEs, one per compartment, expressed as
\begin{equation}\label{equ:cvsim6-ode}
\begin{gathered}
\dot{P}_l(t) = \displaystyle  \frac{Q_{l,in}(t) - Q_{l,out}(t) - \Big(P_l(t) - P_{th}\Big)\dot{C_l}(t)}{C_l(t)},\quad
\dot{P}_a(t) = \displaystyle \frac{ Q_{l,out}(t) - Q_a(t) }{C_a}\\
\dot{P}_v(t) = \displaystyle \frac{Q_a(t) - Q_{r,in}(t)}{C_v},\quad
\dot{P}_r(t) = \displaystyle \frac{ Q_{r,in}(t) - Q_{r,out}(t) - \Big(P_r(t) - P_{th}\Big)\dot{C}_r(t) }{ C_r(t)}  \\
\dot{P}_{pa}(t) = \displaystyle \frac{Q_{r,out}(t) - Q_{pv}(t)}{C_{pa}},\quad
\dot{P}_{pv}(t) = \displaystyle \frac{Q_{pv}(t) - Q_{l,in}(t)}{C_{pv}}.
\end{gathered}
\end{equation} 
Volumetric flows are defined via Ohm's law under a uni-directional valve assumption as
\begin{equation}\label{equ:cvsim6-flow}
    \begin{gathered}
        Q_{l,in}(t) = \displaystyle \frac{P_{pv}(t) - P_l(t)}{R_{l,in}}\mathbb{I}_{P_{pv}(t) > P_l(t)},\quad
        Q_{l,out}(t) = \displaystyle \frac{P_{l}(t) - P_a(t)}{R_{l,out}}\mathbb{I}_{P_{l}(t) > P_a(t)} \\
        Q_a(t) = \displaystyle \frac{P_a(t) - P_v(t)}{R_a},\quad
        Q_{r,in}(t) = \displaystyle \frac{P_{v}(t) - P_r(t)}{R_{r,in}}\mathbb{I}_{P_{v}(t) > P_r(t)}  \\
        Q_{r,out}(t) = \displaystyle \frac{P_{r}(t) - P_{pa}(t)}{R_{r,out}}\mathbb{I}_{P_{r}(t) > P_{pa}(t)},\quad
        Q_{pv}(t) = \displaystyle \frac{P_{pa}(t) - P_{pv}(t)}{R_{pv}}.
    \end{gathered}
\end{equation}
Finally, stressed volumes for each compartment are calculated via a linear pressure-volume relationship of the form
\begin{equation}\label{equ:cvsim6-volume}
\begin{gathered}
        V_{l}(t) = V_{l}^{0} + \big(P_l(t) - P_{th} \big) C_l(t),\quad
        V_{a}(t) = V_{a}^{0} + \big(P_a(t) - \frac{1}{3}P_{th}\big) C_a,\\
        V_{v}(t) = V_{v}^{0} + P_v(t) C_v,\quad
        V_{r}(t) = V_{r}^{0} + \big(P_r(t) - P_{th} \big) C_r(t),\\
        V_{pa}(t) = V_{pa}^{0} + \big(P_{pa}(t) - P_{th} \big) C_{pa},\quad
        V_{pv}(t) = V_{pv}^{0} + \big(P_{pv}(t) - P_{th}\big) C_{pv} \ . 
\end{gathered}
\end{equation}
The evolution over a few cardiac cycles for the pressures, volumes, and flows associated with the default parameter set is shown in Figure~\ref{fig:cvsim_def_sol}. The values of the basic physiological quantities, capacitances, resistances, and unstressed volumes used in this work are listed in Appendix~\ref{appendix_a}.

The CVSim-6 model has two sources of nonlinearity: the unidirectional valves and the time-varying left and right ventricular capacitance, which are responsible for ventricular contraction. 
Both mechanisms induce stiffness in the differential system, particularly during systole, when the opening of the aortic valve couples the left ventricular and systemic compartments, resulting in a particularly short relaxation time (as expressed by the equivalent $RC$ constant). 
The key to achieving a correct periodic response is to carefully use implicit solvers and adaptive time-stepping.

\section{Method}\label{sec:method}

\subsection{eXtreme Theory of Functional Connections}\label{sec:xtfc}

The Theory of Functional Connections (TFC) \cite{mortari2017theory} provides the so-called constrained expression (CE) to approximate the solution of the differential equation in a form depending on the problem constraints \cite{leake2020multivariate,de2021theory,mai2022theory}. 
Consider an initial value problem governing the evolution of the scalar quantity $x\in\mathbb{R}$ of the form
\[
\begin{cases}
dx/dt &= f(x, t)\\
x(0) &= x_{0} 
\end{cases},
\]
where the unknown solution is approximated by the constrained expression \cite{mortari2017theory}
\[
x(t, \bm{\beta}) = g(t, \bm{\beta}) - g(0, \bm{\beta}) + x_{0},
\]
with a user-selected function $g(t, \bm{\beta})$. 
According to the X-TFC framework \cite{schiassi2021extreme}, the function $g(t, \bm{\beta})$ belongs to the family of single-layer random projection neural networks, with input weights and biases assigned randomly before training. It is expressed as 
\begin{equation}\label{equ:weights_biases}
g(t, \bm{\beta}) = \sum_{j=1}^{L}\,\beta_{j}\sigma(w_{j}\,t + b_{j}) = \
\begin{bmatrix}
\sigma(w_{1}\,t + b_{1})\\
\sigma(w_{2}\,t + b_{2})\\
\vdots\\
\sigma(w_{L}\,t + b_{L})\\
\end{bmatrix}\,\bm{\beta} = 
\bm{\sigma}^{T}\,\bm{\beta},\,\,\text{and}\,\,\bm{\sigma}^{T}(0) = 
\begin{bmatrix}
\sigma(0)\\
\sigma(0)\\
\vdots\\
\sigma(0)\\
\end{bmatrix} = \bm{\sigma}^{T}_{0},
\end{equation}
where $w_{j}$, $b_{j}$ and $\beta_{j}, j = 1,\dots,L$ represent the weight, bias, and output weight associated with the $j$-th neuron of the single hidden layer available to the network. 
Nonlinearity is implemented through a user-selected activation function $\sigma(w\,t + b)$ (in this work, $tanh$ or \emph{softplus} activation functions are used). 
A parametric approximation for the solution of the original ODE \cite{mortari2017theory} and its time derivative can thus be written as
\begin{align}
x(t, \bm{\beta}) &= \left[\bm{\sigma} - \bm{\sigma}_{0}\right]^{T}\,\bm{\beta} + x_0,\label{eq:ce}\\
\dot{x}(t, \bm{\beta}) &= \dot{\bm{\sigma}}^{T}\,\bm{\beta}. \label{eq:ce_der}
\end{align}

Black- or gray-box approximation problems are then formulated as determining the value of the coefficients $\beta$ from observations of the unknown solution and/or prior knowledge of the physics (i.e., differential equation).
This problem is formulated on a number of sub-domains obtained by defining $n$ sub-intervals of equal length $h = t_k - t_{k-1}$, for $k = 1,\dots, n$, leading to a collection of initial value problems
\begin{equation}\label{equ:ode_subintervals}
\begin{cases}
dx^{(k)}/dt &= f(x^{(k)},t),\,\,\text{for}\,\,x^{(k)}\in[t_k,t_{k+1}],\\
x_{0}^{(k)} &= x_{f}^{(k-1)}.
\end{cases}
\end{equation}
where continuity of the solution on successive intervals is imposed through the boundary conditions.

\begin{figure}
    \centering
    \includegraphics[width=\textwidth]{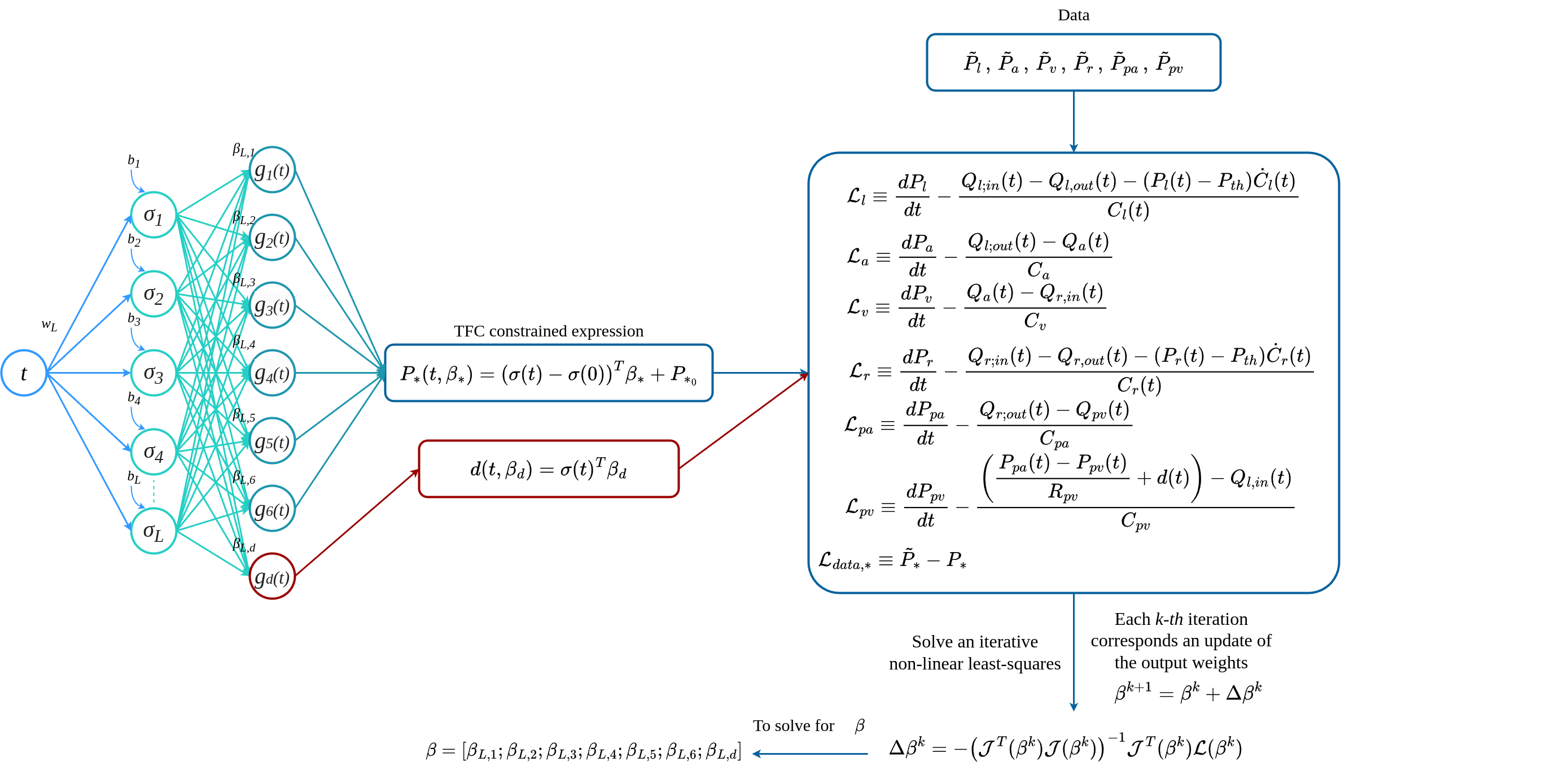}
    \caption{Schematic of the X-TFC algorithm for performing gray-box identification of the pulmonary flux discrepancy. Input weights and biases are randomly selected. The last step solves iteratively a least-squares problem.}
    \label{fig:gbx-tfc}
\end{figure}

We now focus on black-box system identification for a system of $m$ differential equations with $m$ unknown of the form 
\begin{equation}\label{equ:ode_m}
\begin{cases}
\dot{x}_{1} &= f_{1}(x_{1}, x_{2}, \dots, x_{m})\\
\dot{x}_{2} &= f_{2}(x_{1}, x_{2},\dots, x_{m})\\
\vdots\\
\dot{x}_{m} &= f_{m}(x_{1}, x_{2},\dots, x_{m}),
\end{cases}
\end{equation}
in which the right-hand-side functions $f_{1},f_{2},\dots,f_{m}$, are completely unknown. 
The first step is to build the CEs for each state variable under consideration. For this example, we have 
\begin{equation}\label{eq:CEs}
    \begin{cases}
    x_{1}(t) = \left(\bm{\sigma} - \bm{\sigma}_{0}\right)^{T}\,\bm{\beta}_{1} + x_{1}(0)\\
    x_{2}(t) = \left(\bm{\sigma} - \bm{\sigma}_{0}\right)^{T}\,\bm{\beta}_{2} + x_{2}(0)\\
    \vdots\\
    x_{m}(t) = \left(\bm{\sigma} - \bm{\sigma}_{0}\right)^{T}\,\bm{\beta}_{m} + x_{m}(0)
    \end{cases}\,\,
    \text{and their derivatives}\,\,
    \begin{cases}
    \dot{x}_{1}(t) = c\,\dot{\bm{\sigma}}^{T}\,\bm{\beta}_{1}\\
    \dot{x}_{2}(t) = c\,\dot{\bm{\sigma}}^{T}\,\bm{\beta}_{2}\\
    \vdots\\
    \dot{x}_{m}(t) = c\,\dot{\bm{\sigma}}^{T}\,\bm{\beta}_{m},
    \end{cases}
\end{equation}
where $c$ is a scaling factor mapping the time domain $t\in [t_0,t_f]$ (with $t_0 = 0$ and $t_f=T$) into the activation function domain $z \in [z_0, z_f]$, and it is defined as $c = \dfrac{z_f - z_0}{t_f - t_0}$. 
The loss functions we want to minimize are the differences between the observed dynamics $(\widetilde{x}_{1,1},\widetilde{x}_{2,1},\dots,\widetilde{x}_{m,1})$,
$(\widetilde{x}_{1,2},\widetilde{x}_{2,2},\dots,\widetilde{x}_{m,2})$ up to 
$(\widetilde{x}_{1,p},\widetilde{x}_{2,p},\dots,\widetilde{x}_{m,p})$
and their CEs approximations at $p$ time instants $t_{k} < t_{1} < t_{2} < \dots < t_{p} < t_{k+1}$
\begin{equation}\label{equ:loss_data}
\begin{cases}
\mathcal{L}_{\text{data},1}(t_{1}) &= \widetilde{x}_{1,1} - x_{1}(t_{1}),\,\mathcal{L}_{\text{data},1}(t_{2}) = \widetilde{x}_{1,2} - x_{1}(t_{2}),\,\dots\,\mathcal{L}_{\text{data},1}(t_{p}) = \widetilde{x}_{1,p} - x_{1}(t_{p})\\
\mathcal{L}_{\text{data},2}(t_{1}) &= \widetilde{x}_{2,1} - x_{2}(t_{1}),\,\mathcal{L}_{\text{data},2}(t_{2}) = \widetilde{x}_{2,2} - x_{2}(t_{2}),\,\dots\,\mathcal{L}_{\text{data},2}(t_{p}) = \widetilde{x}_{2,p} - x_{2}(t_{p})\\
\vdots\\
\mathcal{L}_{\text{data},m}(t_{1}) &= \widetilde{x}_{m,1} - x_{m}(t_{1}),\,\mathcal{L}_{\text{data},m}(t_{2}) = \widetilde{x}_{m,2} - x_{m}(t_{2}),\,\dots\,\mathcal{L}_{\text{data},m}(t_{p}) = \widetilde{x}_{m,p} - x_{m}(t_{p}).\\
\end{cases}
\end{equation}
The next step is to express the loss as a linear function of the coefficients $\bm{\beta}$ through the Jacobian matrix from a Taylor expansion of the form
\[
\mathcal{L}_{\text{data},i}(t_{j},\bm{\beta}^{k}_{i} + \bm{\Delta\beta}_{i}^{k}) = \mathcal{L}_{\text{data},i}(t_{j},\bm{\beta}^{k}_{i}) +  \bm{\Delta\beta}_{i}^{k}\,\frac{\partial \mathcal{L}_{\text{data},i}(t_{j},\bm{\beta}^{k}_{i})}{\partial \bm{\beta}^{k}_{i}} + o\left[\left(\bm{\Delta\beta}_{i}^{k}\right)^2\right].
\]
Since we would like to achieve a zero loss at the next iteration, we can write
\[
\mathcal{L}_{\text{data},i}(t_{j},\bm{\beta}^{k}_{i}) -  \bm{\Delta\beta}_{i}^{k}\,\left(\bm{\sigma}(t_{j}) - \bm{\sigma}(t_{k})\right)^{T} = 0.
\]
This leads to a Newton-type iteration of the form 
\[
\bm{\beta}_{i}^{k+1} = \bm{\beta}_{i}^{k} + \bm{\Delta\beta}_{i}^{k},\,\,\text{where $\bm{\Delta\beta}_{i}^{k}$ is the solution of}\,\,\bm{\mathcal{J}}_{i}\,\bm{\Delta\beta}_{i}^{k} = \bm{\mathcal{L}}_{i}.
\]
When $p$ observations are available in $[t_{k},t_{k+1}]$ for the $i$-th variable, a left-hand-side matrix $\bm{\mathcal{L}}$ is obtained by stacking vertically the gradient contributions, i.e.
\[
\bm{\mathcal{J}}\in\mathbb{R}^{p\times L} = 
\begin{bmatrix}
\bm{\sigma}(t_{1}) - \bm{\sigma}(t_{k})\\
\bm{\sigma}(t_{2}) - \bm{\sigma}(t_{k})\\
\vdots\\
\bm{\sigma}(t_{p}) - \bm{\sigma}(t_{k})
\end{bmatrix},\,\,\text{and similarly}\,\,\bm{\mathcal{L}} \in\mathbb{R}^{p}
\begin{bmatrix}
\mathcal{L}_{1}\\
\mathcal{L}_{2}\\
\vdots\\
\mathcal{L}_{p}
\end{bmatrix},
\]
leading to a linear system of equations with $p$ equations and $L$ unknowns. For cases where $p>L$, the resulting over-determined system can be solved by least-squares to determine the $k$-th iterate of the coefficient vector $\bm{\beta}$
\begin{equation}\label{eq:beta_comp}
    \bm{\Delta\beta}^{k} = -\left[\bm{\mathcal{J}}^{T}\,\bm{\mathcal{J}}\right]^{-1}\,\bm{\mathcal{J}}^{T}\,\bm{\mathcal{L}}.
\end{equation}
%
A representative schematic of the gray-box X-TFC algorithm is shown in Figure \ref{fig:gbx-tfc}, displaying its main steps.

The above framework is based on a loss function that only encodes model solution errors.
%
In practice, the residuals of the differential system~\eqref{equ:ode_m} are  directly added to the loss~\eqref{equ:loss_data} to give
\[
\begin{gathered}
\mathcal{L}_{x_{1}} = f_{1}(x_{1},x_{2},\dots,x_{m}) - \dot{x}_{1},\,\mathcal{L}_{x_{2}} = f_{2}(x_{1},x_{2}\dots,x_{m}) - \dot{x}_{2},\,\cdots,\,\mathcal{L}_{x_{m}} = f_{m}(x_{1},x_{2},\dots,x_{m}) - \dot{x}_{m},\\
\mathcal{L}_{\text{data},x_{1}} = \widetilde{x}_{1} - x_{1},\,\mathcal{L}_{\text{data},x_{2}} = \widetilde{x}_{2} - x_{2},\,\cdots,\,\mathcal{L}_{\text{data},x_{m}} = \widetilde{x}_{m} - x_{m}.
\end{gathered}
\]
where the functions $f_{1},f_{2}\dots,f_{m}$ are approximated with a small neural network
\[
\begin{cases}
f_{1}(x_{1}, x_{2},\dots,x_{m}) = c\,\bm{\sigma}^{T}\,\widetilde{\bm{\beta}}_{1}\\
f_{2}(x_{1}, x_{2},\dots,x_{m}) = c\,\bm{\sigma}^{T}\,\widetilde{\bm{\beta}}_{2}\\
\vdots\\
f_{m}(x_{1},x_{2},\dots,x_{m}) = c\,\bm{\sigma}^{T}\,\widetilde{\bm{\beta}}_{m}
\end{cases},
\]
where the constant $c$ is unique on subdomains of equal length.
In such a case, the Jacobian matrix becomes
\[
\bm{\mathcal{J}}_{i} = 
\begin{bmatrix}
c\,\bm{\sigma}(t_{i}) & 0 & 0 & -c\,\dot{\bm{\sigma}}(t_{i}) & 0 & 0\\
0 & c\,\bm{\sigma}(t_{i}) & 0 & 0 & -c\,\dot{\bm{\sigma}}(t_{i}) & 0\\
0 & 0 & c\,\bm{\sigma}(t_{i}) & 0 & 0 & -c\,\dot{\bm{\sigma}}(t_{i})\\
\bm{\sigma}(t_{k})-\bm{\sigma}(t_{i}) & 0 & 0 & 0 & 0 & 0\\
0 & \bm{\sigma}(t_{k})-\bm{\sigma}(t_{i}) & 0 & 0 & 0 & 0\\
0 & 0 & \bm{\sigma}(t_{k})-\bm{\sigma}(t_{i}) & 0 & 0 & 0
\end{bmatrix},
\]
and the unknown vector $\bm{\beta} = [\widetilde{\bm{\beta}}_{1},\dots,\widetilde{\bm{\beta}}_{m}, \bm{\beta}_{1}, \dots, \bm{\beta}_{m}]^{T}$ is computed by iteratively solving the linear system 
\[
\bm{\mathcal{J}}\,\bm{\Delta\beta}^{k} = \bm{\mathcal{L}},\,\,\text{where}\,\,
\bm{\mathcal{J}} = 
\begin{bmatrix}
\bm{\mathcal{J}}_{1}\\
\bm{\mathcal{J}}_{2}\\
\vdots\\
\bm{\mathcal{J}}_{p}\\
\end{bmatrix},\,\,\text{and}\,\,
\bm{\mathcal{L}} = 
\begin{bmatrix}
\bm{\mathcal{L}}_{1}\\
\bm{\mathcal{L}}_{2}\\
\vdots\\
\bm{\mathcal{L}}_{p}\\
\end{bmatrix},
\]
with 
\[
\bm{\mathcal{L}}_{i} = 
\begin{bmatrix}
\mathcal{L}_{x_1}(t_{i}) &
\mathcal{L}_{x_2}(t_{i}) &
\cdots &
\mathcal{L}_{x_{m}}(t_{i}) &
\mathcal{L}_{\text{data},x_1}(t_{i}) &
\mathcal{L}_{\text{data},x_2}(t_{i}) &
\cdots &
\mathcal{L}_{\text{data},x_{m}}(t_{i})
\end{bmatrix}^{T}.
\]
Thus, by substituting them into the CEs, CE derivatives, and unknown functions of eqs. (7), (8), and (13), we obtain an approximation for the learned dynamics $[x_{1}(t), x_{2}(t),\dots, x_{m}(t)]^{T}$, their variations in time $[\dot{x}_{1},\dot{x}_{2},\dots,\dot{x}_{m}]^{T}$, and the terms $f_{1}, f_{2},\dots,f_{m}$. 

\begin{remark}
    In this work, we decompose the overall domain into several sub-domains and solve a local least squares problem on each sub-domain. Additionally, X-TFC enforces a $\mathcal{C}^{0}$ continuity condition for the solution $x(t)$ at the boundaries of each sub-domain, consistent with the formulation of CVSim-6 as a first-order initial value problem. The sub-domains are small for the stiff ODEs we consider here so any discontinuity in the slope at the interfaces is negligible.
\end{remark}

\subsection{X-TFC formulation for the CVSim-6 differential system}

This section presents the X-TFC formulation used for parameter estimation in the CVSim-6 ODE system. The parameters to be estimated generate additional unknowns in each least squares solution. 
For improved clarity, here we present a step-by-step example based on \eqref{equ:cvsim6-ode}, with unknown parameters $R_{pv}$ and $C_a$, and data only observed for $\widetilde P_a$ and $\widetilde P_{pa}$.
We first write the constrained expressions and their time derivatives from \eqref{eq:ce} and \eqref{eq:ce_der}
\begin{equation}\label{eq:pressure_ce}
\begin{gathered}
    P_l = (\boldsymbol{\sigma} - \boldsymbol{\sigma}_0) \boldsymbol{\beta}_l + P_{l_0},\,\,\dot P_l = c \boldsymbol{\dot \sigma} \boldsymbol{\beta}_l,\,\,P_a = (\boldsymbol{\sigma} - \boldsymbol{\sigma}_0) \boldsymbol{\beta}_a + P_{a_0},\,\,\dot P_a = c \boldsymbol{\dot \sigma} \boldsymbol{\beta}_a,\\    
    P_v = (\boldsymbol{\sigma} - \boldsymbol{\sigma}_0) \boldsymbol{\beta}_v + P_{v_0},\,\,\dot P_v = c \boldsymbol{\dot \sigma} \boldsymbol{\beta}_v,\,\,P_r = (\boldsymbol{\sigma} - \boldsymbol{\sigma}_0) \boldsymbol{\beta}_r + P_{r_0},\,\,\dot P_r = c \boldsymbol{\dot \sigma} \boldsymbol{\beta}_r,\\
    P_{pa} = (\boldsymbol{\sigma} - \boldsymbol{\sigma}_0) \boldsymbol{\beta}_{pa} + P_{pa_0},\,\,\dot P_{pa} = c \boldsymbol{\dot \sigma} \boldsymbol{\beta}_{pa},\,\,P_{pv} = (\boldsymbol{\sigma} - \boldsymbol{\sigma}_0) \boldsymbol{\beta}_{pv} + P_{pv_0},\,\,\dot P_{pv} = c \boldsymbol{\dot \sigma} \boldsymbol{\beta}_{pv}. 
\end{gathered}
\end{equation}
We can now assemble the loss function from the residuals of the 6 ODEs and 2 observed pressures $(\widetilde{P_a}, \widetilde P_{pa})$, as follows
\begin{equation}
\begin{gathered}
\boldsymbol{\mathcal{L}}_l \equiv \dot{P}_l - \displaystyle  \frac{Q_{l,in} - Q_{l,out} - \Big(P_l - P_{th}\Big)\dot{C_l}(t)}{C_l(t)},\quad\boldsymbol{\mathcal{L}}_a \equiv \dot{P}_a - \displaystyle \frac{ Q_{l,out} - Q_a }{C_a},\quad\boldsymbol{\mathcal{L}}_v \equiv \dot{P}_v - \displaystyle \frac{Q_a - Q_{r,in}(t)}{C_v} \\
\boldsymbol{\mathcal{L}}_r \equiv \dot{P}_r - \displaystyle \frac{ Q_{r,in} - Q_{r,out} - \Big(P_r - P_{th}\Big)\dot{C}_r }{ C_r(t)},\quad\boldsymbol{\mathcal{L}}_{pa} \equiv \dot{P}_{pa} - \displaystyle \frac{Q_{r,out} - Q_{pv}}{C_{pa}},\quad\boldsymbol{\mathcal{L}}_{pv} \equiv \dot{P}_{pv} - \displaystyle \frac{Q_{pv} - Q_{l,in}}{C_{pv}} \\
\boldsymbol{\mathcal{L}}_{a_{data}} \equiv \tilde P_a - P_a,\quad\boldsymbol{\mathcal{L}}_{pa_{data}} \equiv  \tilde P_{pa} - P_{pa} \\
\end{gathered}
\end{equation}
such that
\begin{equation}
    \boldsymbol{\mathcal{L}} = [\boldsymbol{\mathcal{L}}_l\,\, \boldsymbol{\mathcal{L}}_a\,\, \boldsymbol{\mathcal{L}}_v\,\, \boldsymbol{\mathcal{L}}_r\,\, \boldsymbol{\mathcal{L}}_{pa}\,\, \boldsymbol{\mathcal{L}}_{pv} ]^{T}.
\end{equation}
The unknown vector $\boldsymbol{\beta}$ is composed of the unknown output weights of the neural network and the parameters to estimate, such as
\begin{equation}
    \boldsymbol{\beta} = [\boldsymbol{\beta}_l\,\,\boldsymbol{\beta}_a\,\,\boldsymbol{\beta}_v\,\,\boldsymbol{\beta}_r\,\,\boldsymbol{\beta}_{pa}\,\,\boldsymbol{\beta}_{pv}\,\,R_{pv}\,\,C_a]^{T}.
\end{equation}
By computing the derivatives of the loss functions with respect to the unknowns, we get the Jacobian matrix
\begin{equation}
    \boldsymbol{\mathcal{J}} = \begin{bmatrix}  \dfrac{\boldsymbol{\mathcal{L}}_l}{\boldsymbol{\beta}_l} &  \dfrac{\boldsymbol{\mathcal{L}}_l}{\boldsymbol{\beta}_a}    &     \bf{0}                & \bf{0}        &        \bf{0}   &       \dfrac{\boldsymbol{\mathcal{L}}_l}{\boldsymbol{\beta}_{pv}}     &      0    &     0    \\
      \dfrac{\boldsymbol{\mathcal{L}}_a}{\boldsymbol{\beta}_l} &   \dfrac{\boldsymbol{\mathcal{L}}_a}{\boldsymbol{\beta}_a} &   \dfrac{\boldsymbol{\mathcal{L}}_a}{\boldsymbol{\beta}_v}     &     \bf{0}       &         \bf{0}        &          \bf{0}      &           0         & \dfrac{\boldsymbol{\mathcal{L}}_a}{C_a}  \\
       \bf{0}     &     \dfrac{\boldsymbol{\mathcal{L}}_v}{\boldsymbol{\beta}_a}  &  \dfrac{\boldsymbol{\mathcal{L}}_v}{\boldsymbol{\beta}_v}   &  \dfrac{\boldsymbol{\mathcal{L}}_v}{\boldsymbol{\beta}_r}   &       \bf{0}        &          \bf{0}       &          0          &       0        \\
      \bf{0}          &       \bf{0}     &    \dfrac{\boldsymbol{\mathcal{L}}_r}{\boldsymbol{\beta}_v}  &  \dfrac{\boldsymbol{\mathcal{L}}_r}{\boldsymbol{\beta}_r}  &  \dfrac{\boldsymbol{\mathcal{L}}_r}{\boldsymbol{\beta}_{pa}}  &         \bf{0}        &         0     &            0        \\
 \bf{0}        &         \bf{0}    &            \bf{0}   &      \dfrac{\boldsymbol{\mathcal{L}}_{pa}}{\boldsymbol{\beta}_r} &   \dfrac{\boldsymbol{\mathcal{L}}_{pa}}{\boldsymbol{\beta}_{pa}}  &    \dfrac{\boldsymbol{\mathcal{L}}_{pa}}{\boldsymbol{\beta}_{pv}}  &    \dfrac{\boldsymbol{\mathcal{L}}_{pa}}{R_{pv}}   &      0        \\
 \dfrac{\boldsymbol{\mathcal{L}}_{pv}}{\boldsymbol{\beta}_{l}}     &    \bf{0}     &           \bf{0}     &     \bf{0}    &  \dfrac{\boldsymbol{\mathcal{L}}_{pv}}{\boldsymbol{\beta}_{pa}} &   \dfrac{\boldsymbol{\mathcal{L}}_{pv}}{\boldsymbol{\beta}_{pv}}  &  \dfrac{\boldsymbol{\mathcal{L}}_{pv}}{R_{pv}}       &   0        \\     
 \bf{0}    &  \dfrac{\boldsymbol{\mathcal{L}}_{a_{data}}}{\boldsymbol{\beta}_a}  &  \bf{0}     &    \bf{0}  &  \bf{0}  &  \bf{0}  &  0  &  0        \\
 \bf{0}  &  \bf{0}  &  \bf{0}  &  \bf{0}  &  \dfrac{\boldsymbol{\mathcal{L}}_{pa_{data}}}{\boldsymbol{\beta}_{pa}}  &  \bf{0}  &  0  &  0        
 \end{bmatrix}                   
\end{equation}
Finally, we can compute the vector of the unknowns using \eqref{eq:beta_comp}, which provides both the estimated parameters $C_a$ and $R_{pv}$, and the pressure profiles from the constrained expressions \eqref{eq:pressure_ce}.

As previously discussed in Section~\ref{sec:xtfc}, we use domain decomposition in time to formulate the estimation problem on small, sequentially ordered, non overlapping subdomains. Thus, X-TFC is iteratively applied to each subdomain, selecting the initial conditions so variables are continuous across subdomain interfaces.
Final parameter estimates are obtained as averages over point values obtained at each subdomain.
Missing terms in the differential equations or any model discrepancies are estimated by adding a new neural network, similar to how the state variables are estimated. 
For more details, interested readers can refer to the gray-box X-TFC formulation in~\cite{de2023ai}.

\subsection{Uncertainty Quantification}\label{sec:uq}

This section formalizes physics-informed state and parameter estimation for dynamical systems under \emph{total uncertainty}, i.e., aleatoric, epistemic, and model-form uncertainty as defined in Section~\ref{sec:intro}. Consider a statistical model of the form
\[
\widetilde{\bm{x}}(t) = \bm{x}(t,\bm{\beta}) + \bm{x}_{0} + \bm{\epsilon}(t),
\]
where $\widetilde{\bm{x}}(t)$ is the underlying true process, $\bm{x}(t,\bm{\beta})$ is a gray-box X-TFC approximation and $\bm{\epsilon}(t)\sim\mathcal{N}(\bm{0},\bm{C})$ is an heteroscedastic noise model with diagonal covariance matrix, where the square root of the diagonal elements are computed as
\begin{equation}\label{equ:noise_obs}
\sigma_{i} = 0.02\cdot \max_{t}(\vert\widetilde{x}_{i}(t)\vert),\,\,\text{for}\,\,i=1,\dots,m,
\end{equation}
and reported in Table~\ref{tab:stds} for all CVSim-6 state variables.
\begin{table}[ht!]
\centering
\caption{Noise standard deviations for pressure data.}\label{tab:stds}
\begin{tabular}{c c c}
\toprule
Variable & Maximum value (mmHg) & $\sigma_{i}$ (mmHg)\\
\midrule
$P_l$ & 103.6 & 2.07\\
$P_a$ & 103.0 & 2.06\\
$P_v$ & 7.3 & 0.14\\
$P_r$ & 20.4 & 0.41\\
$P_{pa}$ & 20.0 & 0.40\\
$P_{pv}$ & 12.1 & 0.24\\
\bottomrule
\end{tabular}

\end{table}
Let us  also assume the quantity $\bm{x}(t,\bm{\beta})$ to be an approximation for the solution of the system of differential equations~\eqref{equ:cvsim6-ode}. These equations provide only an approximation of the true circulatory response of an individual, and differ from the true response by a vector of \emph{model-form} error components expressed as
\[
\begin{cases}
\dot{x}_{1} = f_{1}(x_{1},x_{2},\dots,x_{m}) + h_{1}(x_{1},x_{2},\dots,x_{m},y_{1},y_{2},\dots,y_{n})\\
\dot{x}_{2} = f_{2}(x_{1},x_{2},\dots,x_{m}) + h_{2}(x_{1},x_{2},\dots,x_{m},y_{1},y_{2},\dots,y_{n})\\
\vdots\\
\dot{x}_{m} = f_{m}(x_{1},x_{2},\dots,x_{m}) + h_{m}(x_{1},x_{2},\dots,x_{m},y_{1},y_{2},\dots,y_{n}),
\end{cases}
\,\text{or in compact form}\,\,
\dot{\bm{x}} = \bm{f}(\bm{x}) + \bm{h}(\bm{x},\bm{y}).\\
\]
We first assume that the variables $\bm{x}$ are sufficient to describe the dynamics for the selected quantities of interest, or, in other words, $\bm{h}(\bm{x},\bm{y}) = \bm{h}(\bm{x})$.
This leads to a modified X-TFC approximation with additional coefficients $\bm{\beta}_{h}$ used to approximate the discrepancy term $\bm{h}$, leading to the modified statistical model
\begin{equation}\label{equ:stats_model_tot_uq}
\widetilde{\bm{x}}(t) = \bm{x}(t,\bm{\beta},\bm{\beta}_{h}) + \bm{x}_{0}+ \bm{\epsilon}(t),
\end{equation}
which contains all three uncertainty mechanisms mentioned above. 

The term $\bm{\epsilon}(t)$ in~\eqref{equ:stats_model_tot_uq} accounts for the irreducible \emph{aleatoric} uncertainty, responsible for variability in the output of repeated model evaluations.
By epistemic uncertainty, we refer to the characterization of the variability in the predicted $\bm{x}$ due to changes in the coefficients $\bm{\beta}$, resulting from the random selection of the weights $\bm{w}$ and biases $\bm{b}$ in~\eqref{equ:weights_biases}, and variability in the observed data consistent with the assumed noise model, which also informs the variability in the initial condition $\bm{x}_{0}$ (perturbed using a zero-mean Gaussian noise with standard deviations in~\eqref{equ:noise_obs}).
Additionally, \emph{model-form} uncertainty consists of epistemic uncertainty on the discrepancy coefficients $\bm{\beta}_{h}$, induced by noise in the pressure data and the random selection of the weights $\bm{w}_{h}$ and biases $\bm{b}_{h}$.

To quantify uncertainty in the reconstructed X-TFC response, we use a simple, scalable, yet effective Monte Carlo approach, which we refer to as MC X-TFC.
Multiple instances of X-TFC are trained independently based on synthetic data with added random noise from a known distribution, each with randomly initialized weights and biases.
MC X-TFC shares similarities with \emph{deep ensembles}, as proposed in \cite{lakshminarayanan2017simple}, which have proven to be highly effective for uncertainty quantification in neural networks, even when such uncertainty arises solely due to the random initialization of weights and biases (e.g., Refs. \cite{ganaie2022ensemble, rahaman2021uncertainty, wenzel2020hyperparameter, gawlikowski2023survey, psaros2023uncertainty, zou2024neuraluq, pickering2022discovering, zou2024correcting, zhang2024discovering}). 
In the next section, we use a simple system to compare the performance of X-TFC and PINNs in the quantification of total uncertainty.

\begin{remark}
The examples discussed in the next sections show how aleatoric, epistemic, and model-form uncertainty are not independent. Aleatoric uncertainty is quantified a priori according to a known probability density and superimposed on synthetically generated data. As such, it is only affected by assumptions related to the precision of the measurement devices used to quantify blood pressures, flows, and volumes. However, its volatility directly affects epistemic and model-form uncertainty in a non-linear fashion.
\end{remark}

\section{Introductory example}\label{sec:pedagogical}

As an introductory example, we examine the decomposition of total prediction uncertainty produced by MC X-TFC when applied to the solution of a simple ODE system. 
The same decomposition is also evaluated using physics-informed neural networks, specifically the deep ensemble method for PINNs~\cite{psaros2023uncertainty, zou2024neuraluq} and Bayesian PINNs~\cite{yang2021b}. 
In Bayesian PINNs, a posterior distribution of the network parameters is first formulated by conditioning on both data and model equations. Samples from such posterior are then generated by Hamiltonian Monte Carlo (HMC)~\cite{neal2012mcmc}, and a posterior predictive distribution is finally identified through forward network evaluations.



\subsection{Decomposition of total uncertainty}\label{sec:pedagogical_1}

\begin{figure}[h!]
\centering
    \includegraphics[width=\textwidth]{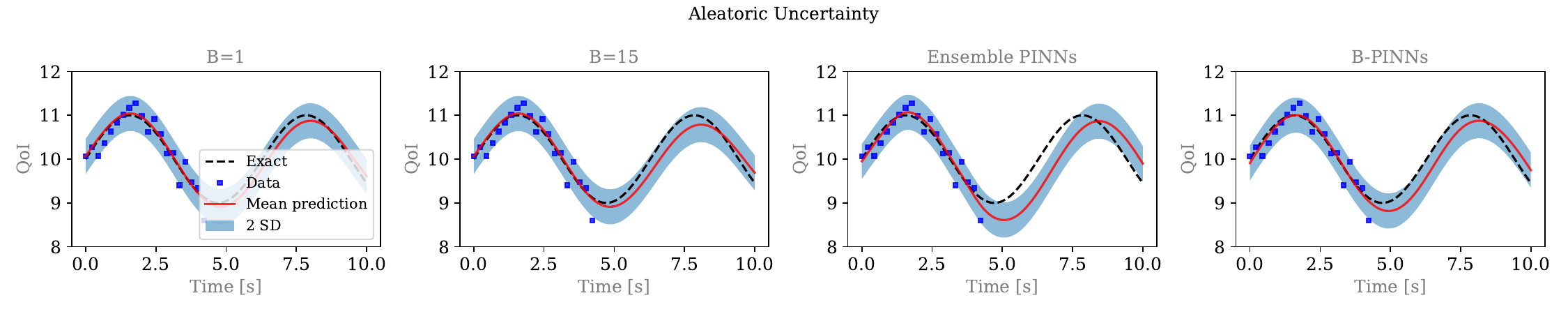}
    \includegraphics[width=\textwidth]{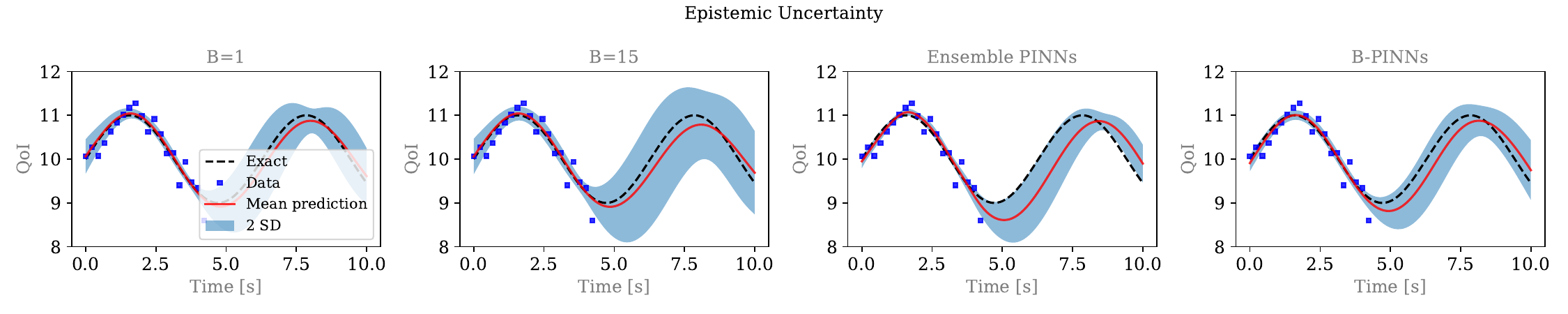}
    \includegraphics[width=\textwidth]{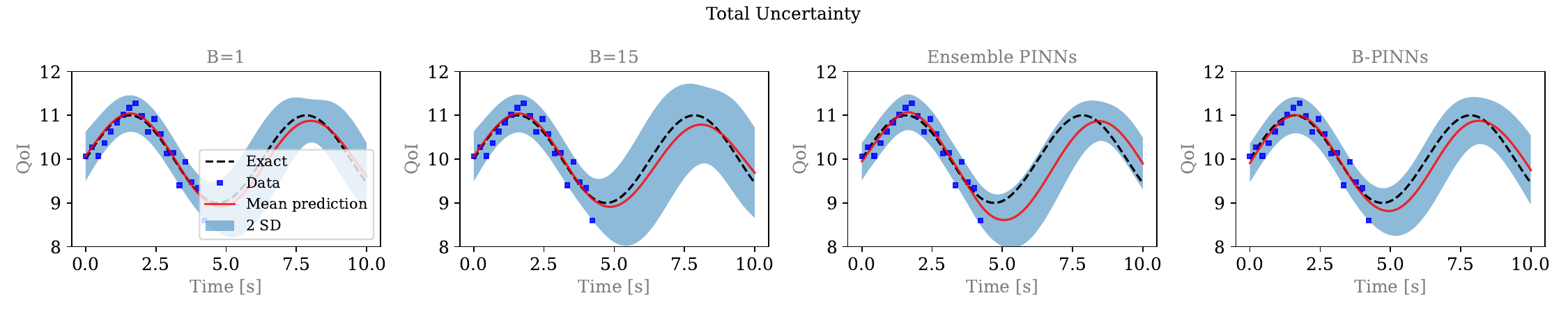}
    \caption{Decomposition of total uncertainty (epistemic and aleatoric) in the reconstruction of a harmonic ODE solution from noisy data using MC X-TFC. $B$ denotes the bound of the uniform distribution $\mathcal{U}[-B, B]$ from which the input weights and biases of the hidden layer are randomly initialized. For comparison, results from ensemble PINN and B-PINN are also reported. The estimated values of $k$ are presented in Table~\ref{tab:example_0}.}
    \label{fig:example_0_0}
\end{figure}

\begin{figure}[h!]
\centering
    \includegraphics[width=\textwidth]{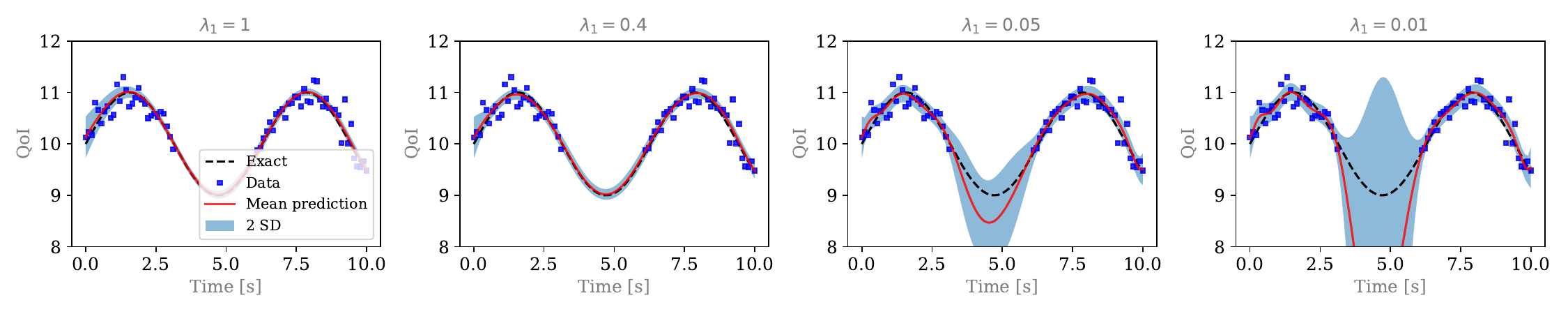}
    \caption{Reconstructed $x(t)$ and epistemic uncertainty computed using MC X-TFC for a varying degree of physics-informed regularization, where $\lambda_1$ in~\eqref{eq:example_0_loss} denotes the penalty coefficient in the loss function. Both the error in the predicted mean and the predicted uncertainty grow significantly as a result of increasingly relying on (missing) data as $\lambda_1$ is reduced.} 
    \label{fig:example_0_1}
\end{figure}


Consider the following initial value problem 
\begin{equation}\label{eq:pedagogical}
    \begin{cases}
    \dfrac{dx(t)}{dt} = \dfrac{\cos(kt)}{k},\\
    x(0) = x_{0} = 10,
    \end{cases}
\end{equation}
with exact solution $x(t) = \sin(kt)/k^2 + x_{0}$, and $k > 0$ is a constant. 
Also consider an \textit{inverse} problem where $k$ is unknown, and $N$ noisy realizations of the solution $x$ are available at irregular time intervals, resulting in a dataset $\{t_i, x_i\}_{i=1}^N$.
To this end, we use MC X-TFC to simultaneously perform three tasks: reconstruct $x(t)$ from partial observations, infer the value of $k$ from both the data and the underlying ODE~\eqref{eq:pedagogical}, and quantify the total uncertainty using $1,000$ MC repetitions.

The reconstructed $x(t)$ is shown in Figure~\ref{fig:example_0_0}, whereas $k$ is estimated as $0.9840\pm0.0536$, which agrees well with the exact value of $k=1$. 
Also, both the reconstructed MC X-TFC solution and its uncertainty are similar to those produced by the ensemble PINN and B-PINN approachs. 
The proposed example only considers data in $[0,5)$, so that extrapolation for $t>5$ mostly relies on satisfaction of \eqref{eq:pedagogical} (i.e., physics-informed regularization) using the predicted value of $k$.
As expected, quantified epistemic uncertainty is smaller in $t\in[0, 5)$ than in $t\in[5, 10)$ due to the uneven distribution of the available data. In addition, the predicted mean of $x(t)$ agrees well with the exact solution, and their difference is appropriately bounded by the predicted uncertainty, as shown in Figure~\ref{fig:example_0_0}.

We also investigate how the choice of the random initialization of weights and bias in the hidden layer affects the predicted uncertainty. To do so, we fix the number of neurons in the hidden layer to $m=20$, and randomly draw initial choices for weights and biases from $\mathcal{U}[-B, B]$ using either $B=1$ or $B=15$. 
A larger $B$ results in larger epistemic (and total) uncertainty, as shown in Figure~\ref{fig:example_0_0}.
Aleatoric uncertainty is determined by the noise model and hence irreducible, and hence is the same across different methods and/or models. 
Finally, the estimated value of $k$ and a computational cost comparison between the different approaches are reported in Table~\ref{tab:example_0}.

\begin{table}[ht!]
\centering
\caption{Estimated values of $k$ (exact value $k=1$) and computational costs of different methods whose results are shown in Figure~\ref{fig:example_0_0}. Estimates of $k$ are reported as $\mu\pm SD$, where $\mu$ and $SD$ are the predicted mean and standard deviation, respectively. Hyperparameters for X-TFC and PINN-based approaches can be found in Section \ref{appendix_b}.}
\label{tab:example_0}
\begin{tabular}{c c c}
\toprule
Methods & Inference of $k$ & Wall time (seconds)\\
\midrule
MC X-TFC ($B=1$) & $0.9840\pm0.0536$ & $17.12$\\
MC X-TFC ($B=15$) & $0.9765\pm0.0747$ & $5.42$ \\
Ensemble ($10$) PINNs & $0.9313\pm0.0317$ & $55.28$\\
B-PINNs & $0.9595\pm0.0354$ & $28.29$\\
\bottomrule
\end{tabular}
\end{table}

X-TFC also allows us to control the relative amount of physics- versus data-informed regularization. To demonstrate this capability, we apply different penalty coefficients to the two main components of the loss function, i.e.
\begin{equation}\label{eq:example_0_loss}
     \boldsymbol{\mathcal{L}} = \lambda_1 \mathcal{L}_{\text{eq}} + \lambda_2 \mathcal{L}_{\text{data}},
\end{equation}
where $\lambda_1$ and $\lambda_2$ are associated with the physics- and data-informed loss component, respectively. 
Specifically, we study the effect of a varying degree of physics-informed regularization under unevenly distributed data. 
When solely relying on data (small $\lambda_1$), the prediction error and corresponding epistemic uncertainty significantly increase, as expected, in regions where observations are missing (see Figure~\ref{fig:example_0_1}).

\subsection{System identification under model-form uncertainty}

\begin{figure}[ht!]
\centering
    \includegraphics[width=\textwidth]{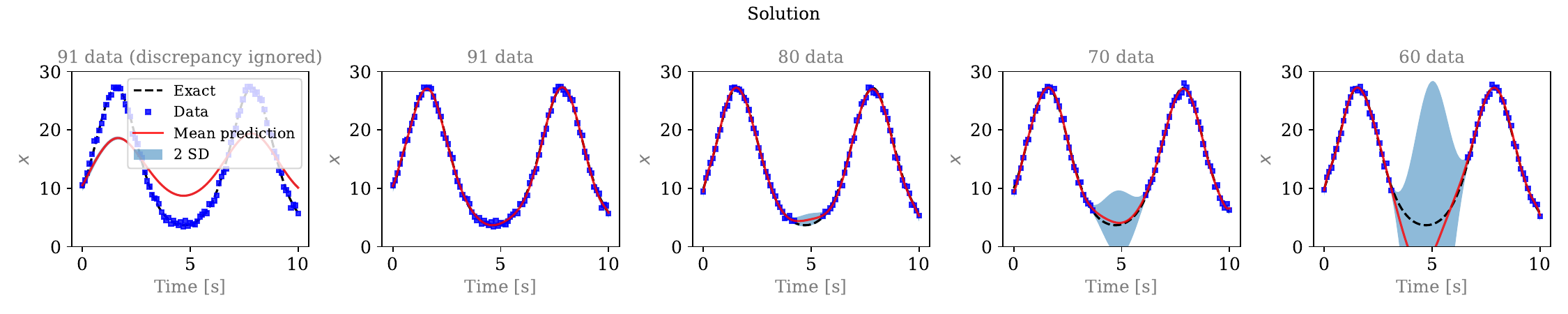}
    \includegraphics[width=\textwidth]{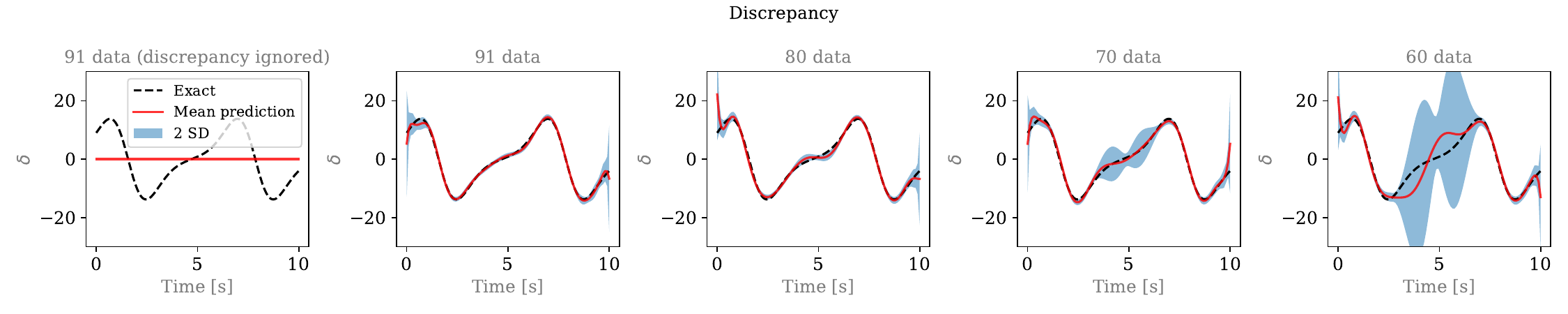}
    \caption{Quantification of epistemic uncertainty in the reconstructed solution of a harmonic equation from an inconsistent dataset of varying size. The first row presents an estimate for the solution $x(t)$ while the second row represents an estimate for the discrepancy $\delta(t)$. The predicted (epistemic) uncertainty of $\delta(t)$ is used to characterize model-form uncertainty. On the leftmost plots, we present the result obtained by ignoring $\delta(t)$, and using the misspecified differential equation, hence demonstrating the use of MC X-TFC in the context of model misspecification.}\label{fig:model_form}
\end{figure}


In this section, we assume that the differential equation~\eqref{eq:pedagogical} is an approximation of a true underlying non-linear model, expressed as the initial value problem
\begin{equation}\label{eq:pedagogical_nonlinear}
    \begin{cases}
    \dfrac{dx(t)}{dt} = \dfrac{x \cos(kt)}{k} \\
    x(0) = x_{0} = 10,
    \end{cases}
\end{equation}
which is used as the data generating process, but whose formulation remains unknown.
This scenario is a source of model-form uncertainty. We deal with this situation by modifying~\eqref{eq:pedagogical} with the addition of an unknown discrepancy, which is learnt, as discussed in Section \ref{sec:xtfc} and Figure \ref{fig:gbx-tfc}, to compensate for the inconsistency between the linear and non-linear models, such as
\begin{equation}
\begin{cases}
\dfrac{dx(t)}{dt} = \dfrac{\cos(kt)}{k} + \delta(t) \\
x(0) = x_{0} = 10.
\end{cases}
\end{equation}
We note that the exact solution is $x(t) = x_0\exp[(1/k^2)\sin(kt)]$ and hence the discrepancy can be computed exactly as $\delta(t) = (x(t)-1)[\cos(kt)/k]$.

Results are presented in Figure~\ref{fig:model_form}, where the reconstructed $x(t)$ and discrepancy $\delta(t)$ are shown in the first row and second row, respectively, together with the quantified uncertainty. 
On the leftmost plot in Figure~\ref{fig:model_form}, we present the consequence of utilizing the misspecified model directly (with $k$ being unknown and learnable) while ignoring the discrepancy, i.e. $\delta(t) = 0, \forall t\in[0, 10]$. 
As shown, MC X-TFC fails to fit the data, as the underlying equation does not agree with the available observations. 
Modeling the discrepancy $\delta(t)$ with an additional network allows the equation loss $\mathcal{L}_{\text{eq}}$ and the data loss $\mathcal{L}_{\text{data}}$ to be simultaneously minimized, so that the data of $x$ are fitted and the (corrected) equation is satisfied~\cite{zou2024correcting}. 
From Figure~\ref{fig:model_form}, we can see that the discrepancy is also accurately captured when irregularly sampled and noisy data are available. 
As the number of data decreases, the accuracy of both the reconstructed $x(t)$ and inferred $\delta(t)$ is reduced, and the errors between their predicted mean and true solution are bounded by the predicted uncertainties.
We note that in this case, we fix $k=1$ to avoid solution multiplicity brought by the unknown parameter $k$ and unknown discrepancy $\delta(t)$.
This is a modeling choice which interacts with the quantification of model-form uncertainty. This aspect will be further discussed in Section~\ref{sec:cvsim6_model_form}.


\section{Results for the CVSim-6 cardiovascular model}\label{sec:results}

We consider two applications of MC X-TFC to the CVSim-6 cardiovascular system.
The first consists of an \emph{ablation} study, where we are interested in determining how much data is needed for MC X-TFC to accurately estimate states related to synthetically generated time histories of blood pressure, flow, and volume while, at the same time, estimating a pulmonary resistance and a systemic compliance parameter. In addition, we would like to quantify the total uncertainty (aleatoric plus epistemic) associated with these predictions. 

The second application focuses on model-form uncertainty, particularly fitting data with an inadequate model. This situation arises very often with lumped parameter models in hemodynamics, for example, when using perfect unidirectional valves without accounting for possible regurgitation~\cite{pant2018lumped}, when neglecting flow contributions from collateral flow, or when excluding atria or organ-level compartments from the model.

\subsection{Ablation study under combined aleatoric and epistemic uncertainty}

The MC X-TFC framework allows to naturally combine information from the available data and the CVSim-6 model equations. Therefore it offers an ideal testbed for determining the minimum amount of data needed to estimate states or parameters of a given system, and also to quantify the uncertainty associated with these estimates. 
To demonstrate this process in the context of computational physiology, we perform an ablation study where pressure data is progressively removed under physics-informed regularization, and first one and then two parameters are simultaneously estimated.
We then report the estimated pressure, flow, and volume traces and their variability under combined aleatoric and epistemic uncertainty. 
This analysis is conducted across six scenarios, as outlined in Table~\ref{table:abl_scenarios}, and considers data acquired on six pressures and two parameters -- the pulmonary venous resistance and the systemic arterial compliance -- since they are clinically relevant in the assessment of cardiovascular function.
\begin{table}[h!]
\centering
\caption{Matrix of measurements and parameters used in the ablation study.}\label{table:abl_scenarios}
\begin{tabular}{c c c c c c c c}
\toprule
Type & Qty & Sc1  & Sc2 & Sc3 & Sc4 & Sc5 & Sc6\\
\midrule
\multirow{6}{*}{Measurements} & $P_l$  & \cmark & \cmark & \xmark & \xmark & \xmark & \xmark\\
 & $P_a$ & \cmark & \cmark & \cmark & \cmark & \cmark & \cmark\\
 & $P_v$  & \cmark & \cmark & \cmark & \cmark & \xmark & \xmark\\
 & $P_r$ & \cmark & \cmark & \cmark & \xmark & \xmark & \xmark\\
 & $P_{pa}$ & \cmark & \cmark & \cmark & \cmark & \cmark & \cmark\\
 & $P_{pv}$ & \cmark & \xmark & \xmark & \xmark & \xmark & \xmark\\
\midrule
\multirow{2}{*}{Parameters} & $r_{pv}$ & \xmark & \xmark & \xmark & \xmark & \xmark & \xmark\\
& $c_{a}$ & \cmark & \cmark & \cmark & \cmark & \cmark & \xmark\\
\bottomrule
\end{tabular}
\end{table}

\subsubsection{State and parameter estimation under total uncertainty}

\begin{figure}[ht!]
\centering
    \includegraphics[width=0.48\textwidth]{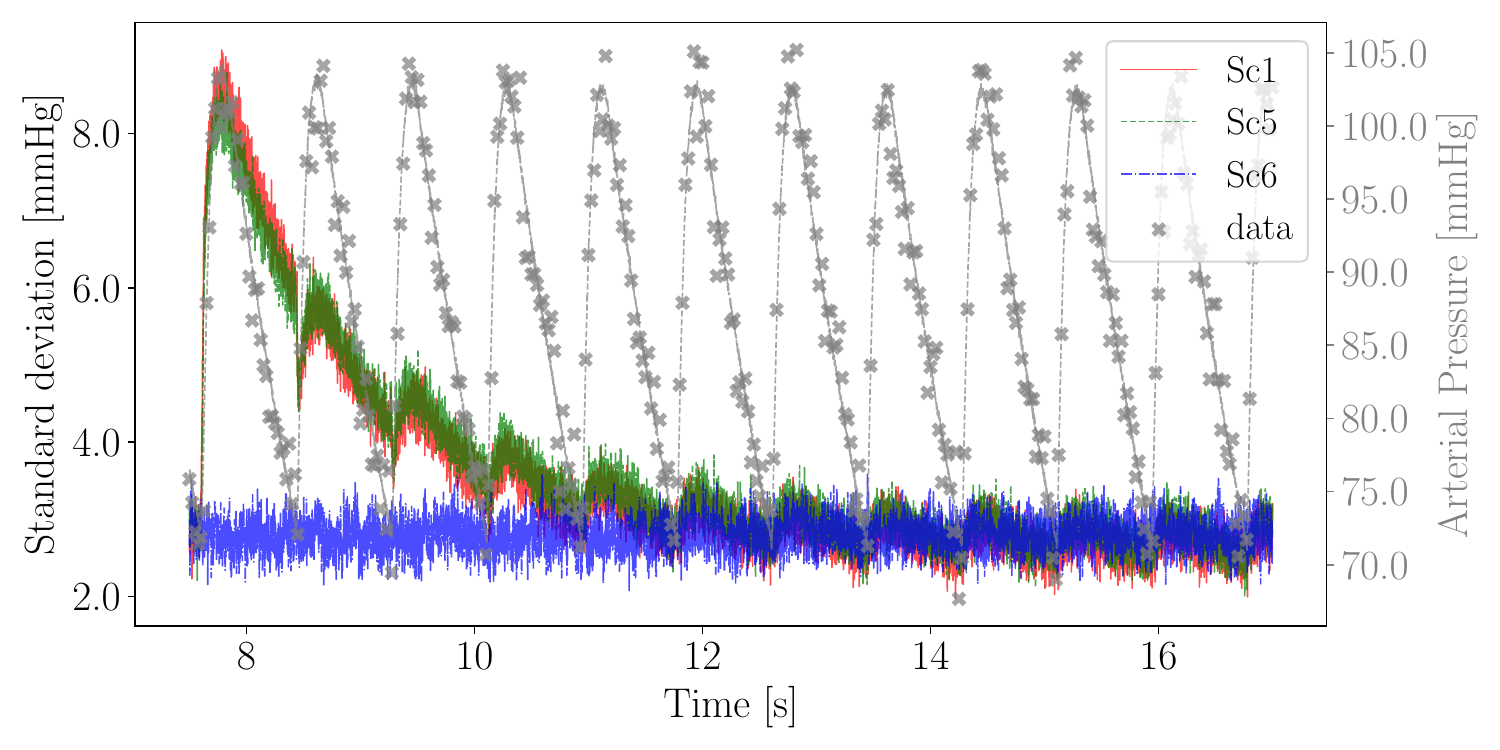}
    \includegraphics[width=0.48\textwidth]{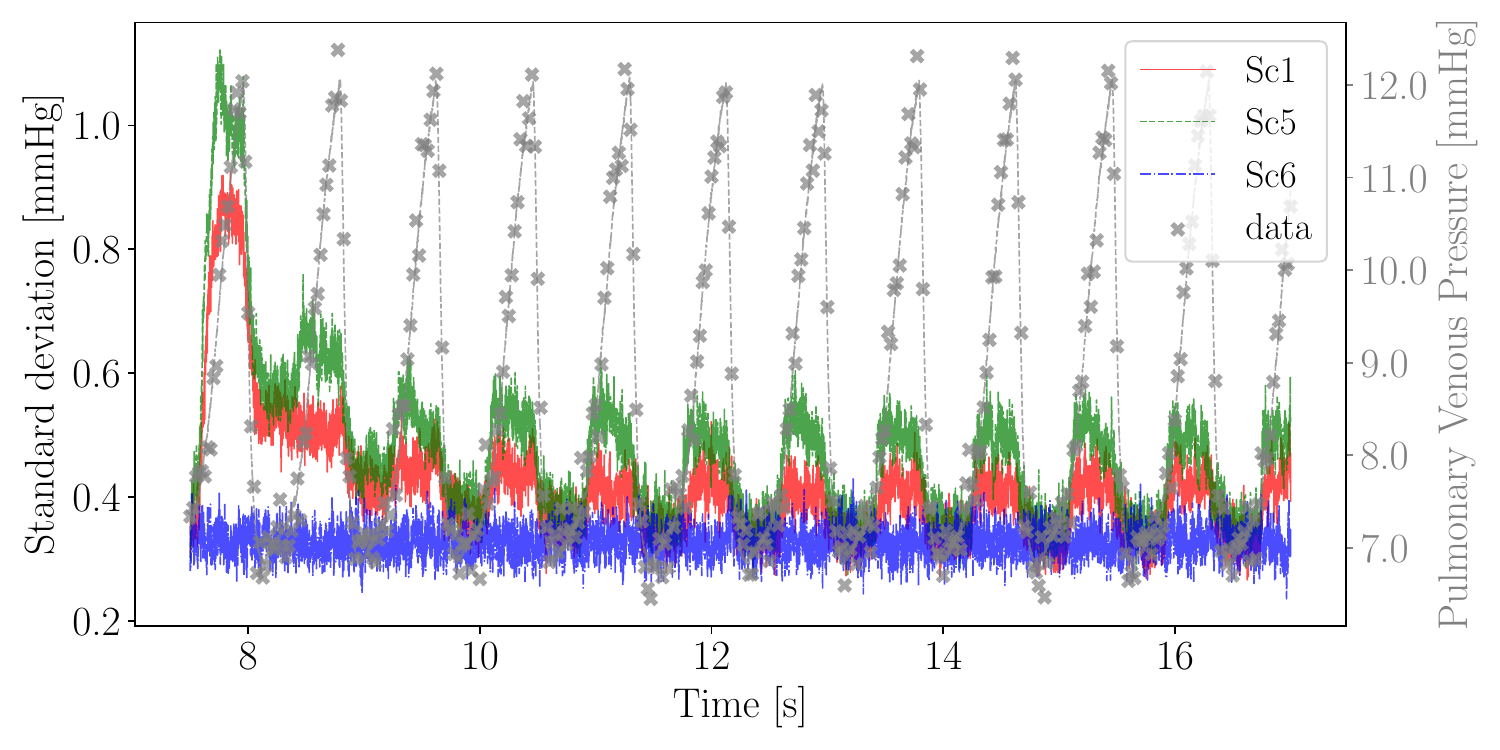}
    \caption{Standard deviations and mean values for reconstructed pressures under aleatoric and epistemic uncertainty. The mean solution for Sc1 and the pressure data are also shown in gray.}\label{fig:abl_pres_std}
\end{figure}

Figure~\ref{fig:abl_pres_std} shows the mean and standard deviation for the estimated systemic arterial and pulmonary venous pressures under total uncertainty.
After a few cardiac cycles, the Monte Carlo standard deviation of the two pressures reduces to approximately 3.0 and 0.5 mmHg for $P_{a}$ and $P_{pv}$, respectively, with values that are only marginally affected by the difference of data availability across scenarios. 
Knowledge of the correct underlying equations allows MC X-TFC to identify the system's response under limited uncertainty, even when simultaneously estimating unknown parameters.
However, the loss of information created by the missing parameters needs to be compensated by providing pressure data on the same compartments (systemic and pulmonary, respectively).

It is also evident from Figure~\ref{fig:abl_pres_std} that the amount of variability reduces with time before reaching a periodic behavior for all scenarios except scenario 6. 
This is the result of a \emph{filtering} process due to physics-informed regularization of random initial conditions, as suggested by Equation~\eqref{equ:ode_subintervals}.
In other words, for Sc1-Sc5, the CVSim-6 equations are alone sufficient to reconstruct the physiological response, and availability of noisy data provides redundant information that is \emph{distilled} over time. 
This is confirmed by the smooth time history of the standard deviation for the pressure state variables under epistemic uncertainty. 
When instead two parameters are estimated in Sc6, pressure data becomes essential to the estimation process, as confirmed by the more \emph{noise-like} time history for the standard deviation under epistemic uncertainty.
In summary, the transition between Sc5 and Sc6 represents a switch from an estimation process, where the physics and data \emph{compete}, to a process where the physics and data \emph{cooperate}.

\begin{figure}[ht!]
\begin{subfigure}[b]{0.58\textwidth}
\begin{subfigure}[b]{\textwidth}
\centering
\includegraphics[width=\textwidth]{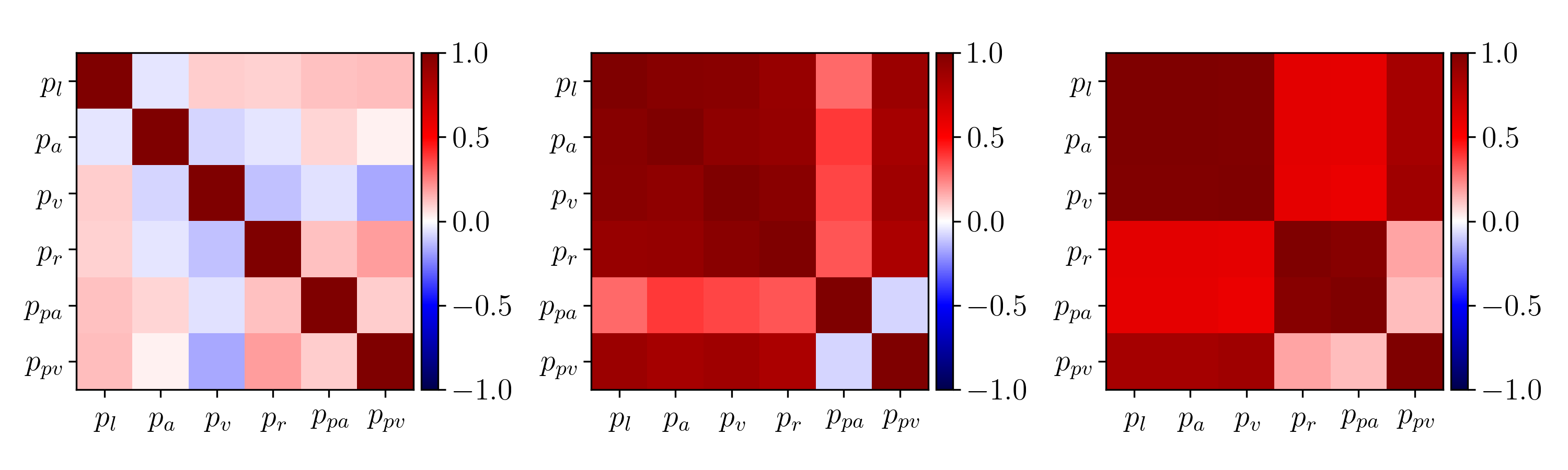}
\caption{Noise correlation for scenario 1.}
\label{fig:noise_corr_sc1}
\end{subfigure}
\begin{subfigure}[b]{\textwidth}
\centering
\includegraphics[width=\textwidth]{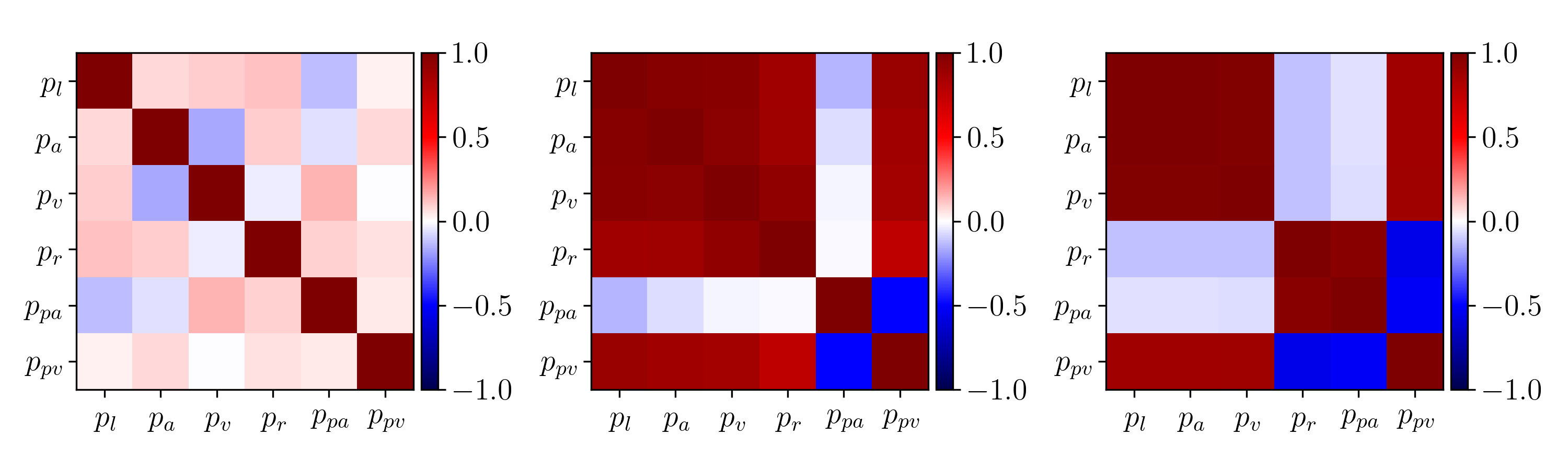}
\caption{Noise correlation for scenario 5.}
\label{fig:noise_corr_sc5}
\end{subfigure}
\end{subfigure}
\begin{subfigure}[b]{0.38\textwidth}
\includegraphics[width=\textwidth]{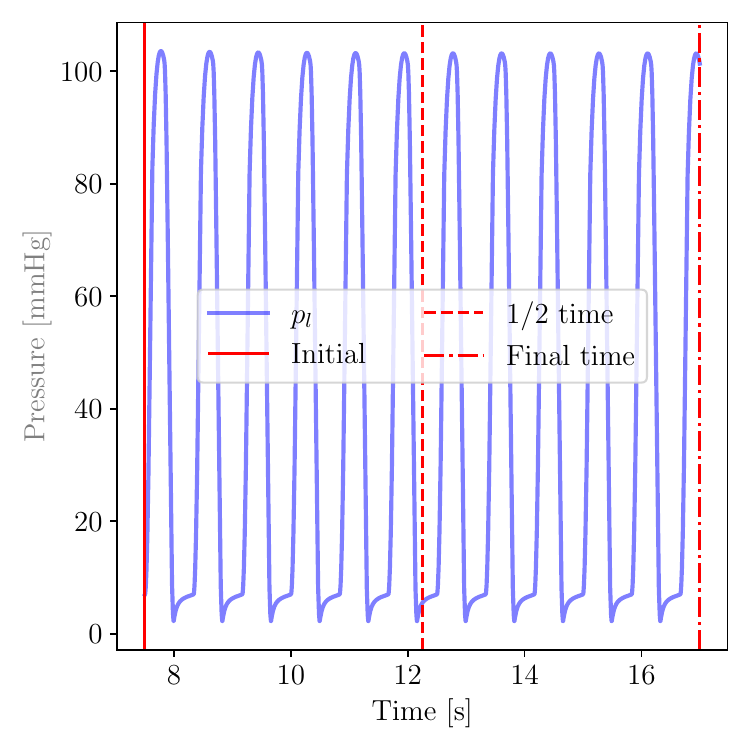}
\caption{Acquisition times for noise correlation.}
\label{fig:noise_time}
\end{subfigure}
\caption{Time snapshots of correlation matrices for pressure unknowns.}
\label{fig:noise_corr}
\end{figure}

To further understand how the statistical structure of the noise is affected by this \emph{filtering} process, the noise correlation matrix for all compartmental pressures is also shown in Figure~\ref{fig:noise_corr}. 
Correlations are representative of three solution snapshots, at the beginning, half-time, and final simulation time, which correspond to early systole, diastole, and systole, respectively. Correlation matrices are shown for scenarios 1 and 5 in Figure~\ref{fig:noise_corr_sc1} and~\ref{fig:noise_corr_sc5}, respectively. 
The first snapshot confirms that noise in the initial condition is independently applied on each pressure.
The second and third snapshots (see Figure~\ref{fig:noise_time}) show a high correlation between all pressure components except $P_{pa}$ in diastole and ($P_{pa}$, $P_{r}$) in systole. 
The smooth pressure reconstruction achieved by X-TFC results in highly correlated noise among different pressure components, facilitated by the communication between compartments following valve openings. In contrast, the lack of correlation between $P_{pa}$ and $P_{r}$ is attributed to variability in the $R_{pv}$ parameter, which fluctuates as estimates are updated over time. Unlike Sc1 and Sc5, in Sc6 the noise in different pressure components remains uncorrelated over time (not shown).

In scenarios 1 to 5, the CoV is approximately 3\% for $P_{a}$, $P_{v}$ and $P_{pa}$, 6\% for $P_{l}$ and $P_{r}$ (due to higher measurement noise) and 5\% for $P_{pv}$, as a result of the estimation of $r_{pv}$ in Sc1 to Sc5.
CoV is instead below $2\%$ for all volumes except for pulmonary veins where it slightly above $2.5\%$.
Flow uncertainty is sensibly higher, with CoV for the systemic flow (i.e., systemic arterial flow, left ventricular outflow, and right ventricular inflow) equal to approximately $2.5\%$, $8.5\%$ for the pulmonary flow CoV, except for the right ventricular outflow where it is approximately $13.5\%$.
Since flows are estimated by minimizing a residual that contains derivatives of a pressure approximated from noisy observations, this higher variability is expected.

A regime shift is observed for Sc6, where the arterial compliance $c_{a}$ is also estimated as part of the solution process. 
In this scenario, the uncertainty doubles for the left and right ventricular pressures and increases substantially for the left ventricular volume, inflow, and outflow. 
Satisfaction of noisy pressure measurements on the systemic pressure (remember that only $P_{a}$ and $P_{pa}$ are observed in this scenario) can be realized by either changing the left ventricular volume or the aortic compliance (whose estimated value changes with time). If this was a Bayesian estimation problem, we would say that the posterior marginal of left ventricular volume and aortic compliance shows a negative correlation due to a lack of identifiability from pressure data.

\begin{figure}
\centering
\includegraphics[width=\textwidth]{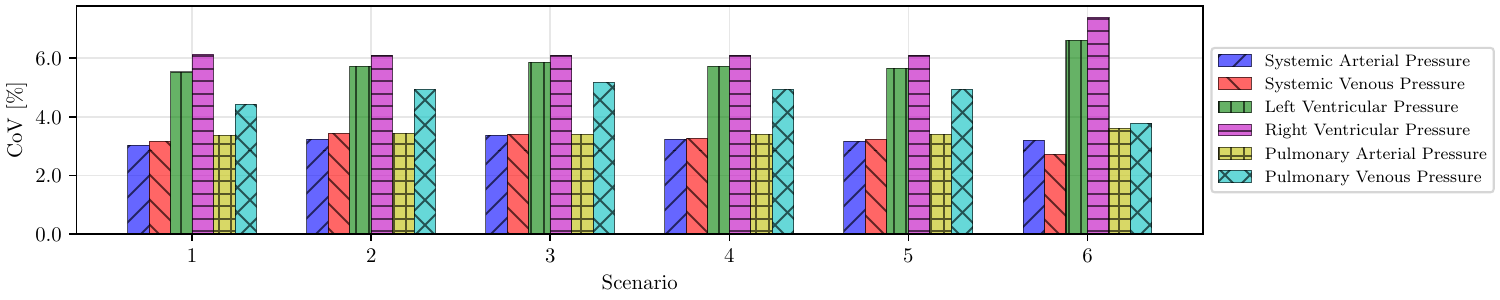}\\
\includegraphics[width=\textwidth]{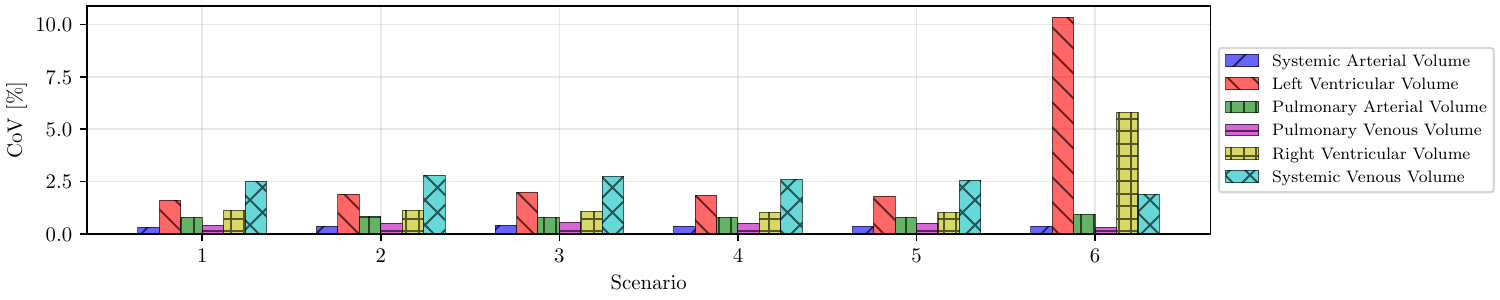}\\
\includegraphics[width=\textwidth]{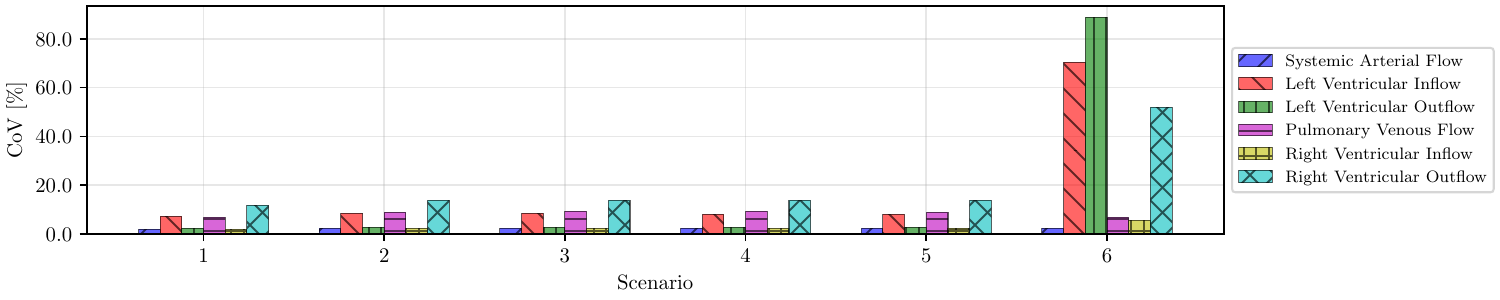}
\caption{Ablation study summary. Time-average coefficients of variations (last two cardiac cycles) for pressure (top), volume (center), and flow (bottom) quantities of interest and all scenarios.}\label{fig:abl_summary}
\end{figure}

\begin{figure}[ht!]
\centering
    \includegraphics[width=0.32\textwidth]{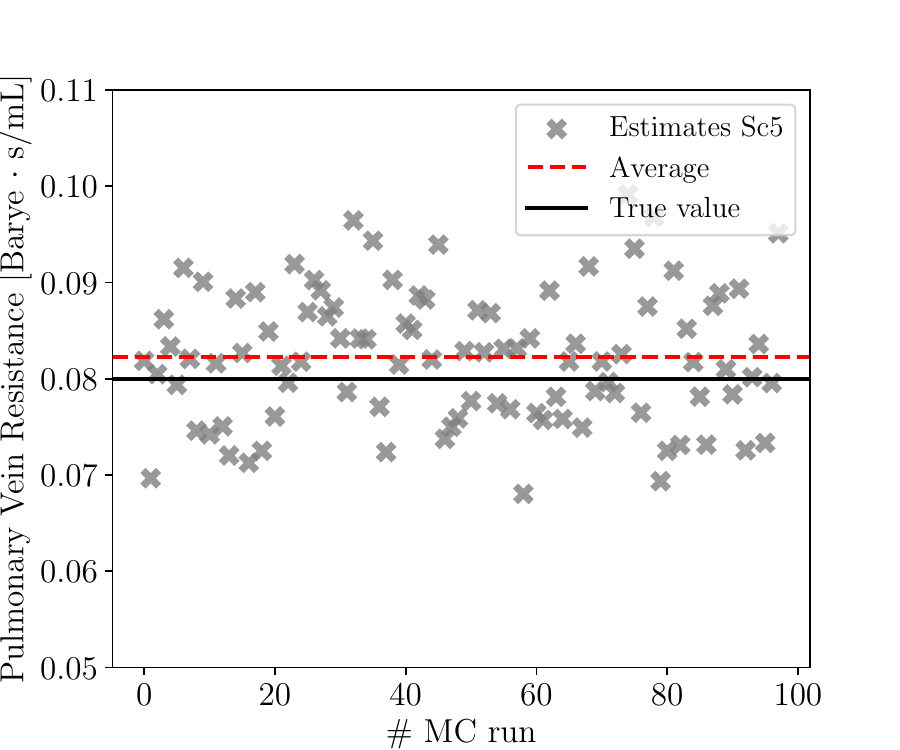}
    \includegraphics[width=0.32\textwidth]{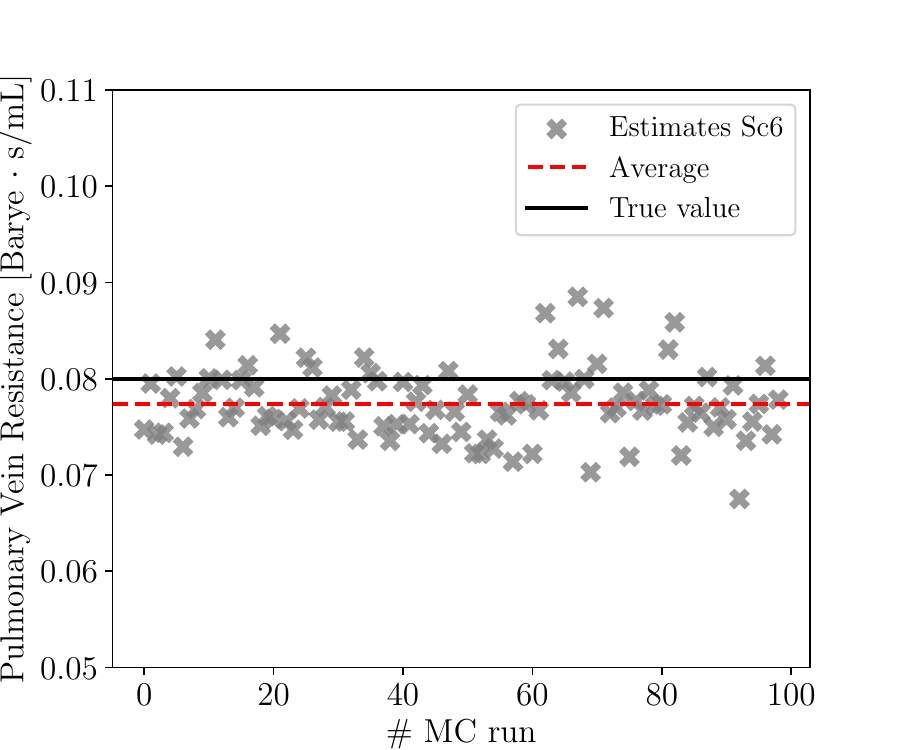}
    \includegraphics[width=0.32\textwidth]{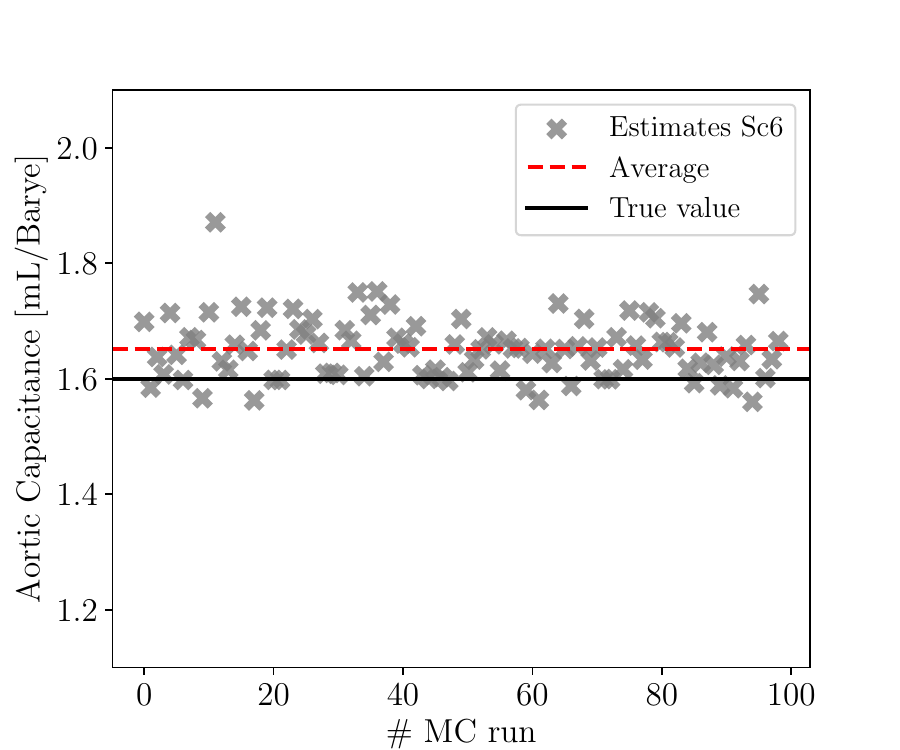}
    \caption{Estimates of pulmonary resistance and aortic compliance for Scenarios 5 and 6.}\label{fig:params}
\end{figure}

We also investigate the ability of MC X-TFC to estimate physiologically relevant parameters. Parameter estimates are computed on a per-simulation basis and averaged over a number of Monte Carlo runs, as shown in Figure~\ref{fig:params}. Final averages agree well with true parameter values.

\begin{figure}[ht!]
\centering
    \includegraphics[width=0.48\textwidth]{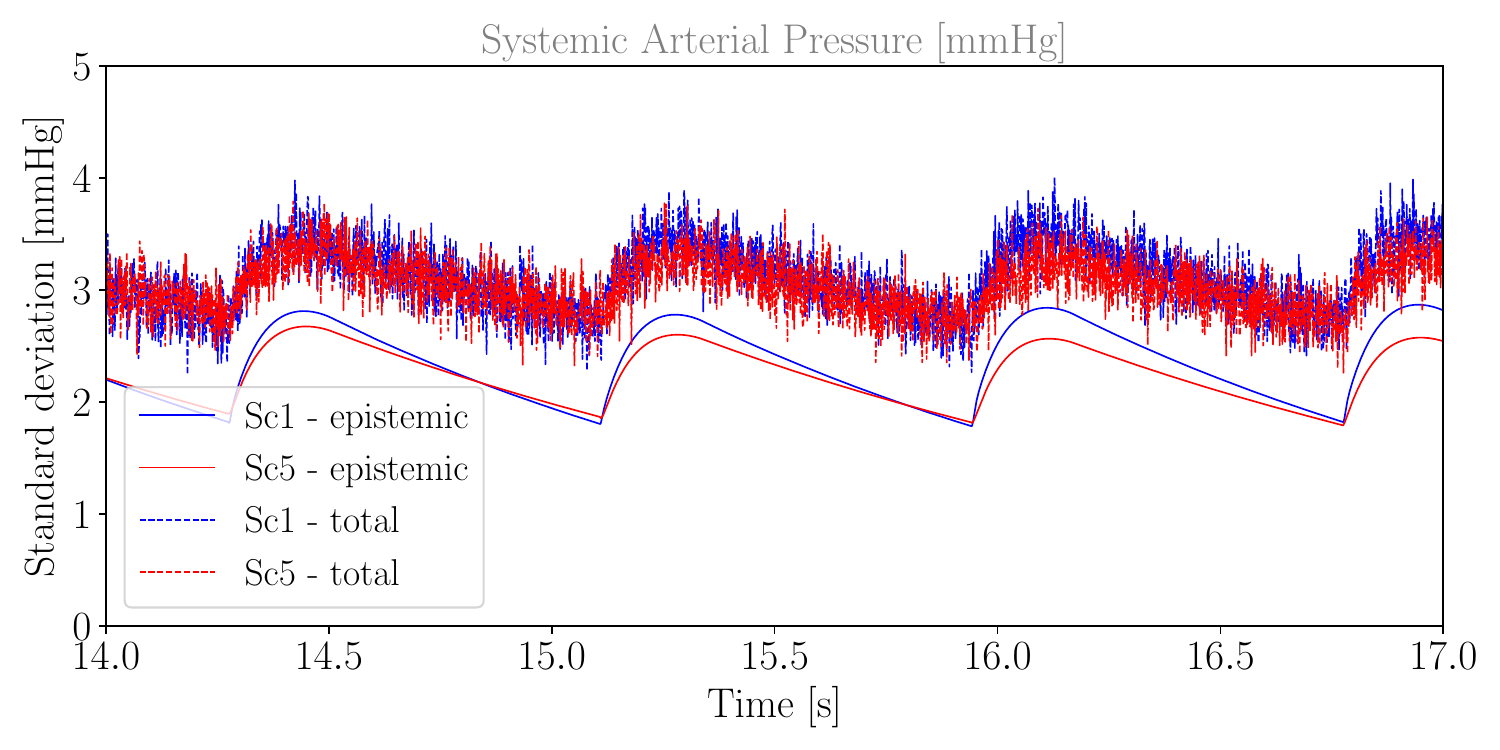}
    \includegraphics[width=0.48\textwidth]{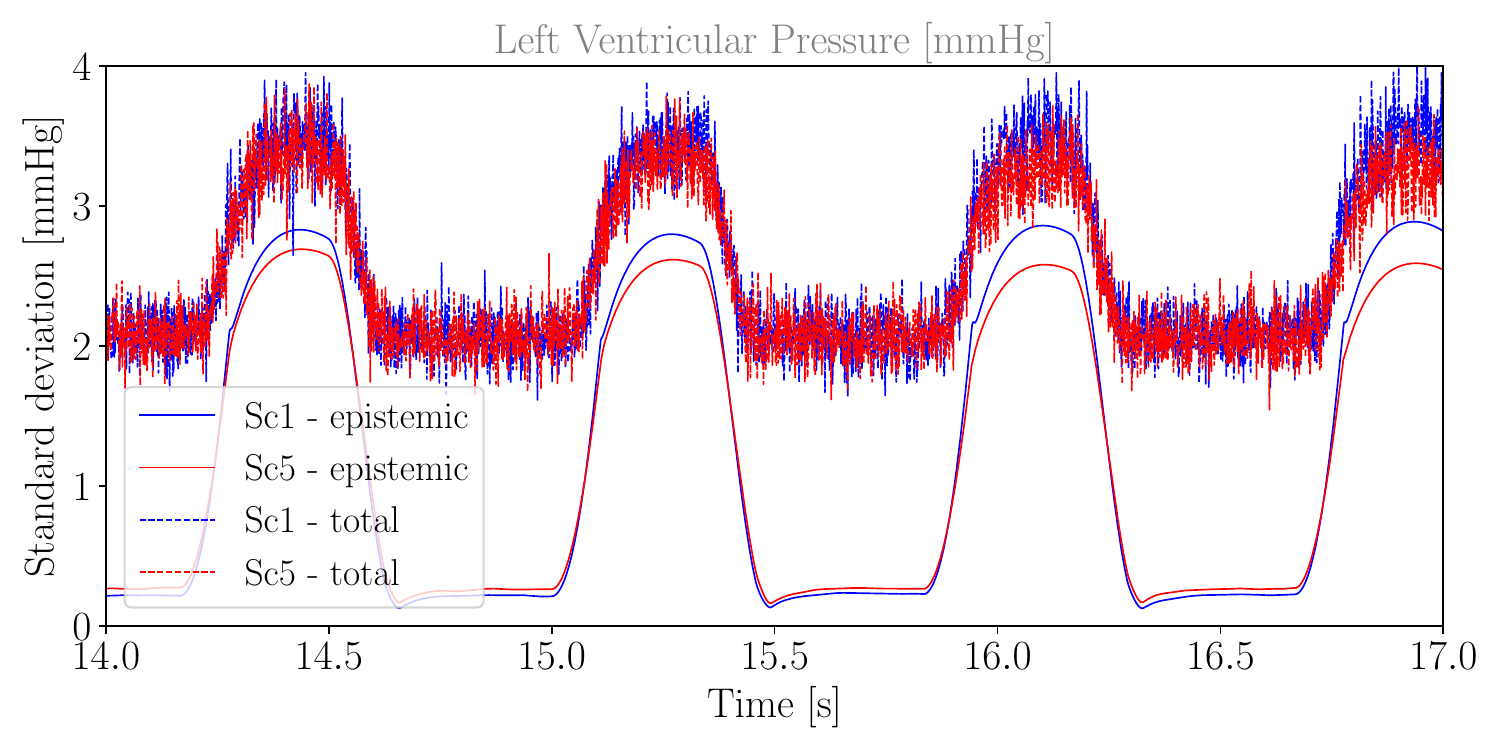}\\
    \includegraphics[width=0.48\textwidth]{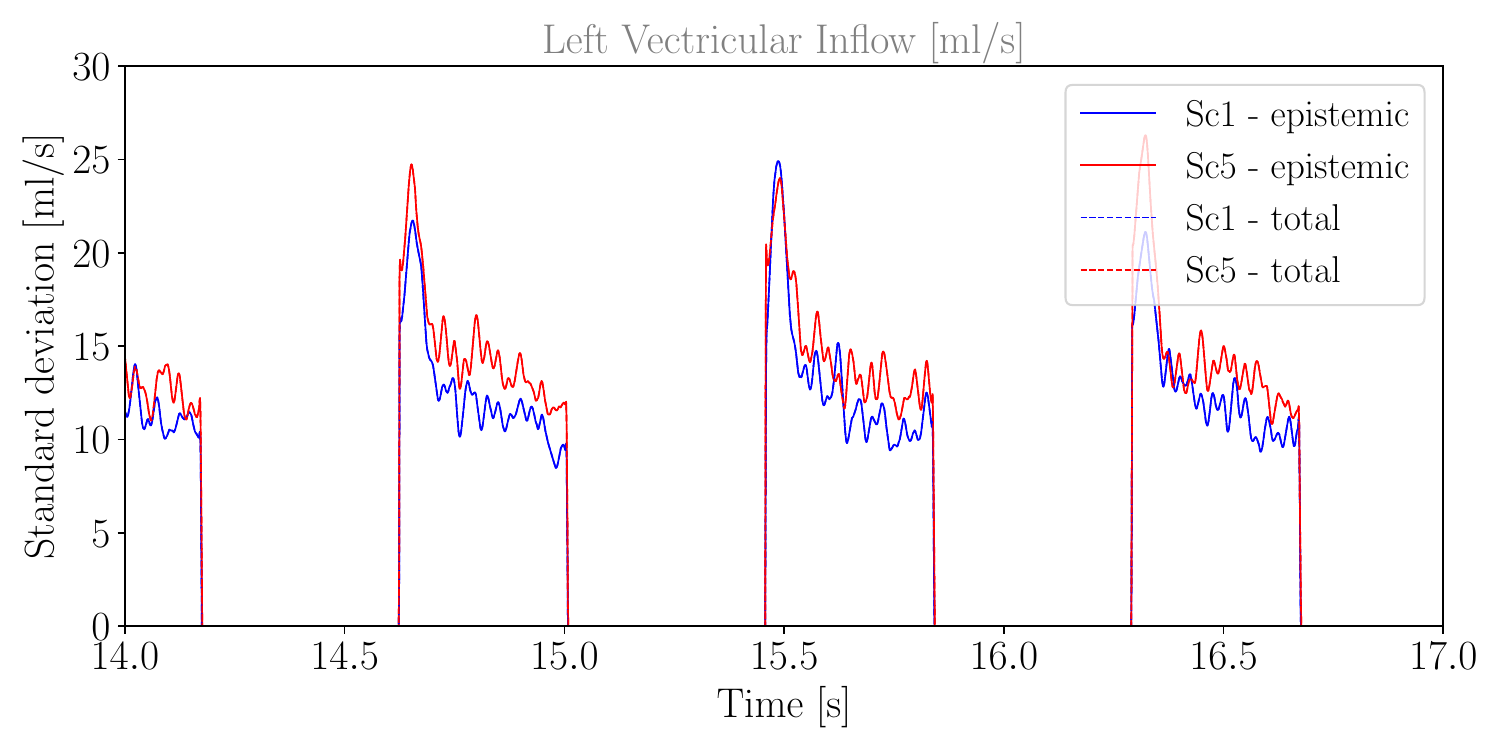}
    \includegraphics[width=0.48\textwidth]{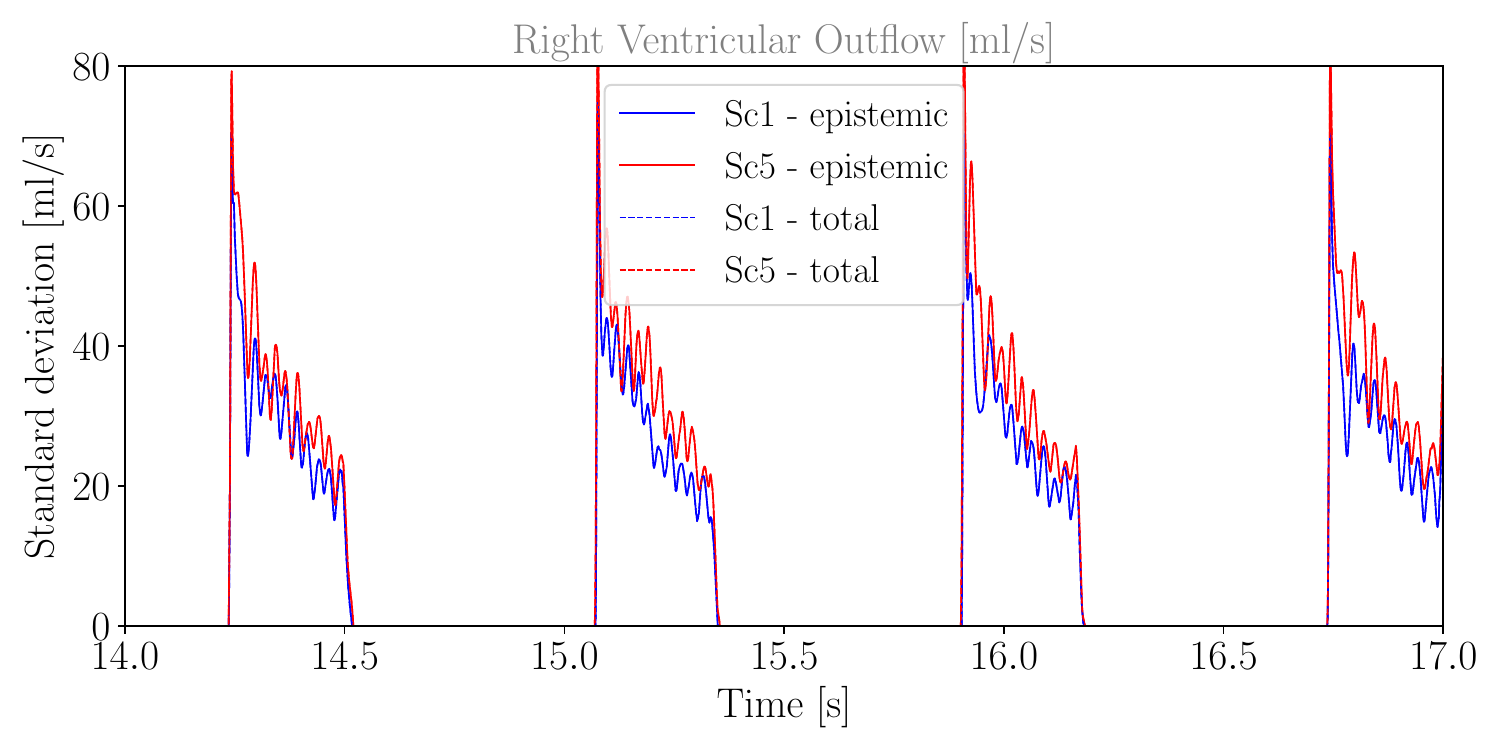}
    \caption{Decomposition of total uncertainty in Sc1 and Sc5 for selected pressure and flow QoIs. For demonstration purposes, we have assumed the possibility of measuring flow with infinite precision, resulting in negligible aleatoric uncertainty.}\label{dec_uq_pres}
\end{figure}

Finally, we analyze how total uncertainty can be decomposed into its aleatoric and epistemic components, considering only scenarios 1 and 5. 
Figure~\ref{dec_uq_pres} illustrates this decomposition for $P_{a}$, $P_{l}$, $Q_{li}$, and $Q_{lo}$. The oscillating nature of the total uncertainty is consistent with its calculation as a Monte Carlo estimate based on 100 samples.
For systemic arterial and left ventricular pressures, the uncertainty is mostly epistemic, except at diastole, where the signal-to-noise ratio is smaller. 
Flow and volume quantities of interest are estimated from MC X-TFC as a result of the CVSim-6 equations and observed data. As a result, any characterization of aleatoric uncertainty comes directly from assumptions about the precision of the tool or device used to measure such quantities. 
In Figure~\ref{dec_uq_pres}, we highlight this by assuming infinite precision, causing epistemic and total uncertainty to coincide. In other words, epistemic uncertainty in \emph{derived} variables (e.g., blood flows and volumes) does not affect the estimation process, and can be added in post-processing.

The results for all scenarios are obtained with subdomains of length 0.001 seconds, 5 collocation points per subdomain, and 5 neurons. With this setup, the computational time is about 6 seconds per MC realization and approximately 10 minutes for quantifying total uncertainty through 100 MC repetitions. Therefore, the fast execution times for MC X-TFC and the fact that no offline training is required make it ideal for online state and parameter estimation.

\subsection{Misspecified compartment: linear approximation of non-linear pulmonary resistance}\label{sec:cvsim6_model_form}

In this section, we investigate an important aspect of simulating circulatory systems with lumped parameter models, that typically provide an oversimplified representation of the physiological response based on linear RLC circuits, ideal unidirectional valves, and that often selectively consider the presence of organ-level compartments, depending on the application.
With reference to the CVSim-6 system, we consider a specific equation from~\eqref{equ:cvsim6-flow} governing flow in the pulmonary compartment
\begin{equation}\label{eq:lin_pv_model}
Q_{\text{lin},pv}(t) = \displaystyle \frac{P_{pa}(t) - P_{pv}(t)}{R_{pv}}.
\end{equation}
Due to the large number of branches that typically characterize the pulmonary arterial tree, we consider this linear relation to be an approximation of the more accurate nonlinear one
\begin{equation}\label{eq:nonlin_pv_model}
Q_{\text{nonlin},pv}(t) = \displaystyle \frac{P_{pa}(t) - P_{pv}(t)}{R_{pv}(Q_{pv})},
\end{equation}
where a nonlinear resistance is added, which accounts for the larger contribution of minor pressure losses at bifurcations for an increasing pulmonary flow (see Figure~\ref{fig:pulm_res}).
To account for the difference between a linear and non-linear pressure-flow behavior, we use the flexibility of MC X-TFC to add a \emph{discrepancy} term $\delta(t,\bm{\beta})$ to~\eqref{eq:lin_pv_model}, such that
\begin{equation}\label{eq:disc_pv_model}
Q_{\text{disc},pv}(t) = \displaystyle \frac{P_{pa}(t) - P_{pv}(t)}{R_{pv}} + \delta(t,\bm{\beta}),
\end{equation}
and estimate this term from the available pressure data plus the satisfaction of the CVSim-6 differential equations, by approximating it with another NN, such as
\begin{equation}
    \delta(t,\bm{\beta}) = \boldsymbol{\sigma} \boldsymbol{\beta}_{\delta}. 
\end{equation}
Then, we verify the capability of MC X-TFC to correctly estimate the pulmonary flow discrepancy under deterministic conditions  and both aleatoric and epistemic uncertainty.
In the following, we will refer to equations~\eqref{eq:lin_pv_model}, \eqref{eq:nonlin_pv_model} and~\eqref{eq:disc_pv_model} as the \emph{linear}, \emph{nonlinear}, and \emph{linear + discrepancy} model, respectively.

\begin{figure}
\begin{subfigure}[c]{0.48\textwidth}
\centering
\includegraphics[width=0.7\textwidth]{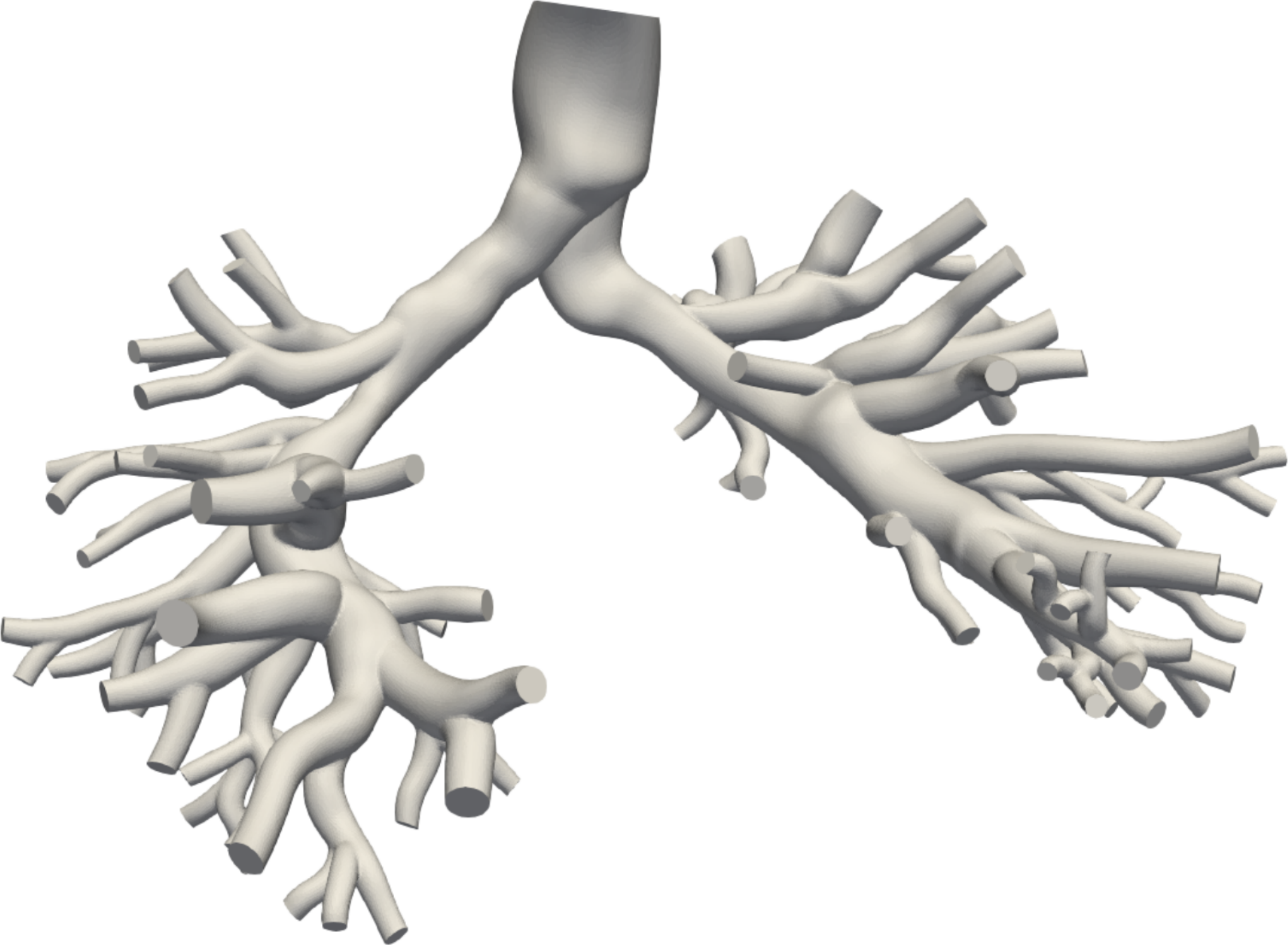}
\caption{Selected pulmonary arterial geometry.}
\label{fig:pulm_res_nl_img}
\end{subfigure}
\begin{subfigure}[c]{0.48\textwidth}
\centering
\begin{tabular}{l c c c c} 
\toprule
{\bf Qty} & {\bf $Q_{pv}$ [L/min]} & {\bf $R_{pv}$ [Barye$\cdot$s/ml]}\\
\midrule
{\bf Diastolic} & 0.12 & 13.0\\
{\bf Mean} & 5.8 & 104.9\\
{\bf Systolic} & 19.8 & 327.8\\
\bottomrule
\end{tabular}
\caption{Pulmonary flows and resistance from three-dimensional rigid wall CFD analysis.}\label{fig:pulm_res_nl_table}
\end{subfigure}
\caption{Determination of nonlinear pulmonary resistance from three-dimensional arterial tree model.}
\end{figure}

The values of linear and non-linear resistance for this application are selected with reference to the pulmonary arterial anatomy shown in Figure~\ref{fig:pulm_res_nl_img}. 
In Table~\ref{fig:pulm_res_nl_table} we report the equivalent model resistance under mean, systolic and diastolic flow, where the strong dependence of resistance from flow is evident. 
Note how the resistance under mean flow is very similar to the value of $R_{pv}$ in Table~\ref{table:params_res}. 
Therefore, we consider a linear model with the default resistance $P_{pv}$ in Table~\ref{table:params_res}, and use linear interpolation to determine a flow-dependent resistance to be used in the nonlinear model.

\begin{figure}[ht!]
\centering
    \includegraphics[width=0.32\textwidth]{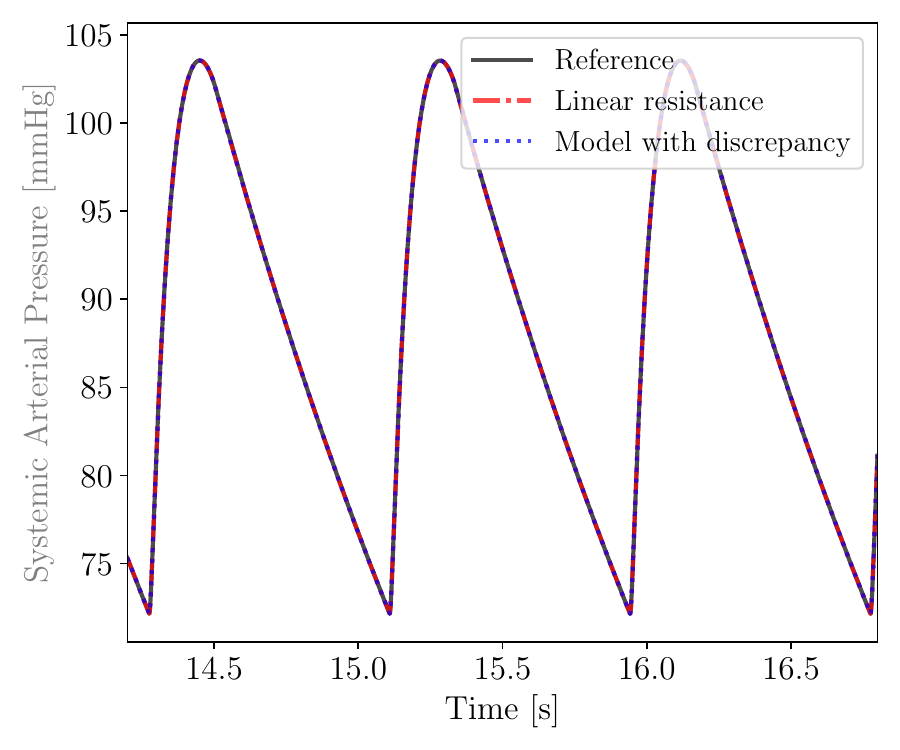}
    \includegraphics[width=0.32\textwidth]{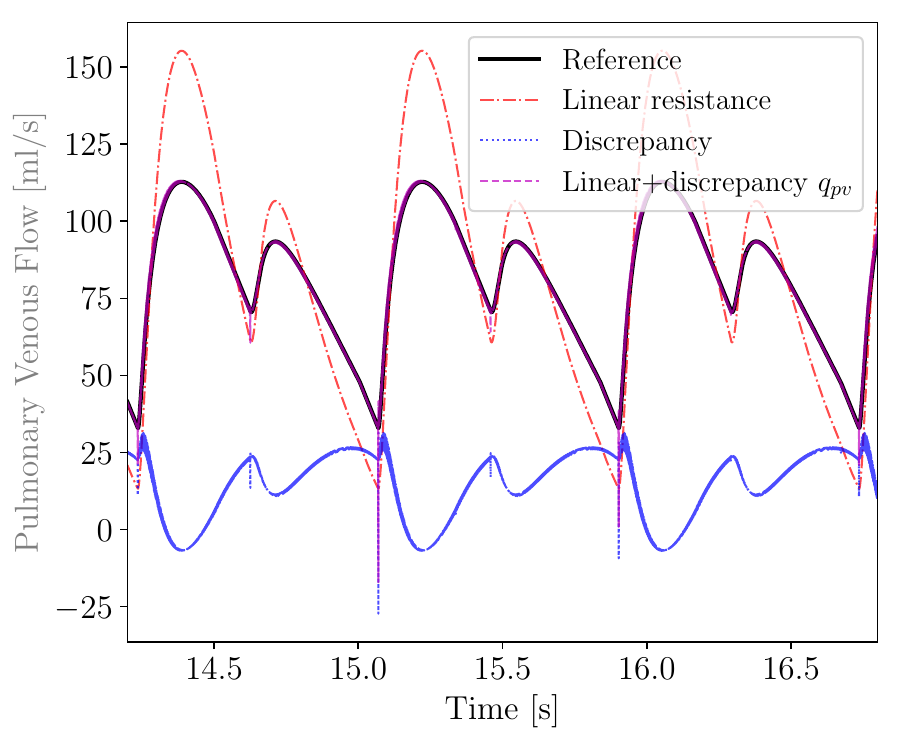}
    \includegraphics[width=0.32\textwidth]{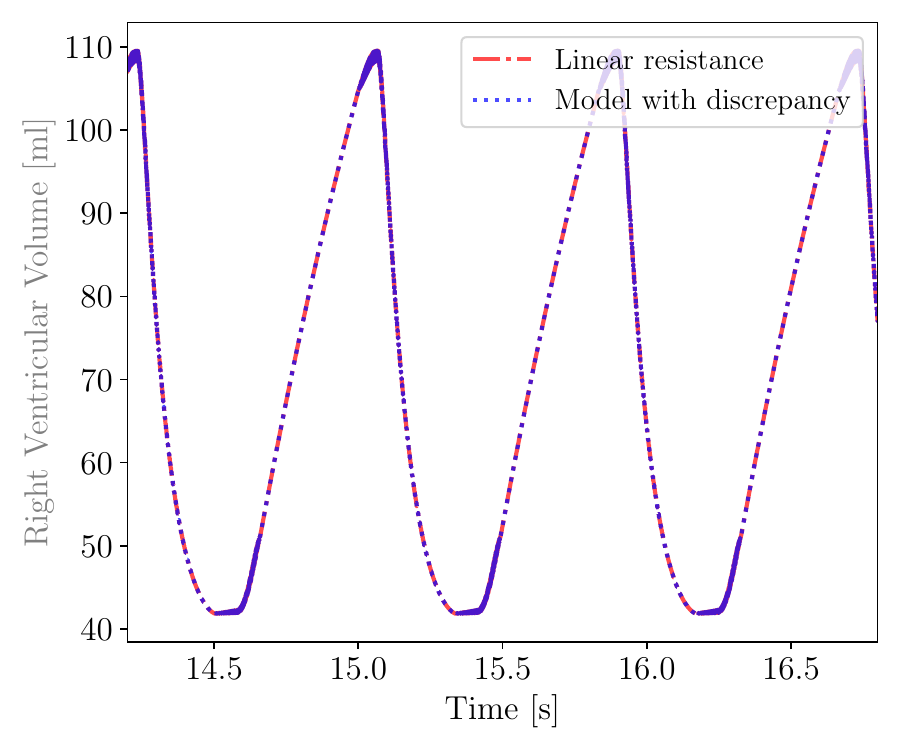}
    \caption{Difference in arterial pressure (left), pulmonary venous flow (center) and right ventricular volume (right) between true nonlinear model, and linear model with and without discrepancy, under ideal noiseless data. The plot on the left and right show how the use of a misspecified linear resistor only affects flow and pressure in the pulmonary compartment, under noiseless data synthetically generated by the true underlying non linear model. As shown in the central figure, the flow estimated by a misspecified linear model overestimates the true systolic flow as a result of considering a constant rather than flow dependent pulmonary resistance. The central figure also shows how the discrepancy learned by X-TFC correctly restores the correct pulmonary flow.}\label{fig:disc_noUQ}
\end{figure}

We then investigate the ability of MC X-TFC to recover the correct discrepancy under ideal noiseless measurements and complete knowledge of the CVSim-6 equations. 
We generate ideal data from the true nonlinear model and reconstruct the physiological response using the linear and discrepancy models.
The resulting difference in arterial pressure, pulmonary venous flow, and right ventricular volume between these three models is illustrated in Figure~\ref{fig:disc_noUQ}. 
Since the linear resistance was selected equal to the resistance at mean flow, and the linear and nonlinear models have identical capacitance, the difference in response between the two models remains confined to the pulmonary venous compartment. 
Therefore, under ideal noiseless conditions, X-TFC correctly recovers the flow discrepancy, as shown in Figure \ref{fig:disc_noUQ}.

\begin{figure}[ht!]
\centering
    \includegraphics[width=0.24\textwidth]{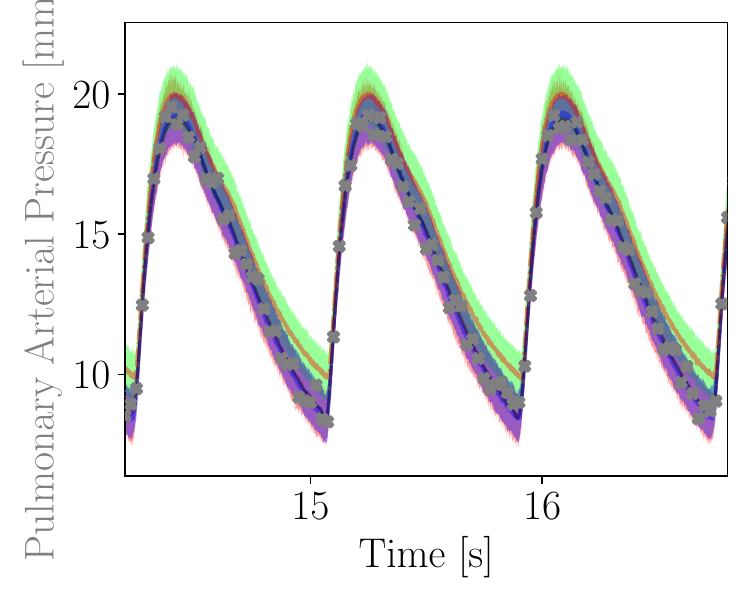}
    \includegraphics[width=0.24\textwidth]{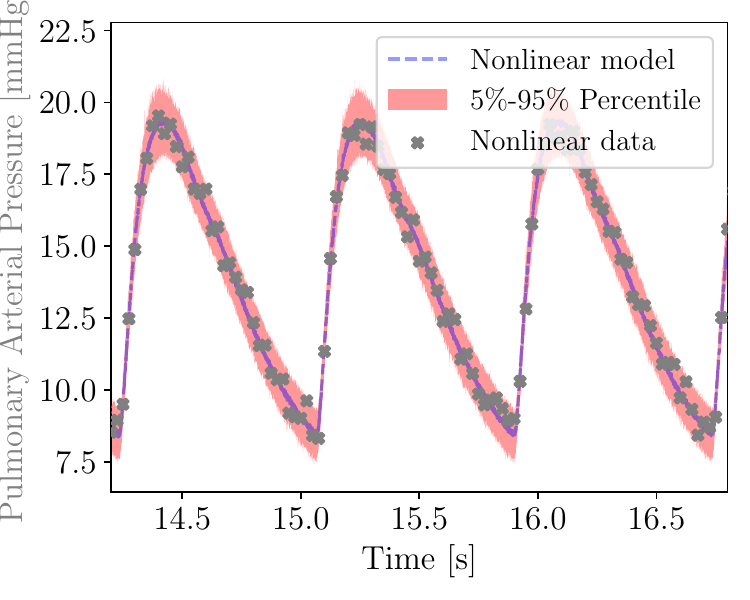}
    \includegraphics[width=0.24\textwidth]{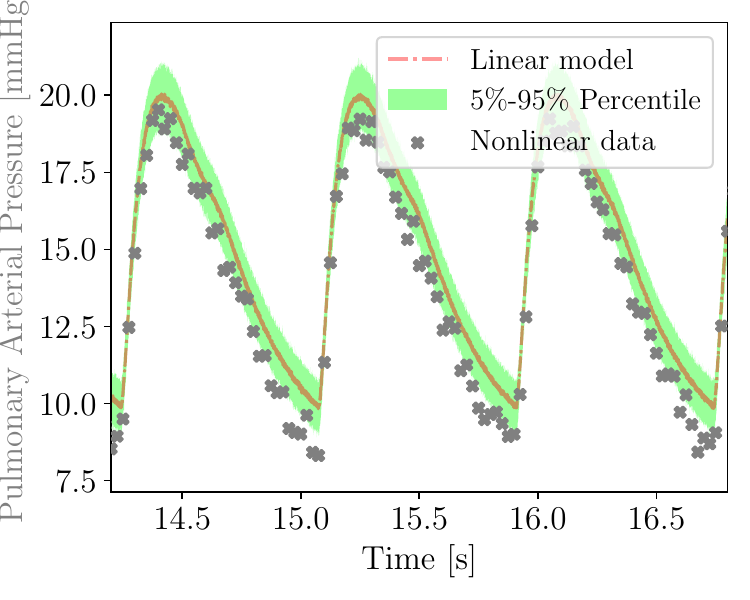}
    \includegraphics[width=0.24\textwidth]{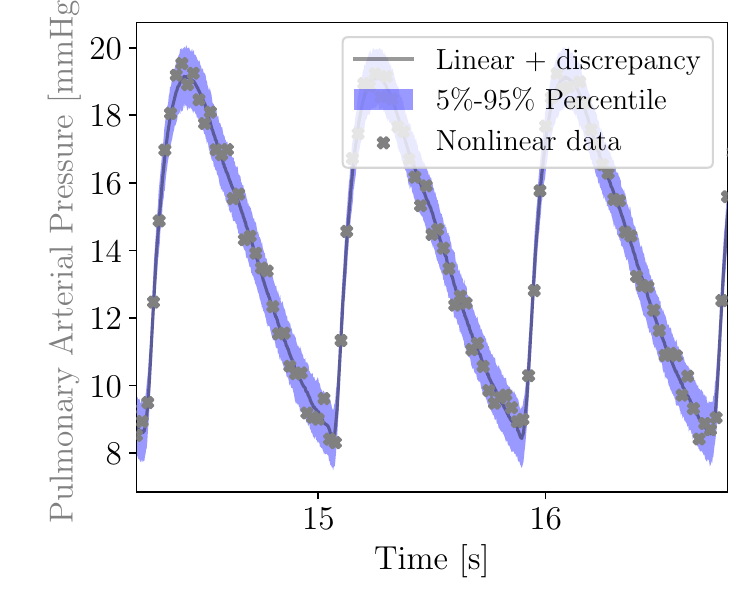}

    \includegraphics[width=0.24\textwidth]{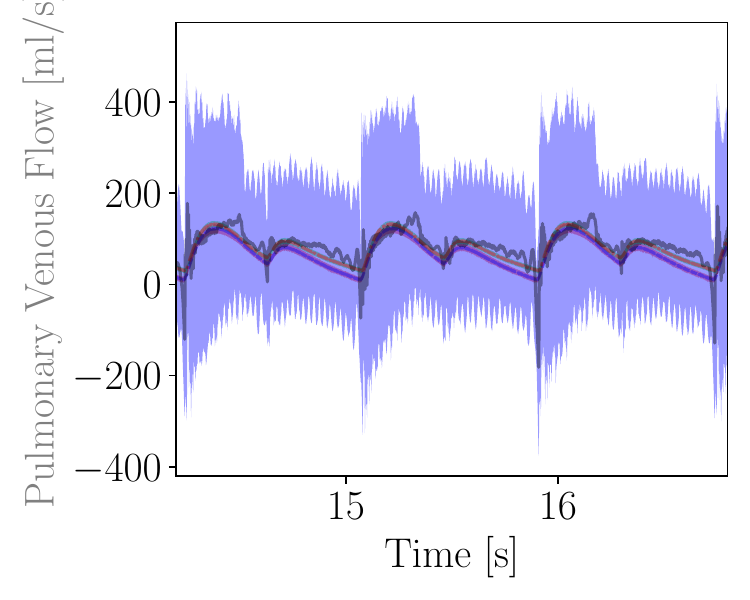}
    \includegraphics[width=0.24\textwidth]{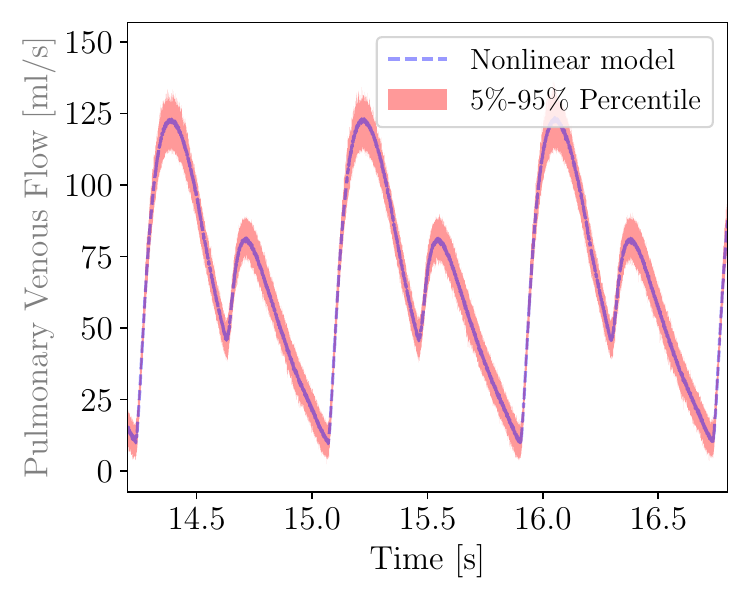}
    \includegraphics[width=0.24\textwidth]{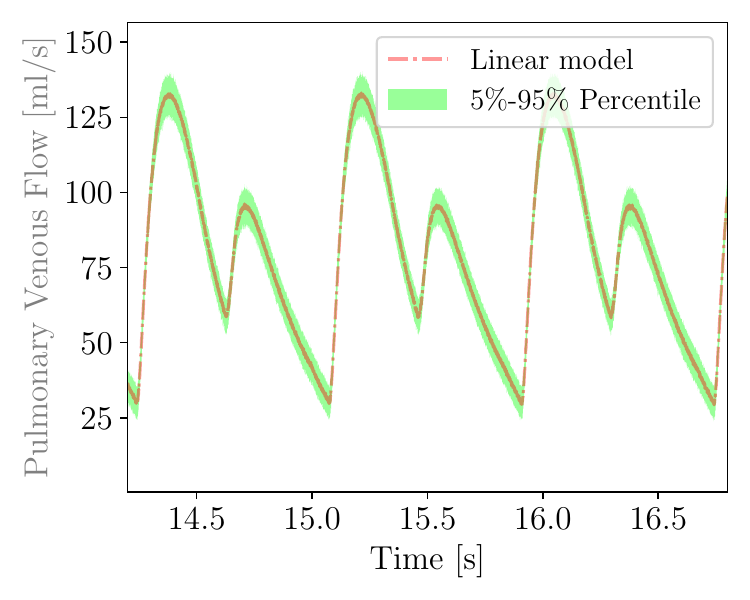}
    \includegraphics[width=0.24\textwidth]{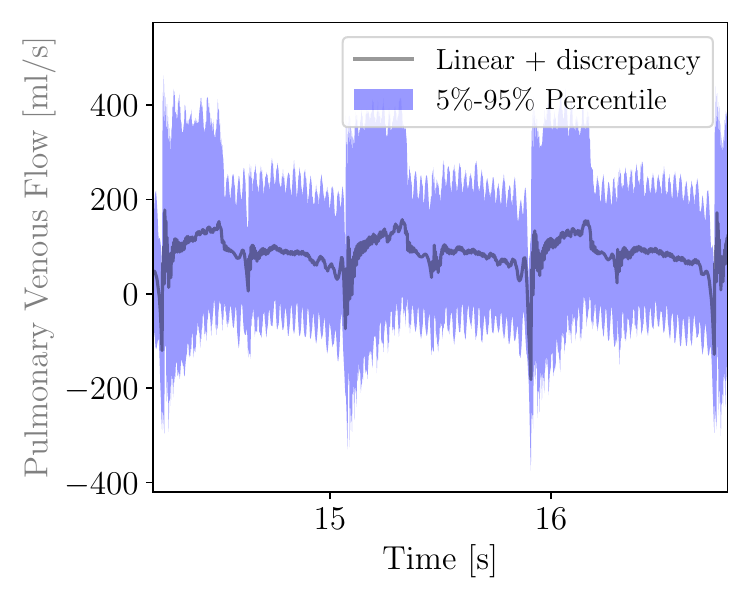}
    \caption{Reconstructed pulmonary arterial pressure (top) and pulmonary venous flow (bottom) under total uncertainty, for true nonlinear (second column), misspecified linear (third column) and linear plus discrepancy (right) model formulations. The first column shows a superposition of all average reconstructions and associated 5-95\% confidence regions, whereas the same reconstructions and confidence regions are shown separately on the other columns for improved clarity. For the pressure predictions shown in the top row, the available data are also plotted using gray crosses. Pressure bias due to a misspecified linear pulmonary resistance (top row) is significantly reduced with the addition of a discrepancy term as shown in~\eqref{eq:disc_pv_model}. However, stiffness in the CVSim-6 ODE equations and lack of intertia result in large confidence intervals and even non physiological negative pulmonary flow (bottom row).}
    \label{fig:totuq_pressure_pa_pv}
\end{figure}

We now consider MC X-TFC reconstruction under total uncertainty, including model-form uncertainty.
Results for pulmonary arterial pressure and pulmonary venous flow are reported in Figure~\ref{fig:totuq_pressure_pa_pv}. 
While no bias was observed for the linear model under ideal conditions, adding noise to the pressure observations induces bias consisting of a moderate increase in the pulmonary arterial pressure for the linear model. 
However, once a discrepancy is introduced and estimated, the bias is practically eliminated, and the total uncertainty in the pressure almost coincides with the uncertainty of the true nonlinear model.
Conversely, the mean estimated pulmonary flow shows significant oscillations and is associated with substantial total uncertainty, even nonphysical negative values. This is due to two factors: the first relates to modeling assumptions in the CVSim-6 model, which lacks any form of inertance, allowing sudden variations in flow to go unopposed by the system. 
The second relates to the regularity of the flow discrepancy, which is governed by the equation of a capacitor, which includes the derivative of pressure reconstructions for $P_{pa}$ and $P_{pv}$ that are affected by significant oscillations due to the stiffness of CVSim-6 at systole.
In other words, modeling assumptions combined with the specific choice of the variable chosen to represent the discrepancy directly affect the quantification of model-form and total uncertainty. 

\begin{figure}[ht!]
\centering
    \includegraphics[width=0.24\textwidth]{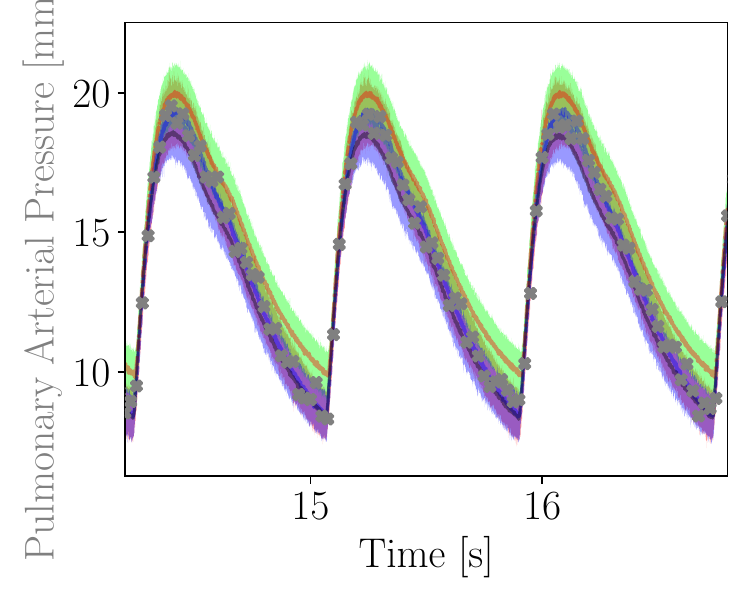}
    \includegraphics[width=0.24\textwidth]{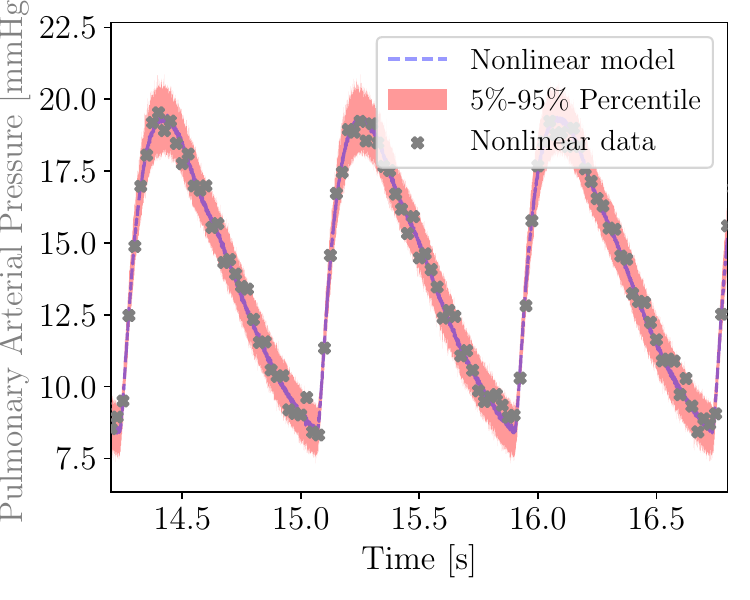}
    \includegraphics[width=0.24\textwidth]{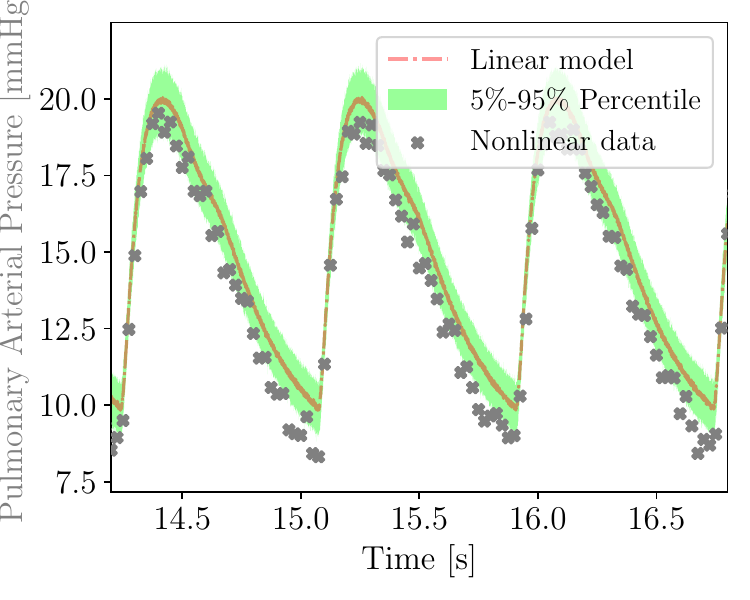}
    \includegraphics[width=0.24\textwidth]{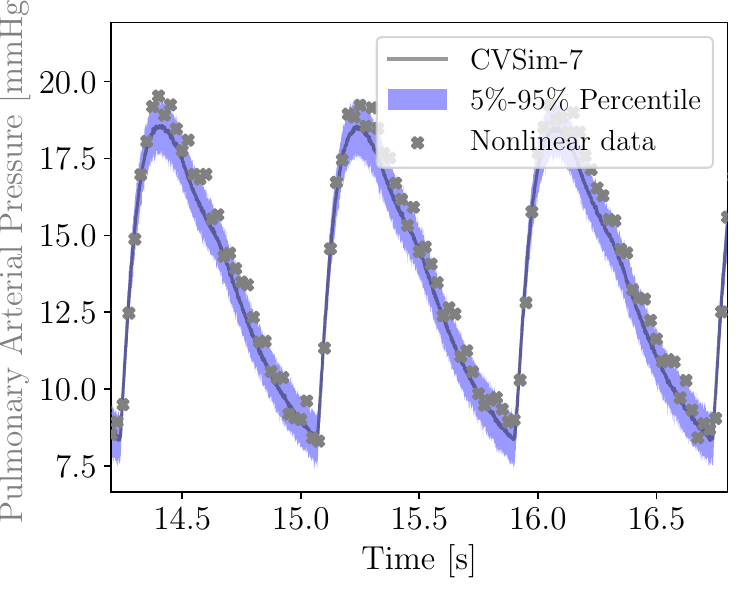}

    \includegraphics[width=0.24\textwidth]{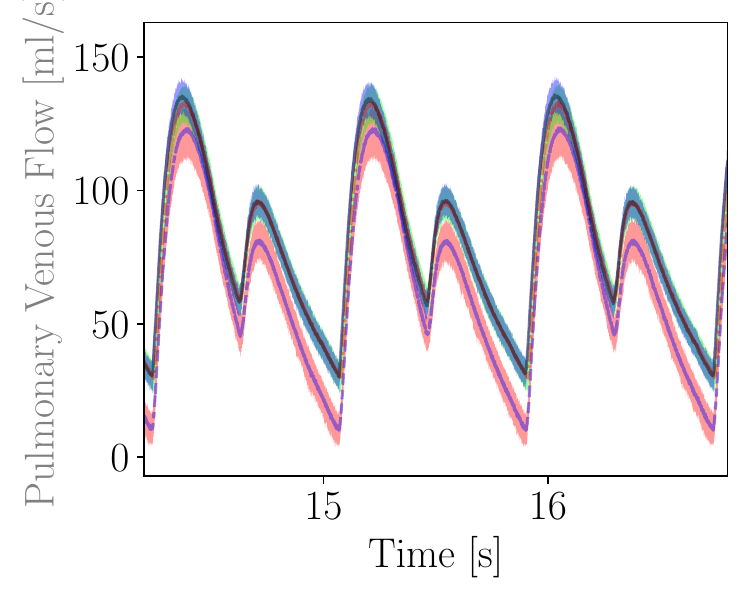}
    \includegraphics[width=0.24\textwidth]{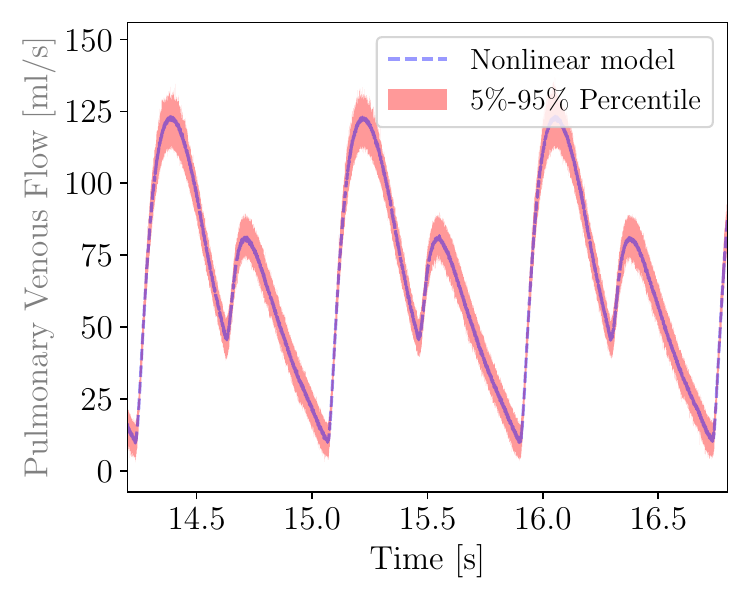}
    \includegraphics[width=0.24\textwidth]{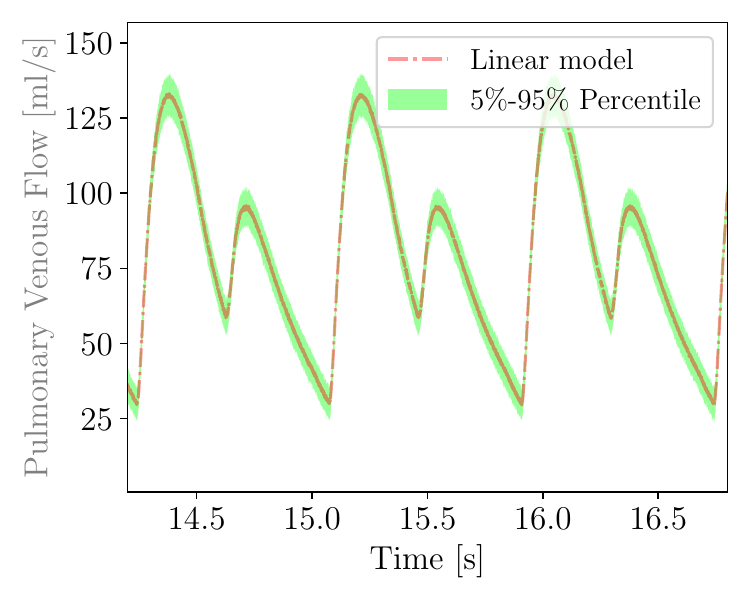}
    \includegraphics[width=0.24\textwidth]{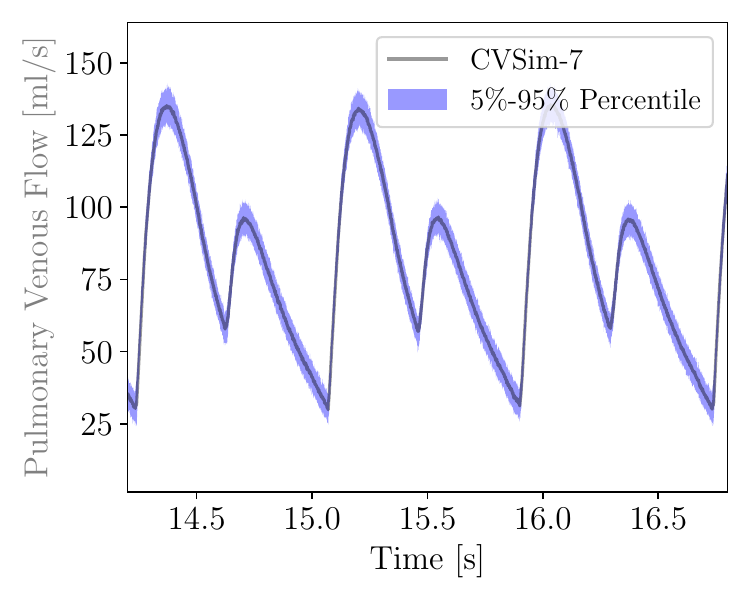}
    \caption{Reconstructed pulmonary arterial pressure (top) and pulmonary venous flow (bottom) under total uncertainty, for true nonlinear (second column), misspecified linear (third column) and linear plus discrepancy (right) model formulations for the system with additional inertia.
    The first column shows a superposition of all average reconstructions and associated 5-95\% confidence regions, whereas the same reconstructions and confidence regions are shown separately on the other columns for improved clarity. 
    For the pressure predictions shown in the top row, the available data are also plotted using gray crosses. 
    Pressure bias due to a misspecified linear pulmonary resistance (top row) is still reduced with the addition of a discrepancy term, this time formulated as an inductance equation in~\eqref{equ:discr_two}.
    Additionally, a more physiologically consistent discrepancy formulation greatly reduces total uncertainty in pulmonary flow (bottom row).}
    \label{fig:results_inductance}
\end{figure}

A simple remedy is to add an inductor equation for the flow discrepancy of the form 
\begin{equation}\label{equ:discr_two}
    \dot{\delta}(t) = \frac{P_{pa}(t) - P_{pv}(t)}{L_{pv}},
\end{equation}
where $L_{pv}$ represents the inductance of the pulmonary flow discrepancy, for this case chosen as $L_{pv} = 10\,r_{pv}$, and the $\delta(t)$ is now approximated via a new TFC constrained expression as 
\begin{equation}
    \delta(t,\bm{\beta}) = (\boldsymbol{\sigma} - \boldsymbol{\sigma}_0) \boldsymbol{\beta}_{\delta} + \delta_{0},    
\end{equation}
where the initial condition $\delta_{0}$ is arbitrarily set to 0. This addresses two problems at the same time by both adding inertia and, therefore, improving the realism of the CVSim-6 model, and posing additional conditions on the regularity of the flow discrepancy. We refer to this new model configuration as \emph{CVSim-7} due to the additional differential equation introduced for the discrepancy. 
The results are shown in Figure~\ref{fig:results_inductance}, where the discrepancy is able to reduce the bias in $P_{pa}$ while producing a physiologically consistent pulmonary venous flow.

The results for the discrepancy models are obtained with subdomains of length 0.01 seconds, 10 collocation points per subdomain, and 10 neurons. Computing the average discrepancy requires a computational time of about 6 seconds per MC realization, totaling about 10 minutes for the whole simulation and total uncertainty quantification for 100 MC realizations. Again, MC X-TFC confirms its robustness and efficiency for online estimation under model-form uncertainty and does not require offline training.

\section{Discussion and conclusions}\label{sec:conclusions}

This study focuses on characterizing aleatoric, epistemic, and model-form uncertainty in the physics-informed reconstruction of synthetic responses generated by ODE systems.
We first illustrate the performance of PINNs, B-PINNs, and MC X-TFC on a single-equation differential model where the solution is reconstructed under total uncertainty, i.e., combining aleatoric, epistemic, and model-form uncertainty.
We then use MC X-TFC to estimate the physiologic response of the CVSim-6 compartmental model, first considering aleatoric and epistemic uncertainty in an ablation study, and then adding model-form uncertainty to consider misspecification in the pulmonary compartment.
MC X-TFC produces accurate reconstructions with limited uncertainty (up to $\sim$3 mmHg on systemic arterial pressure, and generally smaller than variability in the clinical assessment of pressures, flows, and volumes), with fast execution and without requiring offline training.

By progressively removing data and estimating an increasing number of parameters under a correctly formulated model, we observe the transition from a competitive to a cooperative interaction between data-informed and physics-informed approximation of the true underlying ODE solution. 
In addition, we have shown that the specific formulation of an unobserved discrepancy term, introduced to compensate for model misspecification, strongly affects model-form uncertainty. 
While this issue can be mitigated by adding data, we chose to provide additional regularization through physics by adding an inductance, which injects inertia to the CVSim-6 system, while also posing additional conditions on the rate of change of the pulmonary flow. 
This reduces estimation bias in the pulmonary arterial pressure, and also results in a physiologically sound estimate of the pulmonary flow.

It’s important to note that this study focuses on the online reconstruction of an ODE solution based on equations and data, differing from other parameter estimation approaches that rely on offline processes. In those approaches, offline work is used to learn the forward or inverse map between model parameters and outputs, or to establish prior or posterior information that informs the online process. This distinction is evident from the results of Sc6 in our ablation study. When two parameters need to be estimated jointly, the information removed from the system must be compensated by data from co-located compartments to ensure the underlying physical response is well-defined. The independence between compartments and the lack of inertia in the CVSim-6 equations seem to limit the number of parameters that can be estimated from a given dataset, relying on correlations between parameters or redundancies in model components. This observation suggests potential future work in extending X-TFC with a hybrid offline/online estimation approach.

While this study focuses on understanding the interaction between physics-informed regularization and total uncertainty for compartmental models in physiology, several limitations may restrict the ability to generalize these findings to more realistic conditions. First, we consider a Gaussian noise model on top of synthetically generated data instead of real patient-specific data. Second, some of the pressure measurements considered available in our study would be difficult to continuously measure in patients, or would only be available as time statistics, e.g., as mean, systolic or diastolic values. Third, the CVSim-6 is a relatively simple model for the human physiology and does not account for a number of important physiological mechanisms, including a four-chamber hearth model, cardiorespiratory coupling or pressure autoregulation mechanisms. Finally, the range of model parameter considered in this study relate to healthy conditions in human adults, but the model and X-TFC estimation process could be adapted to specific conditions and measurement scenarios to inform on specific conditions.

Future work will focus on using MC X-TFC to estimate a larger number of physiologically relevant parameters, more complex models, or on applications to ICU patients, where clinical data is continuously acquired from multiple sensors. In such scenario, fast online estimation informed by continuously acquired clinical signals is crucial for enhancing model-based patient monitoring and assessing critical conditions.

\section*{Acknowledgements}\label{sec:acks}

MD, ZZ and GEK acknowledge the support from the NIH grant ``Neural Operator Learning to Predict Aneurysmal Growth and Outcomes'' (R01HL168473) and the NIH grant ``CRCNS: Waste-clearance flows in the brain inferred using physics-informed neural networks'' (R01AT012312). DES acknowledges support from NSF CAREER award \#1942662 and NSF CDS\&E award \#2104831. 

%

\appendix

\section{Default parameters from the CVSim-6 cardiovascular model}\label{appendix_a}

This section provides a list of default CVSim-6 model parameters, including basic parameters (Table~\ref{table:params_basic}), and values of capacitance (Table~\ref{table:params_cap}), resistance (Table~\ref{table:params_res}) and unstressed volume (Table~\ref{table:params_vols}), corresponding to a healthy physiological response. 

\begin{table}[ht!]
\caption{Basic physiological quantities.}\label{table:params_basic}
\centering
\begin{tabular}{@{} l l l l @{}}
\toprule
Num. & Description & Ref. & Unit\\
\midrule
1. & Heart rate ($Hr$) & 72.00 & (bpm)\\
2. & Transthoracic pressure ($P_{th}$) & -4.00 & (mmHg)\\
3. & Systolic ratio per cardiac cycle ($r_{sys}$) & 0.33 & $-$\\ 
\bottomrule
\end{tabular}
\end{table}

\begin{table}[ht!]
\caption{Capacitances of each compartment.}\label{table:params_cap}
\centering
\begin{tabular}{@{} l l l l @{}}
\toprule
Num. & Description & Ref. & Unit\\
\midrule
4. & Left ventricular diastolic capacitance ($C_{l,dia}$) & $7.50\cdot 10^{-3}$ & (mL/Barye)\\ 
5. & Left ventricular systolic capacitance ($C_{l,sys}$) & $3.00\cdot 10^{-4}$ & (mL/Barye)\\ 
6. & Arterial capacitance ($C_a$) & $1.20\cdot 10^{-3}$  & (mL/Barye)\\ 
7. & Venous capacitance ($C_v$) & $7.50\cdot 10^{-2}$   & (mL/Barye)\\ 
8. & Right ventricular diastolic capacitance ($C_{r,dia}$) & $1.50\cdot 10^{-2}$ & (mL/Barye)\\ 
9. & Right ventricular systolic capacitance ($C_{r,sys}$) & $9.00\cdot 10^{-4}$ & (mL/Barye)\\ 
10. & Pulmonary arterial capacitance ($C_{pa}$) & $3.23\cdot10^{-3}$ & (mL/Barye)\\ 
11. & Pulmonary venous capacitance ($C_{pv}$) & $6.30\cdot 10^{-3}$ & (mL/Barye)\\ 
\bottomrule
\end{tabular}
\end{table}

\begin{table}[ht!]
\caption{Resistance of each compartment (Outflow resistance equal to the inflow resistance of the following compartment.)}\label{table:params_res}
\centering
\begin{tabular}{@{} l l l l @{}}
\toprule
Num. & Description & Ref. & Unit\\
\midrule
12. & Left ventricular input resistance ($R_{l,in}$)  & 13.33  & (Barye$\cdot$s/mL)\\ 
13. & Left ventricular output resistance ($R_{l,out}$) & 8.00 & (Barye$\cdot$s/mL)\\  
14. & Arterial resistance ($R_a$) & 1333.22  &(Barye$\cdot$s/mL)\\ 
15. & Right ventricular input resistance ($R_{r,in}$)  & 66.66  & (Barye$\cdot$s/mL)\\ 
16. & Right ventricular output resistance ($R_{r,out}$) & 4.00  &(Barye$\cdot$s/mL)\\  
17. & Pulmonary venous resistance ($R_{pv}$) & 106.66 & (Barye$\cdot$s/mL)\\ 
\bottomrule
\end{tabular}
\end{table}

\begin{table}[ht!]
\caption{Unstressed volume of each compartment.}\label{table:params_vols}
\centering
\begin{tabular}{@{} l l l l @{}}
\toprule
Num. & Description & Ref. & Unit\\
\midrule
18. & Unstressed left ventricular volume ($V_{l}^0$) & 15.00 & (mL) \\
19. & Unstressed arterial volume ($V_{a}^0$) & 715.00 & (mL) \\
20. & Unstressed venous volume ($V_{v}^0$) & 2500.00 & (mL) \\
21. & Unstressed right ventricular volume ($V_{r}^0$) & 15.00 & (mL) \\
22. & Unstressed pulmonary arterial volume ($V_{pa}^0$) & 90.00 & (mL) \\
23. & Unstressed pulmonary venous volume ($V_{pv}^0$) & 490.00 & (mL) \\
\bottomrule
\end{tabular}
\end{table}

\section{Details of methods compared in the introductory example}\label{appendix_b}
This section provides details of methods (the MC X-TFC method, the ensemble PINNs method, and the B-PINNs method) compared in Section \ref{sec:pedagogical}. $1,000$ Monte Carlo simulations were performed for the MC X-TFC method, $10$ PINNs were independently trained in the ensemble PINNs method, and $1,000$ posterior samples were taken in Hamiltonian Monte Carlo (HMC) along with $2,000$ burn-in samples for satisfactory acceptance rate \cite{yang2021b, zou2024neuraluq} in the B-PINNs method. The open-source NeuralUQ library \cite{zou2024neuraluq} was used for fast and convenient implementations of HMC in B-PINNs.
We note that although in this work, the training of X-TFC and PINNs was conducted sequentially, both the MC X-TFC and the ensemble PINNs methods are able to be accelerated by leveraging advanced vectorization and/or parallelization techniques for fast uncertainty quantification \cite{psaros2023uncertainty}.
In the MC X-TFC method, the width of the hidden layer was set to $20$, while in the ensemble PINNs and the B-PINNs methods, the neural network was chosen to have one hidden layer with $50$ neurons for acceptable results. The activation function was set to the hyperbolic tangent in all methods. 
In the training of PINNs, the Adam optimizer \cite{kingma2014adam} with $1\times10^{-3}$ was employed for $30,000$ iterations.
The comparison was performed on a standard laptop CPU (13th Gen Intel(R) Core(TM) i9-13900HX with 2.20 GHz processor).

\section{Ablation study of MC X-TFC in quantifying epistemic uncertainty for the harmonic ODE solution}\label{appendix_c}

\begin{figure}[h!]
\centering
    \includegraphics[width=\textwidth]{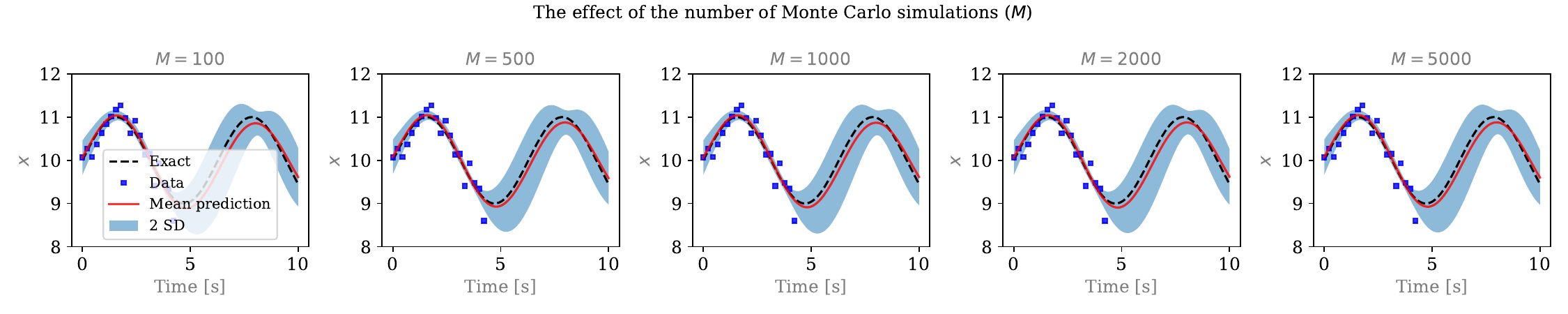}
    \includegraphics[width=\textwidth]{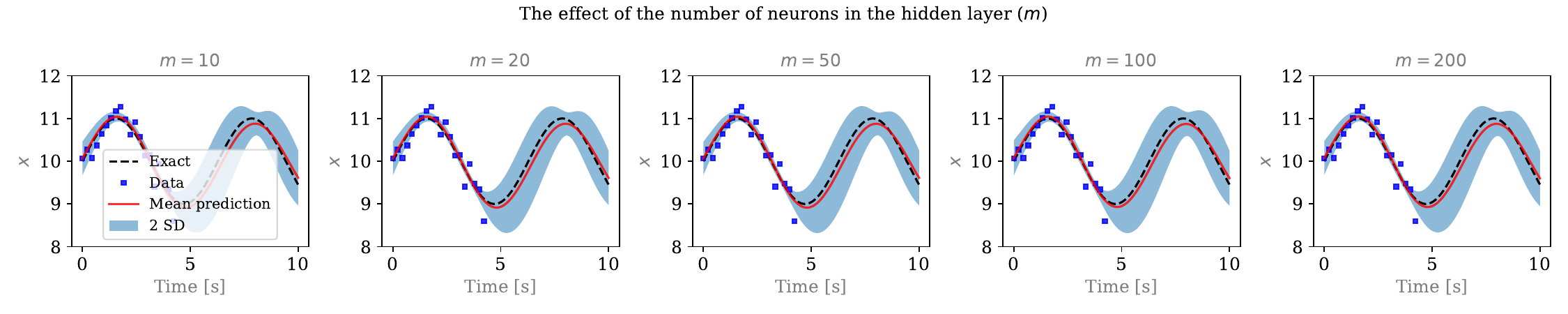}
    \includegraphics[width=\textwidth]{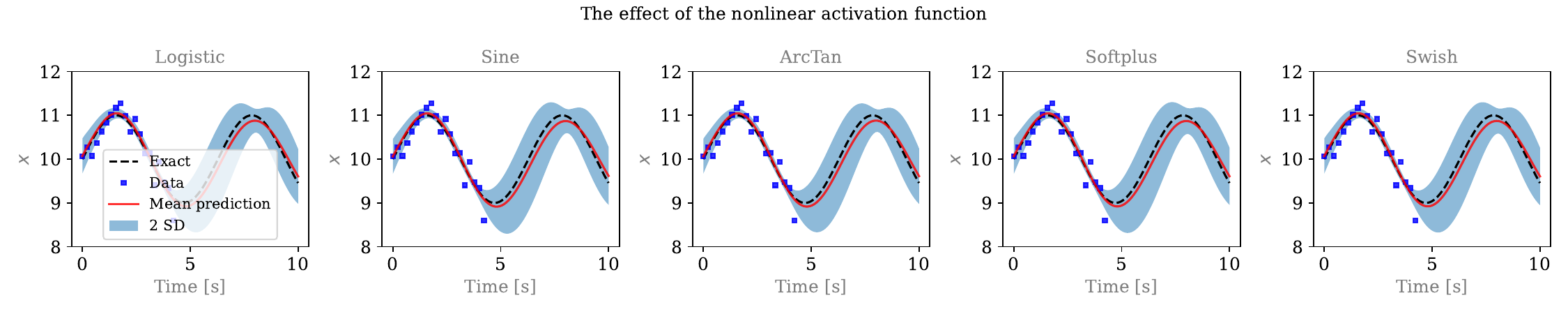}
    \includegraphics[width=\textwidth]{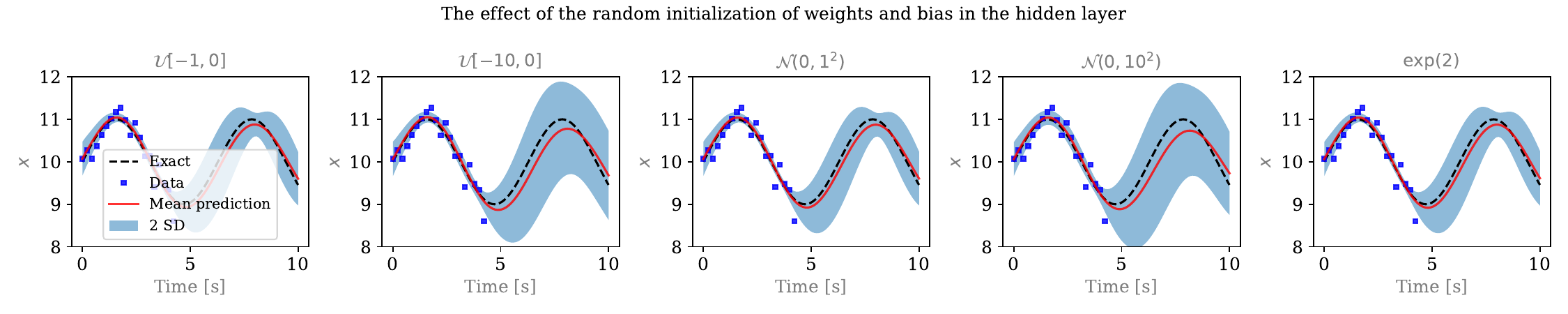}
    \caption{Ablation study of MC X-TFC in quantifying epistemic uncertainty for the harmonic ODE solution in Section \ref{sec:pedagogical_1}. Here $M$ denotes the number of Monte Carlo simulations, $m$ denotes the number of neurons in the hidden layer of X-TFC, $\mathcal{U}$ denotes the 1D uniform distribution, $\mathcal{N}$ denotes the 1D Gaussian distribution, and $\exp(\mu)$ denotes the exponential distribution with mean $\mu$.}
    \label{fig:example_0_ablation}
\end{figure}

In this section we conduct an ablation study of the proposed method, i.e. MC X-TFC, in quantifying epistemic uncertainty for the example presented in Section \ref{sec:pedagogical_1}. Specifically, we investigate the effect of the number of Monte Carlo simulations (denoted as $M$), the number of neurons in the hidden layer (denoted as $m$), the nonlinear activation function, and the initialization method of the weights and biases in the hidden layer. Here we investigate five activation functions (apart from the hyperbolic tangent): logistic function, sine function, inverse tangent function, Softplus activation function, and Swish activation function \cite{ramachandran2017searching}.
The baseline parameters used in all the examples in the paper are $M=1,000$, $m=20$, hyperbolic tangent and the uniform distribution on $[-1, 1]$ (denoted as $\mathcal{U}[-1, 1]$), unless stated otherwise.

As shown in the first three rows of Figure \ref{fig:example_0_ablation}, results of MC X-TFC are consistent across different numbers of Monte Carlo simulations, numbers of neurons in the hidden layer, and activation functions.
The fourth row of Figure \ref{fig:example_0_ablation} presents results of MC X-TFC with the following five random initialization distributions: the uniform distribution $\mathcal{U}[-1, 0]$, the uniform distribution $\mathcal{U}[-10, 0]$, the normal distribution $\mathcal{N}(0, 1^2)$, the normal distribution $\mathcal{N}(0, 10^2)$, and $\exp(2)$ where $\exp(\mu)$ denotes the exponential distribution with mean $\mu$. We note that the random initialization distribution $\mathcal{U}[-1, 0]$ only yields non-positive weights and biases in the hidden layer, while $\exp(2)$ generates only strictly positive values. Nonetheless, they are able to produce results similar to the one with $\mathcal{U}[-1, 1]$ (presented in Figure \ref{fig:example_0_0}), demonstrating that MC X-TFC is robust to the choice of random initialization. However, we observe that the predicted epistemic uncertainty is sensitive to the scale of randomly initialized weights and biases, which is consistent with results presented in Section \ref{sec:pedagogical_1}, i.e., a larger scale leads to a larger epistemic uncertainty.

\bibliography{main}

\begin{thebibliography}{100}

\bibitem{psaros2023uncertainty}
Apostolos~F Psaros, Xuhui Meng, Zongren Zou, Ling Guo, and George~Em Karniadakis.
\newblock Uncertainty quantification in scientific machine learning: Methods, metrics, and comparisons.
\newblock {\em Journal of Computational Physics}, 477:111902, 2023.

\bibitem{raissi2019physics}
Maziar Raissi, Paris Perdikaris, and George~E Karniadakis.
\newblock Physics-informed neural networks: A deep learning framework for solving forward and inverse problems involving nonlinear partial differential equations.
\newblock {\em Journal of Computational physics}, 378:686--707, 2019.

\bibitem{eivazi2022physics}
Hamidreza Eivazi, Mojtaba Tahani, Philipp Schlatter, and Ricardo Vinuesa.
\newblock Physics-informed neural networks for solving {Reynolds-averaged Navier--Stokes} equations.
\newblock {\em Physics of Fluids}, 34(7), 2022.

\bibitem{raissi2020hidden}
Maziar Raissi, Alireza Yazdani, and George~Em Karniadakis.
\newblock Hidden fluid mechanics: Learning velocity and pressure fields from flow visualizations.
\newblock {\em Science}, 367(6481):1026--1030, 2020.

\bibitem{cheng2021deep}
Chen Cheng and Guang-Tao Zhang.
\newblock Deep learning method based on physics informed neural network with {r}esnet block for solving fluid flow problems.
\newblock {\em Water}, 13(4):423, 2021.

\bibitem{wessels2020neural}
Henning Wessels, Christian Wei{\ss}enfels, and Peter Wriggers.
\newblock The neural particle method--{An} updated {Lagrangian} physics informed neural network for computational fluid dynamics.
\newblock {\em Computer Methods in Applied Mechanics and Engineering}, 368:113127, 2020.

\bibitem{shukla2024neurosem}
Khemraj Shukla, Zongren Zou, Chi~Hin Chan, Additi Pandey, Zhicheng Wang, and George~Em Karniadakis.
\newblock {NeuroSEM: A hybrid framework for simulating multiphysics problems by coupling PINNs and spectral elements}.
\newblock {\em arXiv preprint arXiv:2407.21217}, 2024.

\bibitem{linka2022bayesian}
Kevin Linka, Amelie Sch{\"a}fer, Xuhui Meng, Zongren Zou, George~Em Karniadakis, and Ellen Kuhl.
\newblock {Bayesian} physics informed neural networks for real-world nonlinear dynamical systems.
\newblock {\em Computer Methods in Applied Mechanics and Engineering}, 402:115346, 2022.

\bibitem{kharazmi2021identifiability}
Ehsan Kharazmi, Min Cai, Xiaoning Zheng, Zhen Zhang, Guang Lin, and George~Em Karniadakis.
\newblock Identifiability and predictability of integer-and fractional-order epidemiological models using physics-informed neural networks.
\newblock {\em Nature Computational Science}, 1(11):744--753, 2021.

\bibitem{schiassi2021physics}
Enrico Schiassi, Mario De~Florio, Andrea D’Ambrosio, Daniele Mortari, and Roberto Furfaro.
\newblock Physics-informed neural networks and functional interpolation for data-driven parameters discovery of epidemiological compartmental models.
\newblock {\em Mathematics}, 9(17):2069, 2021.

\bibitem{zou2024neuraluq}
Zongren Zou, Xuhui Meng, Apostolos~F Psaros, and George~E Karniadakis.
\newblock {NeuralUQ}: A comprehensive library for uncertainty quantification in neural differential equations and operators.
\newblock {\em SIAM Review}, 66(1):161--190, 2024.

\bibitem{lu2021physics}
Lu~Lu, Raphael Pestourie, Wenjie Yao, Zhicheng Wang, Francesc Verdugo, and Steven~G Johnson.
\newblock Physics-informed neural networks with hard constraints for inverse design.
\newblock {\em SIAM Journal on Scientific Computing}, 43(6):B1105--B1132, 2021.

\bibitem{gao2022physics}
Han Gao, Matthew~J Zahr, and Jian-Xun Wang.
\newblock Physics-informed graph neural galerkin networks: A unified framework for solving pde-governed forward and inverse problems.
\newblock {\em Computer Methods in Applied Mechanics and Engineering}, 390:114502, 2022.

\bibitem{cuomo2022scientific}
Salvatore Cuomo, Vincenzo~Schiano Di~Cola, Fabio Giampaolo, Gianluigi Rozza, Maziar Raissi, and Francesco Piccialli.
\newblock Scientific machine learning through physics--informed neural networks: Where we are and what’s next.
\newblock {\em Journal of Scientific Computing}, 92(3):88, 2022.

\bibitem{karniadakis2021physics}
George~Em Karniadakis, Ioannis~G Kevrekidis, Lu~Lu, Paris Perdikaris, Sifan Wang, and Liu Yang.
\newblock Physics-informed machine learning.
\newblock {\em Nature Reviews Physics}, 3(6):422--440, 2021.

\bibitem{blei2017variational}
D.M. Blei, A.~Kucukelbir, and J.D. McAuliffe.
\newblock Variational inference: A review for statisticians.
\newblock {\em Journal of the American Statistical Association}, 112(518):859--877, 2017.

\bibitem{hoffman2013stochastic}
Matthew~D Hoffman, David~M Blei, Chong Wang, and John Paisley.
\newblock Stochastic variational inference.
\newblock {\em Journal of Machine Learning Research}, 2013.

\bibitem{ranganath2014black}
R.~Ranganath, S.~Gerrish, and D.~Blei.
\newblock Black box variational inference.
\newblock In {\em Artificial Intelligence and Statistics}, pages 814--822, 2014.

\bibitem{gal2016dropout}
Yarin Gal and Zoubin Ghahramani.
\newblock {Dropout as a Bayesian approximation: Representing model uncertainty in deep learning}.
\newblock In {\em international conference on machine learning}, pages 1050--1059. PMLR, 2016.

\bibitem{lakshminarayanan2017simple}
Balaji Lakshminarayanan, Alexander Pritzel, and Charles Blundell.
\newblock Simple and scalable predictive uncertainty estimation using deep ensembles.
\newblock {\em Advances in Neural Information Processing Systems}, 30, 2017.

\bibitem{yang2021b}
Liu Yang, Xuhui Meng, and George~Em Karniadakis.
\newblock {B-PINNs: Bayesian} physics-informed neural networks for forward and inverse pde problems with noisy data.
\newblock {\em Journal of Computational Physics}, 425:109913, 2021.

\bibitem{zou2024leveraging}
Zongren Zou, Tingwei Meng, Paula Chen, J{\'e}r{\^o}me Darbon, and George~Em Karniadakis.
\newblock {Leveraging viscous Hamilton-Jacobi PDEs for uncertainty quantification in scientific machine learning}.
\newblock {\em arXiv preprint arXiv:2404.08809}, 2024.

\bibitem{zou2024correcting}
Zongren Zou, Xuhui Meng, and George~Em Karniadakis.
\newblock Correcting model misspecification in physics-informed neural networks {(PINNs)}.
\newblock {\em Journal of Computational Physics}, 505:112918, 2024.

\bibitem{zou2023uncertainty}
Zongren Zou, Xuhui Meng, and George~Em Karniadakis.
\newblock Uncertainty quantification for noisy inputs-outputs in physics-informed neural networks and neural operators.
\newblock {\em arXiv preprint arXiv:2311.11262}, 2023.

\bibitem{zou2023hydra}
Zongren Zou and George~Em Karniadakis.
\newblock {L-HYDRA: Multi-head physics-informed neural networks}.
\newblock {\em arXiv preprint arXiv:2301.02152}, 2023.

\bibitem{meng2022learning}
Xuhui Meng, Liu Yang, Zhiping Mao, Jos{\'e} del {\'A}guila~Ferrandis, and George~Em Karniadakis.
\newblock Learning functional priors and posteriors from data and physics.
\newblock {\em Journal of Computational Physics}, 457:111073, 2022.

\bibitem{mortari2017theory}
Daniele Mortari.
\newblock The theory of connections: Connecting points.
\newblock {\em Mathematics}, 5(4):57, 2017.

\bibitem{mortari2017least}
Daniele Mortari.
\newblock Least-squares solution of linear differential equations.
\newblock {\em Mathematics}, 5(4):48, 2017.

\bibitem{mortari2019high}
Daniele Mortari, Hunter Johnston, and Lidia Smith.
\newblock High accuracy least-squares solutions of nonlinear differential equations.
\newblock {\em Journal of computational and applied mathematics}, 352:293--307, 2019.

\bibitem{de2021theory}
Mario De~Florio, Enrico Schiassi, Andrea D’Ambrosio, Daniele Mortari, and Roberto Furfaro.
\newblock Theory of functional connections applied to linear odes subject to integral constraints and linear ordinary integro-differential equations.
\newblock {\em Mathematical and computational applications}, 26(3):65, 2021.

\bibitem{leake2020multivariate}
Carl Leake, Hunter Johnston, and Daniele Mortari.
\newblock The multivariate theory of functional connections: Theory, proofs, and application in partial differential equations.
\newblock {\em Mathematics}, 8(8):1303, 2020.

\bibitem{schiassi2021extreme}
Enrico Schiassi, Roberto Furfaro, Carl Leake, Mario De~Florio, Hunter Johnston, and Daniele Mortari.
\newblock Extreme theory of functional connections: A fast physics-informed neural network method for solving ordinary and partial differential equations.
\newblock {\em Neurocomputing}, 457:334--356, 2021.

\bibitem{huang2006extreme}
Guang-Bin Huang, Qin-Yu Zhu, and Chee-Kheong Siew.
\newblock Extreme learning machine: theory and applications.
\newblock {\em Neurocomputing}, 70(1-3):489--501, 2006.

\bibitem{dwivedi2020physics}
Vikas Dwivedi and Balaji Srinivasan.
\newblock Physics informed extreme learning machine (pielm)--a rapid method for the numerical solution of partial differential equations.
\newblock {\em Neurocomputing}, 391:96--118, 2020.

\bibitem{dong2021local}
Suchuan Dong and Zongwei Li.
\newblock Local extreme learning machines and domain decomposition for solving linear and nonlinear partial differential equations.
\newblock {\em Computer Methods in Applied Mechanics and Engineering}, 387:114129, 2021.

\bibitem{wang2024extreme}
Yiran Wang and Suchuan Dong.
\newblock An extreme learning machine-based method for computational pdes in higher dimensions.
\newblock {\em Computer Methods in Applied Mechanics and Engineering}, 418:116578, 2024.

\bibitem{sun2024local}
Jingbo Sun, Suchuan Dong, and Fei Wang.
\newblock Local randomized neural networks with discontinuous galerkin methods for partial differential equations.
\newblock {\em Journal of Computational and Applied Mathematics}, 445:115830, 2024.

\bibitem{alvarez2024nonlinear}
Hector~Vargas Alvarez, Gianluca Fabiani, Nikolaos Kazantzis, Ioannis~G Kevrekidis, and Constantinos Siettos.
\newblock Nonlinear discrete-time observers with physics-informed neural networks.
\newblock {\em Chaos, Solitons \& Fractals}, 186:115215, 2024.

\bibitem{fabiani2023parsimonious}
Gianluca Fabiani, Evangelos Galaris, Lucia Russo, and Constantinos Siettos.
\newblock Parsimonious physics-informed random projection neural networks for initial value problems of odes and index-1 daes.
\newblock {\em Chaos: An Interdisciplinary Journal of Nonlinear Science}, 33(4), 2023.

\bibitem{dong2022computing}
Suchuan Dong and Jielin Yang.
\newblock On computing the hyperparameter of extreme learning machines: Algorithm and application to computational pdes, and comparison with classical and high-order finite elements.
\newblock {\em Journal of Computational Physics}, 463:111290, 2022.

\bibitem{calabro2021extreme}
Francesco Calabr{\`o}, Gianluca Fabiani, and Constantinos Siettos.
\newblock Extreme learning machine collocation for the numerical solution of elliptic pdes with sharp gradients.
\newblock {\em Computer Methods in Applied Mechanics and Engineering}, 387:114188, 2021.

\bibitem{dong2023method}
Suchuan Dong and Yiran Wang.
\newblock A method for computing inverse parametric pde problems with random-weight neural networks.
\newblock {\em Journal of Computational Physics}, 489:112263, 2023.

\bibitem{patsatzis2024physics}
Dimitrios~G Patsatzis, Lucia Russo, and Constantinos Siettos.
\newblock A physics-informed neural network method for the approximation of slow invariant manifolds for the general class of stiff systems of odes.
\newblock {\em arXiv preprint arXiv:2403.11591}, 2024.

\bibitem{liu2023bayesian}
Xu~Liu, Wen Yao, Wei Peng, and Weien Zhou.
\newblock Bayesian physics-informed extreme learning machine for forward and inverse pde problems with noisy data.
\newblock {\em Neurocomputing}, 549:126425, 2023.

\bibitem{fabiani2024randonet}
Gianluca Fabiani, Ioannis~G Kevrekidis, Constantinos Siettos, and Athanasios~N Yannacopoulos.
\newblock Randonet: Shallow-networks with random projections for learning linear and nonlinear operators.
\newblock {\em arXiv preprint arXiv:2406.05470}, 2024.

\bibitem{drozd2021energy}
Kristofer Drozd, Roberto Furfaro, Enrico Schiassi, Hunter Johnston, and Daniele Mortari.
\newblock Energy-optimal trajectory problems in relative motion solved via theory of functional connections.
\newblock {\em Acta Astronautica}, 182:361--382, 2021.

\bibitem{de2021solutions}
Mario De~Florio, Enrico Schiassi, Roberto Furfaro, Barry~D Ganapol, and Domiziano Mostacci.
\newblock Solutions of chandrasekhar’s basic problem in radiative transfer via theory of functional connections.
\newblock {\em Journal of quantitative spectroscopy and radiative transfer}, 259:107384, 2021.

\bibitem{johnston2020fuel}
Hunter Johnston, Enrico Schiassi, Roberto Furfaro, and Daniele Mortari.
\newblock Fuel-efficient powered descent guidance on large planetary bodies via theory of functional connections.
\newblock {\em The journal of the astronautical sciences}, 67(4):1521--1552, 2020.

\bibitem{drozd2019constrained}
K~Drozd, R~Furfaro, and D~Mortari.
\newblock Constrained energy-optimal guidance in relative motion via theory of functional connections and rapidly-explored random trees.
\newblock In {\em Proceedings of the 2019 Astrodynamics Specialist Conference, Portland, ME, USA}, pages 11--15, 2019.

\bibitem{de2022physics}
Mario De~Florio, Enrico Schiassi, and Roberto Furfaro.
\newblock Physics-informed neural networks and functional interpolation for stiff chemical kinetics.
\newblock {\em Chaos: An Interdisciplinary Journal of Nonlinear Science}, 32(6), 2022.

\bibitem{schiassi2020new}
Enrico Schiassi, Andrea D’Ambrosio, Hunter Johnston, Mario De~Florio, Kristofer Drozd, Roberto Furfaro, Fabio Curti, Daniele Mortari, et~al.
\newblock Physics-informed extreme theory of functional connections applied to optimal orbit transfer.
\newblock In {\em Proceedings of the AAS/AIAA Astrodynamics Specialist Conference, Lake Tahoe, CA, USA}, pages 9--13, 2020.

\bibitem{de2021physics}
Mario De~Florio, Enrico Schiassi, Barry~D Ganapol, and Roberto Furfaro.
\newblock Physics-informed neural networks for rarefied-gas dynamics: Thermal creep flow in the bhatnagar--gross--krook approximation.
\newblock {\em Physics of Fluids}, 33(4), 2021.

\bibitem{de2022poiseuille}
Mario De~Florio, Enrico Schiassi, Barry~D Ganapol, and Roberto Furfaro.
\newblock Physics-informed neural networks for rarefied-gas dynamics: Poiseuille flow in the bgk approximation.
\newblock {\em Zeitschrift f{\"u}r angewandte Mathematik und Physik}, 73(3):126, 2022.

\bibitem{schiassi2022orbit}
Enrico Schiassi, Andrea D’Ambrosio, Kristofer Drozd, Fabio Curti, and Roberto Furfaro.
\newblock Physics-informed neural networks for optimal planar orbit transfers.
\newblock {\em Journal of Spacecraft and Rockets}, 59(3):834--849, 2022.

\bibitem{de2024physics}
Mario De~Florio, Enrico Schiassi, Francesco Calabr{\`o}, and Roberto Furfaro.
\newblock Physics-informed neural networks for 2nd order odes with sharp gradients.
\newblock {\em Journal of Computational and Applied Mathematics}, 436:115396, 2024.

\bibitem{furfaro2022physics}
Roberto Furfaro, Andrea D'Ambrosio, Enrico Schiassi, and Andrea Scorsoglio.
\newblock Physics-informed neural networks for closed-loop guidance and control in aerospace systems.
\newblock In {\em AIAA SCITECH 2022 Forum}, page 0361, 2022.

\bibitem{schiassi2022physics}
Enrico Schiassi, Mario De~Florio, Barry~D Ganapol, Paolo Picca, and Roberto Furfaro.
\newblock Physics-informed neural networks for the point kinetics equations for nuclear reactor dynamics.
\newblock {\em Annals of Nuclear Energy}, 167:108833, 2022.

\bibitem{d2021pontryagin}
Andrea D’ambrosio, Enrico Schiassi, Fabio Curti, and Roberto Furfaro.
\newblock Pontryagin neural networks with functional interpolation for optimal intercept problems.
\newblock {\em Mathematics}, 9(9):996, 2021.

\bibitem{schiassi2020physics}
Enrico Schiassi, Andrea D'Ambrosio, Mario De~Florio, Roberto Furfaro, and Fabio Curti.
\newblock Physics-informed extreme theory of functional connections applied to data-driven parameters discovery of epidemiological compartmental models.
\newblock {\em arXiv preprint arXiv:2008.05554}, 2020.

\bibitem{ahmadi2024ai}
Nazanin Ahmadi~Daryakenari, Mario De~Florio, Khemraj Shukla, and George~Em Karniadakis.
\newblock Ai-aristotle: A physics-informed framework for systems biology gray-box identification.
\newblock {\em PLOS Computational Biology}, 20(3):e1011916, 2024.

\bibitem{de2023ai}
Mario De~Florio, Ioannis~G Kevrekidis, and George~Em Karniadakis.
\newblock Ai-lorenz: A physics-data-driven framework for black-box and gray-box identification of chaotic systems with symbolic regression.
\newblock {\em arXiv preprint arXiv:2312.14237}, 2023.

\bibitem{chowell2017fitting}
Gerardo Chowell.
\newblock Fitting dynamic models to epidemic outbreaks with quantified uncertainty: A primer for parameter uncertainty, identifiability, and forecasts.
\newblock {\em Infectious Disease Modelling}, 2(3):379--398, 2017.

\bibitem{hastings2013population}
Alan Hastings.
\newblock {\em Population biology: concepts and models}.
\newblock Springer Science \& Business Media, 2013.

\bibitem{smolen2000mathematical}
Paul Smolen, Douglas~A Baxter, and John~H Byrne.
\newblock Mathematical modeling of gene networks.
\newblock {\em Neuron}, 26(3):567--580, 2000.

\bibitem{wang2017review}
Lin Wang, Satyakam Dash, Chiam~Yu Ng, and Costas~D Maranas.
\newblock A review of computational tools for design and reconstruction of metabolic pathways.
\newblock {\em Synthetic and systems biotechnology}, 2(4):243--252, 2017.

\bibitem{stamatakis2006phylogenetic}
Alexandros Stamatakis.
\newblock Phylogenetic models of rate heterogeneity: a high performance computing perspective.
\newblock In {\em Proceedings 20th IEEE international parallel \& distributed processing symposium}, pages 8--pp. IEEE, 2006.

\bibitem{harvey2014works}
William Harvey.
\newblock {\em The Works of {W}illiam {H}arvey}.
\newblock University of Pennsylvania Press, 2014.

\bibitem{sutera1993history}
Salvatore~P Sutera and Richard Skalak.
\newblock The history of {P}oiseuille's law.
\newblock {\em Annual review of fluid mechanics}, 25(1):1--20, 1993.

\bibitem{frank1990basic}
Otto Frank.
\newblock The basic shape of the arterial pulse. first treatise: mathematical analysis.
\newblock {\em Journal of molecular and cellular cardiology}, 22(3):255--277, 1990.

\bibitem{milivsic2004analysis}
V.~Mili{\v{s}}i{\'c} and A.~Quarteroni.
\newblock Analysis of lumped parameter models for blood flow simulations and their relation with {1D} models.
\newblock {\em ESAIM: Mathematical Modelling and Numerical Analysis}, 38(4):613--632, 2004.

\bibitem{hughes1973one}
Thomas~JR Hughes and J~Lubliner.
\newblock On the one-dimensional theory of blood flow in the larger vessels.
\newblock {\em Mathematical Biosciences}, 18(1-2):161--170, 1973.

\bibitem{bazilevs2013computational}
Yuri Bazilevs, Kenji Takizawa, and Tayfun~E Tezduyar.
\newblock {\em Computational fluid-structure interaction: methods and applications}.
\newblock John Wiley \& Sons, 2013.

\bibitem{brown2023patient}
Jordan~A Brown, Jae~H Lee, Margaret~Anne Smith, David~R Wells, Aaron Barrett, Charles Puelz, John~P Vavalle, and Boyce~E Griffith.
\newblock Patient--specific immersed finite element--difference model of transcatheter aortic valve replacement.
\newblock {\em Annals of biomedical engineering}, 51(1):103--116, 2023.

\bibitem{verzicco2022electro}
Roberto Verzicco.
\newblock Electro-fluid-mechanics of the heart.
\newblock {\em Journal of Fluid Mechanics}, 941:P1, 2022.

\bibitem{zingaro2024electromechanics}
Alberto Zingaro, Michele Bucelli, Roberto Piersanti, Francesco Regazzoni, Alfio Quarteroni, et~al.
\newblock An electromechanics-driven fluid dynamics model for the simulation of the whole human heart.
\newblock {\em Journal of Computational Physics}, 504:112885, 2024.

\bibitem{sank2011sc}
Sethuraman Sankaran and Alison~L. Marsden.
\newblock {A Stochastic Collocation Method for Uncertainty Quantification and Propagation in Cardiovascular Simulations}.
\newblock {\em Journal of Biomechanical Engineering}, 133(3):031001, 02 2011.

\bibitem{schiavazzi2017generalized}
D.E. Schiavazzi, A.~Doostan, G.~Iaccarino, and A.L. Marsden.
\newblock A generalized multi-resolution expansion for uncertainty propagation with application to cardiovascular modeling.
\newblock {\em Computer methods in applied mechanics and engineering}, 314:196--221, 2017.

\bibitem{seo2020effects}
Jongmin Seo, Daniele~E Schiavazzi, Andrew~M Kahn, and Alison~L Marsden.
\newblock The effects of clinically-derived parametric data uncertainty in patient-specific coronary simulations with deformable walls.
\newblock {\em International journal for numerical methods in biomedical engineering}, 36(8):e3351, 2020.

\bibitem{tran2019uncertainty}
J.S. Tran, D.E Schiavazzi, A.M. Kahn, and A.L. Marsden.
\newblock Uncertainty quantification of simulated biomechanical stimuli in coronary artery bypass grafts.
\newblock {\em Computer Methods in Applied Mechanics and Engineering}, 345:402--428, 2019.

\bibitem{maher2021geometric}
Gabriel~D Maher, Casey~M Fleeter, Daniele~E Schiavazzi, and Alison~L Marsden.
\newblock Geometric uncertainty in patient-specific cardiovascular modeling with convolutional dropout networks.
\newblock {\em Computer methods in applied mechanics and engineering}, 386:114038, 2021.

\bibitem{harrod2021predictive}
Karlyn~K Harrod, Jeffrey~L Rogers, Jeffrey~A Feinstein, Alison~L Marsden, and Daniele~E Schiavazzi.
\newblock Predictive modeling of secondary pulmonary hypertension in left ventricular diastolic dysfunction.
\newblock {\em Frontiers in physiology}, 12:666915, 2021.

\bibitem{tran2016automated}
J.S. Tran, D.E. Schiavazzi, A.~Bangalore~Ramachandra, A.M. Kahn, and A.L. Marsden.
\newblock Automated tuning for parameter identification and uncertainty quantification in multi-scale coronary simulations.
\newblock {\em Computers \& Fluids}, 142:128--138, 2017.

\bibitem{schiavazzi2016patient}
D.E. Schiavazzi, A.~Baretta, G.~Pennati, T.Y. Hsia, and A.L. Marsden.
\newblock Patient-specific parameter estimation in single-ventricle lumped circulation models under uncertainty.
\newblock {\em International Journal for Numerical Methods in Biomedical Engineering}, 33(3), 2017.

\bibitem{schiavazzi2016uncertainty}
Daniele~E Schiavazzi, Gr{\'e}gory Arbia, Catriona Baker, Anthony~M Hlavacek, Tain-Yen Hsia, Alison~L Marsden, Irene~E Vignon-Clementel, and The~Modeling of~Congenital Hearts Alliance (MOCHA)~Investigators.
\newblock Uncertainty quantification in virtual surgery hemodynamics predictions for single ventricle palliation.
\newblock {\em International journal for numerical methods in biomedical engineering}, 32(3):e02737, 2016.

\bibitem{fleeter2020multilevel}
C.M. Fleeter, G.~Geraci, D.E. Schiavazzi, A.M. Kahn, and A.L. Marsden.
\newblock Multilevel and multifidelity uncertainty quantification for cardiovascular hemodynamics.
\newblock {\em Computer methods in applied mechanics and engineering}, 365:113030, 2020.

\bibitem{seo2020multifidelity}
Jongmin Seo, Casey Fleeter, Andrew~M Kahn, Alison~L Marsden, and Daniele~E Schiavazzi.
\newblock Multifidelity estimators for coronary circulation models under clinically informed data uncertainty.
\newblock {\em International Journal for Uncertainty Quantification}, 10(5), 2020.

\bibitem{pfaller2022automated}
Martin~R Pfaller, Jonathan Pham, Aekaansh Verma, Luca Pegolotti, Nathan~M Wilson, David~W Parker, Weiguang Yang, and Alison~L Marsden.
\newblock Automated generation of {0D} and {1D} reduced-order models of patient-specific blood flow.
\newblock {\em International Journal for Numerical Methods in Biomedical Engineering}, 38(10):e3639, 2022.

\bibitem{zanoni2024improved}
Andrea Zanoni, Gianluca Geraci, Matteo Salvador, Karthik Menon, Alison~L Marsden, and Daniele~E Schiavazzi.
\newblock Improved multifidelity monte carlo estimators based on normalizing flows and dimensionality reduction techniques.
\newblock {\em Computer Methods in Applied Mechanics and Engineering}, 429:117119, 2024.

\bibitem{schaferglobal}
Friederike Sch{\"a}fer, Daniele~E Schiavazzi, Leif~Rune Hellevik, and Jacob Sturdy.
\newblock Global sensitivity analysis with multifidelity monte carlo and polynomial chaos expansion for vascular haemodynamics.
\newblock {\em International Journal for Numerical Methods in Biomedical Engineering}, page e3836, 2024.

\bibitem{lee2024probabilistic}
John~D Lee, Jakob Richter, Martin~R Pfaller, Jason~M Szafron, Karthik Menon, Andrea Zanoni, Michael~R Ma, Jeffrey~A Feinstein, Jacqueline Kreutzer, Alison~L Marsden, et~al.
\newblock A probabilistic neural twin for treatment planning in peripheral pulmonary artery stenosis.
\newblock {\em International journal for numerical methods in biomedical engineering}, page e3820, 2024.

\bibitem{davis1991teaching}
Timothy~Lloyd Davis.
\newblock {\em Teaching physiology through interactive simulation of hemodynamics}.
\newblock PhD thesis, Massachusetts Institute of Technology, 1991.

\bibitem{heldt2010cvsim}
Thomas Heldt, Ramakrishna Mukkamala, George~B Moody, and Roger~G Mark.
\newblock {{CVS}im: an open-source cardiovascular simulator for teaching and research}.
\newblock {\em The open pacing, electrophysiology \& therapy journal}, 3:45, 2010.

\bibitem{mai2022theory}
Tina Mai and Daniele Mortari.
\newblock Theory of functional connections applied to quadratic and nonlinear programming under equality constraints.
\newblock {\em Journal of Computational and Applied Mathematics}, 406:113912, 2022.

\bibitem{ganaie2022ensemble}
Mudasir~A Ganaie, Minghui Hu, Ashwani~Kumar Malik, Muhammad Tanveer, and Ponnuthurai~N Suganthan.
\newblock Ensemble deep learning: A review.
\newblock {\em Engineering Applications of Artificial Intelligence}, 115:105151, 2022.

\bibitem{rahaman2021uncertainty}
Rahul Rahaman et~al.
\newblock Uncertainty quantification and deep ensembles.
\newblock {\em Advances in neural information processing systems}, 34:20063--20075, 2021.

\bibitem{wenzel2020hyperparameter}
Florian Wenzel, Jasper Snoek, Dustin Tran, and Rodolphe Jenatton.
\newblock Hyperparameter ensembles for robustness and uncertainty quantification.
\newblock {\em Advances in Neural Information Processing Systems}, 33:6514--6527, 2020.

\bibitem{gawlikowski2023survey}
Jakob Gawlikowski, Cedrique Rovile~Njieutcheu Tassi, Mohsin Ali, Jongseok Lee, Matthias Humt, Jianxiang Feng, Anna Kruspe, Rudolph Triebel, Peter Jung, Ribana Roscher, et~al.
\newblock A survey of uncertainty in deep neural networks.
\newblock {\em Artificial Intelligence Review}, 56(Suppl 1):1513--1589, 2023.

\bibitem{pickering2022discovering}
Ethan Pickering, Stephen Guth, George~Em Karniadakis, and Themistoklis~P Sapsis.
\newblock Discovering and forecasting extreme events via active learning in neural operators.
\newblock {\em Nature Computational Science}, 2(12):823--833, 2022.

\bibitem{zhang2024discovering}
Zhen Zhang, Zongren Zou, Ellen Kuhl, and George~Em Karniadakis.
\newblock Discovering a reaction--diffusion model for {Alzheimer’s disease by combining PINNs} with symbolic regression.
\newblock {\em Computer Methods in Applied Mechanics and Engineering}, 419:116647, 2024.

\bibitem{neal2012mcmc}
Radford~M Neal.
\newblock {MCMC using Hamiltonian dynamics}.
\newblock {\em arXiv preprint arXiv:1206.1901}, 2012.

\bibitem{pant2018lumped}
Sanjay Pant, Chiara Corsini, Catriona Baker, Tain-Yen Hsia, Giancarlo Pennati, and Irene~E Vignon-Clementel.
\newblock A lumped parameter model to study atrioventricular valve regurgitation in stage 1 and changes across stage 2 surgery in single ventricle patients.
\newblock {\em IEEE Transactions on Biomedical Engineering}, 65(11):2450--2458, 2018.

\bibitem{kingma2014adam}
Diederik Kingma and Jimmy Ba.
\newblock Adam: A method for stochastic optimization.
\newblock In {\em International Conference on Learning Representations (ICLR)}, San Diega, CA, USA, 2015.

\bibitem{ramachandran2017searching}
Prajit Ramachandran, Barret Zoph, and Quoc~V Le.
\newblock Searching for activation functions.
\newblock {\em arXiv preprint arXiv:1710.05941}, 2017.

\end{thebibliography}

\end{document}